\documentclass[12pt]{article}

\usepackage{scicite}
\usepackage{dsfont}
\usepackage{amsmath,amssymb}
\usepackage{times}

\usepackage{xcolor} 
\usepackage{bm}
\usepackage{xspace}
\usepackage{graphicx}
\usepackage{caption}
\usepackage{subcaption}
\usepackage{multirow}
\usepackage[toc]{appendix}
\usepackage{url} 
\usepackage[pagebackref=true,breaklinks=true,colorlinks,bookmarks=false,citecolor=blue]{hyperref}
\usepackage{relsize}
\usepackage{sectsty}
\usepackage{times}
\usepackage[shortlabels]{enumitem}

\newcommand{\gaussian}{\mathcal{N}}
\newcommand{\uniform}{\mathcal{U}}
\newcommand{\dclaw}{D'Claw\xspace}
\newcommand{\bigdataset}{\mathbb{B}}
\newcommand{\smalldataset}{\mathbb{S}}
\newcommand{\expsymbol}{\mathcal{E}}
\newcommand{\stusymbol}{\mathcal{S}}

\newcommand{\expert}{\pi^\expsymbol}
\newcommand{\student}{\pi^\stusymbol}

\newcommand{\figref}[1]{Figure~\ref{#1}}
\newcommand{\tblref}[1]{Table~\ref{#1}}
\renewcommand{\eqref}[1]{Equation~\ref{#1}}
\newcommand{\secref}[1]{Section~\ref{#1}}
\newcommand{\obs}{\bm{o}_t\xspace}

\newcommand{\act}{\bm{a}_t\xspace}

\newenvironment{sciabstract}{%
\begin{quote} \bf}
{\end{quote}}

\title{Visual Dexterity: In-Hand Reorientation of Novel and Complex Object Shapes}

\author
{Tao Chen$^{1,2}$, Megha Tippur$^{2}$, Siyang Wu$^{3}$, Vikash Kumar$^{4}$, \\ Edward Adelson$^{2}$, Pulkit Agrawal$^{\ast 1,2,5}$\\
\\
\normalsize{$^{1}$Improbable AI Laboratory, Massachusetts Institute of Technology}\\
\normalsize{Cambridge, MA 02139, USA}\\
\normalsize{$^{2}$Computer Science and Artificial Intelligence Laboratory (CSAIL)},\\ \normalsize{Massachusetts Institute of Technology,}\\
\normalsize{Cambridge, MA 02139, USA}\\
\normalsize{$^{3}$Institute for Interdisciplinary Information Sciences, }\\
\normalsize{Tsinghua University, Beijing, 100084, China} \\
\normalsize{$^{4}$Meta AI, Pittsburgh, PA 15213, USA} \\
\normalsize{$^{5}$Institute of Artificial Intelligence and Advanced Interactions (IAIFI)}\\
\normalsize{Massachusetts Institute of Technology,}\\
\normalsize{Cambridge, MA 02139, USA}\\
\normalsize{$^\ast$To whom correspondence should be addressed; E-mail: pulkitag@mit.edu.}
}

\date{}

\begin{document}

\baselineskip24pt

\maketitle

\begin{sciabstract}
In-hand object reorientation is necessary for performing many dexterous manipulation tasks, such as tool use in less structured environments that remain beyond the reach of current robots. Prior works built reorientation systems assuming one or many of the following: reorienting only specific objects with simple shapes, limited range of reorientation, slow or quasistatic manipulation, simulation-only results, the need for specialized and costly sensor suites, and other constraints which make the system infeasible for real-world deployment. We present a general object reorientation controller that does not make these assumptions. It uses readings from a single commodity depth camera to dynamically reorient complex and new object shapes by any rotation in real-time, with the median reorientation time being close to seven seconds. The controller is trained using reinforcement learning in simulation and evaluated in the real world on new object shapes not used for training, including the most challenging scenario of reorienting objects held in the air by a downward-facing hand that must counteract gravity during reorientation. Our hardware platform only uses open-source components that cost less than five thousand dollars. Although we demonstrate the ability to overcome assumptions in prior work, there is ample scope for improving absolute performance. For instance, the challenging duck-shaped object not used for training was dropped in 56 percent of the trials. When it was not dropped, our controller reoriented the object within 0.4 radians (23 degrees) 75 percent of the time.
  
\end{sciabstract}

\section*{Summary}
A real-time controller that dynamically reorients complex and new objects by any amount
using a single depth camera.

\section*{Introduction}
The human hand's dexterity is vital to a wide range of daily tasks such as re-arranging objects, loading dishes in a dishwasher, fastening bolts, cutting vegetables, and other forms of tool use both inside and outside households. Despite a long-standing interest in creating similarly capable robotic systems, current robots are far behind in their versatility, dexterity, and robustness. In-hand object reorientation, illustrated in Figure 1, is a specific dexterous manipulation problem where the goal is to manipulate a hand-held object from an arbitrary initial orientation to an arbitrary target orientation~\cite{mason1989robot,salisbury1982articulated,rus1999hand,mordatch2012contact,bai2014dexterous,kumar2014real,chen2021system}. Object reorientation occupies a special place in manipulation because it is a pre-cursor to flexible tool use. After picking a tool, the robot must orient the tool in an appropriate configuration to use it. For example, a screwdriver can only be used if its head is aligned with the top of the screw. Object reorientation is, therefore, not only a litmus test for dexterity but also an enabler for many downstream manipulation tasks. 

A reorientation system ready for the real world should satisfy multiple criteria: it should be able to reorient objects into any orientation, generalize to new objects, and operate in real-time using data from commodity sensors. Some seemingly benign setup choices can make the system impractical for real-world deployment. For instance, consider the choice of placing multiple cameras around the workspace to reduce occlusion in viewing the object being manipulated \cite{andrychowicz2020learning,akkaya2019solving}. For a mobile manipulator, such camera placements are impractical. Similarly, performing reorientation under the assumption that the hand is below the object (upwards facing hand configuration) \cite{andrychowicz2020learning,akkaya2019solving,nagabandi2020deep} instead of the hand holding the object from the top (downwards facing hand configuration) is much easier. With a downward-facing hand, the hand must manipulate the object while simultaneously counteracting gravity. Small errors in finger motion can result in the object falling down. The upward-facing hand assumption makes control easier, but it limits the downstream use of the reorientation skill in many tool-use applications. 

Even without real-world setup constraints, object reorientation is challenging because it requires coordinated movement between multiple fingers resulting in a high-dimensional control space. The robot must control the amount of applied force, when to apply it, and where the fingers should make and break contact with the object. The combination of continuous and discrete decisions leads to a challenging continuous-discrete optimization problem that is often computationally intractable. For computational feasibility, a majority of prior works constrain manipulation to simple convex shapes such as polygons or cylinders~\cite{furukawa2006dynamic,ishihara2006dynamic,kumar2014real,kumar2016learning,calli2017vision,sundaralingam2019relaxed, andrychowicz2020learning,van2015learning,abondance2020dexterous,bhatt2022surprisingly,calli2018path,zhu2019dexterous,rajeswaran2017learning,khandate2022feasibility}. Other simplifying assumptions include designing specific movement patterns of fingers~\cite{morgan2022complex,bhatt2022surprisingly}, assuming fingers never make and break contact with the object~\cite{sundaralingam2019relaxed,jeong2020learning}, hand being in an upward-facing configuration~\cite{bai2014dexterous,nagabandi2020deep,andrychowicz2020learning} or the manipulation being quasi-static~\cite{morgan2022complex,sievers2022learning}. Such assumptions restrict the applicability of reorientation to a limited set of objects, scenarios, or orientations (for example, along only a single axis). 

Complementary to the control problem is the issue of measuring the state information the controller requires, such as the object's pose, surface friction, whether the finger is in contact with the object, etc. Touch sensors provide local contact information but are not widely available as a plug-and-play module. The difficulty in using visual sensing is that fingers occlude the object during reorientation. Recent works employed RGBD (RGB and depth) cameras to estimate object pose but require a separate pose estimator to be trained per object, which limits their generalization to new object shapes~\cite{allshire2022transferring, morgan2022complex, akkaya2019solving, andrychowicz2020learning}. 

Due to challenges in perception and control, no prior work has demonstrated a real-world ready reorientation system. Although controlling directly from perception is hard, given the full low-dimensional representation of relevant state information such as the object's position, velocity, pose, and manipulator's proprioceptive state, it is possible to build a controller using deep reinforcement learning (RL) that successfully reorients diverse objects in simulation~\cite{chen2021system}. RL effectively leverages large amounts of interaction data to find an approximate solution to the computationally challenging optimization problem of solving for reorientation. However, as a result of requiring large amounts of data and full state information, today, such RL controllers can only be trained in simulation. This leaves at least two open questions: how to train controllers with sensors available in the real world such as visual inputs and whether controllers trained in simulation transfer to the real world (sim-to-real transfer problem). 

The difficulty in training RL controllers from visual inputs stems from the learner's need to simultaneously solve the problem of inferring the relevant state information (feature learning) and determining the optimal actions. If the optimal actions were known in advance, it would be simpler to train a model that predicts these actions from visual inputs (supervised learning). Such a two-stage teacher-student training paradigm, where first a control policy is trained via RL with full state information (teacher) and then a second student policy trained via supervised learning to mimic the teacher has been successfully used for several applications~\cite{chen2020learning,margolis2021learning,chen2021system,kumar2021rma,lee2020learning}. We found the major roadblock in learning a visual policy that works across diverse objects is the slow speed of rendering in simulation which resulted in training times of over 20 days with our compute resources. Such slow training makes experimentation infeasible. We devised a two-stage approach for training the vision policy that first uses a synthetic point cloud without the need for rendering and is then finetuned with rendered point cloud to reduce the sim-to-real gap. Our pipeline makes training $5\times$ times faster. The second consideration was the use of a sparse convolution neural network to represent the policy to process point clouds at the speed required for real-time feedback control ($12$Hz in our case). By directly predicting actions from point clouds, our approach bypasses the problem of consistently defining pose/keypoints across different objects, allowing for generalization to new shapes. 

The next challenge is in overcoming the sim-to-real gap. In dynamic in-hand object reorientation, both the robot and the object move quickly. Achieving precise control in a system with fast-changing dynamics is challenging. It becomes even more challenging when using a downward-facing hand as control failures are irreversible. Therefore, dynamic in-hand object reorientation poses a substantial sim-to-real transfer challenge. Some reasons for the sim-to-real gap are differences in motor/object dynamics, perception noise, and modeling approximations made by the simulator. For instance, contact models in fast simulators tend to be a crude approximation of reality, especially for non-convex objects~\cite{xu20216dls}. Whether sim-to-real transfer of reorientation controller is even possible for these complex object shapes remained unclear. 

The systematic choices of identifying the manipulator dynamics (details in Method section), domain randomization~\cite{tobin2017domain}, the design of reward function, and the hardware considerations, including the number of fingers and the fingertip material, reduced the sim-to-real gap.
We conducted experiments in the challenging downward-facing hand configuration. We tested the controller's ability to make use of an external support surface for reorientation (extrinsic dexterity~\cite{dafle2014extrinsic}) and the harder condition when the object is in the air without any supporting surface. The results show progress towards developing a real-time controller capable of dynamically reorienting new objects with complex shapes and diverse materials by any amount in the full space of rotations (SO(3), special orthogonal group in three dimensions) using inputs from just a single commodity depth camera and joint encoders. While there is substantial room for improvement, especially in achieving precise reorientation, our results provide evidence that sim-to-real transfer is possible for challenging tasks involving dynamic and contact-rich manipulation in less-structured settings than previously demonstrated. 

Finally, many prior efforts used custom or expensive manipulators (such as the Shadow Hand~\cite{andrychowicz2020learning,akkaya2019solving,nagabandi2020deep} costing over $\$100,000$) and often relied on sophisticated sensing equipments such as a motion capture system. Such a hardware stack is hard to replicate due to its cost and complexity. In contrast, our hardware setup costs less than $\$5,000$ and uses only open-source components, making it easier to replicate. Furthermore, our platform is not specific to object reorientation and can be used for other dexterous manipulation tasks. Due to the low barrier to entry, and the evidence that such a system can tackle a challenging manipulation task, our platform can democratize research in dexterous manipulation.

\section*{Results}
\begin{figure}[!htb]
    \centering
    \includegraphics[width=\linewidth]{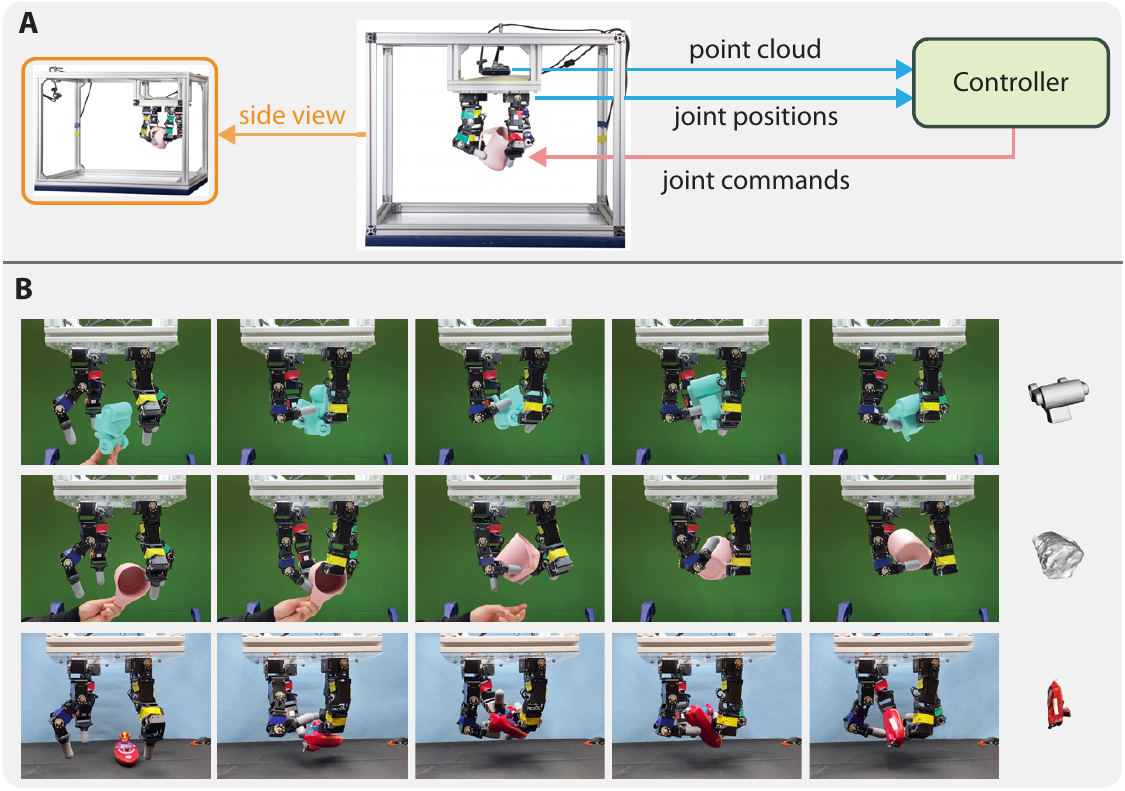}
    \caption{\textbf{Illustration of the robot system}. \textbf{(A)}: the front and side views of our real-world setup. The controller is a neural network that uses depth recordings from a single camera along with the joint positions of the manipulator to predict the change in joint positions. \textbf{(B)}: Visualization of the same controller reorienting three different objects. The rightmost column shows the target orientation. The first two rows are instances of a four-fingered hand reorienting objects in the air. The last row shows reorientation with the help of a supporting surface (extrinsic dexterity).}
    \label{fig:teaser}
\end{figure}

We trained a single controller to reorient $150$ objects from an arbitrary initial to a target configuration in simulation. The learned controllers are deployed in the real world on the open-source three-fingered \dclaw manipulator~\cite{ahn2020robel} and a modified four-fingered version with nine and twelve degrees of freedom (DoFs), respectively. The robot's observation is a depth image captured from a single Intel RealSense camera and the proprioceptive state of the fingers. The goal is provided as the point cloud of the object in a target configuration in the SO(3) space. The initial configuration of the object is a random transformation in SE(3)(special
Euclidean group in three dimensions) space within the range of the robot's fingers -- either the object is set on a table or handed over by a human to the robot. 

We experimented with the hand in the downward-facing configuration in two settings: with and without a supporting table. Our system runs in real-time at a control frequency of $12$ Hz using a commodity workstation. Figure 1 shows the intermediate steps of manipulating three objects to target orientations depicted in the rightmost column. The proposed controller reorients a diverse set of new objects with complex geometries not used for training. The main text movie provides a short summary of our results with audio. Movie S1 shows our system reorienting many objects and provides a more detailed summary of our major findings. Movie S2 visualizes the setting where the robot is tasked with a sequence of target orientations. In such a scenario, it has to stop when it reaches the current target orientation and then restart to achieve the next target.

For quantitative evaluation, we use seven objects from the training dataset ($\bigdataset$), which we refer to as in-distribution, and five objects from the held-out test dataset ($\smalldataset$), which we refer to as out-of-distribution (OOD). Objects are shown in Figure 2A. We test each object 20 times with random initial and goal orientation in each testing condition. We 3D print these objects to ensure the shape of objects in simulation and the real world is identical, which is helpful in evaluating the extent of sim-to-real transfer. While the shape of these seven objects is included in the training set, the surface properties such as friction of the real-world objects, may not correspond to any object used for training in simulation. Evaluation on five OOD objects tests generalization to shapes. To further showcase generalization to shapes and different material properties, we also present results on some rigid objects from daily life.  The orientation errors are measured using an OptiTrack motion capture system that tracks object pose. We define error as the distance between the goal and the object's orientation when the controller predicts it has reached the goal and stops. The motion capture is only used for evaluation and is not required by our controller otherwise.

\begin{figure}[!htb]
    \centering
    \includegraphics[width=0.82\linewidth]{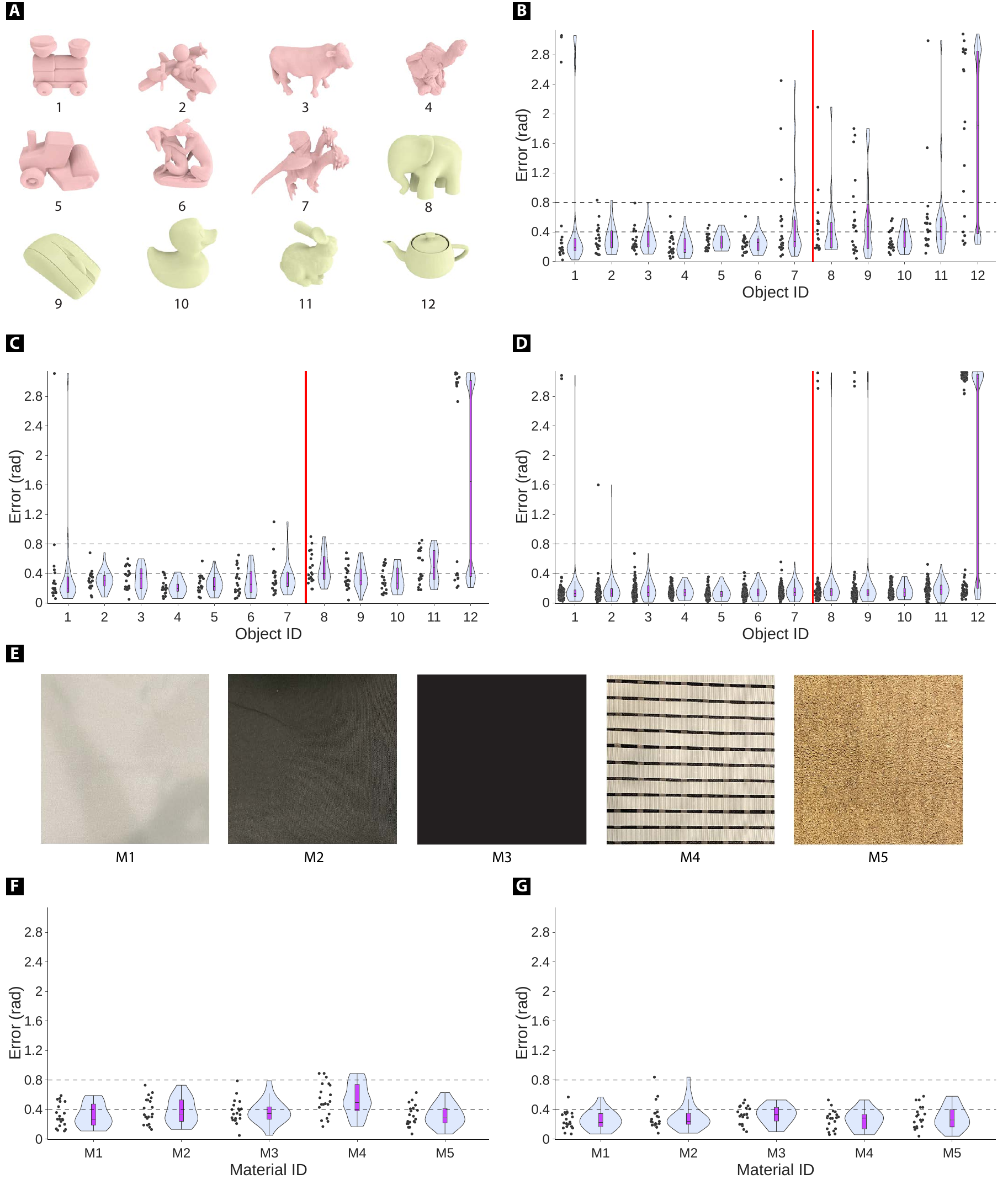}
    \caption{
    \textbf{Experimental results of reorientation}. \textbf{(A)}: twelve objects with their IDs. The first seven objects are from the training dataset $\bigdataset$, and the last five are from the testing dataset $\smalldataset$. \textbf{(B)}, \textbf{(C)} show the real-world error distribution when using rigid and soft fingertips, respectively, on material M1. \textbf{(D)} shows the error distribution in simulation for each object as a violin plot \cite{hintze1998violin}. The violet rectangle shows the errors within [25\%, 75\%] percentile and the horizontal bar in the rectangle depicts the median error. Train objects can mostly be reoriented within an error of 0.4 radians, with similar performance for rigid and soft fingertips. The error on test objects is higher, and soft fingertips exhibit better generalization. \textbf{(E)}: five table materials. \textbf{(F)} and \textbf{(G)} show the error distribution on different materials for object $\#5$ and $\#10$, respectively.}
    \label{fig:error_soft_rigid}
\end{figure}

\subsection*{Extrinsic dexterity: object reorientation with a supporting surface}
We first report results on the easier problem of reorienting objects when the table is present below the hand to support the object. Using an external surface to aid reorientation has been referred to as extrinsic dexterity \cite{dafle2014extrinsic} and is necessary in many real-world use cases. Visualization of the proposed controller reorienting a diverse set of objects is provided in Figure 3. To demonstrate the versatility of our system, we present results of the robot manipulating objects of different shapes, materials, surfaces, fingertip materials, and varying numbers of fingers. 

\begin{figure}[!tb]
    \centering
\includegraphics[width=\linewidth]{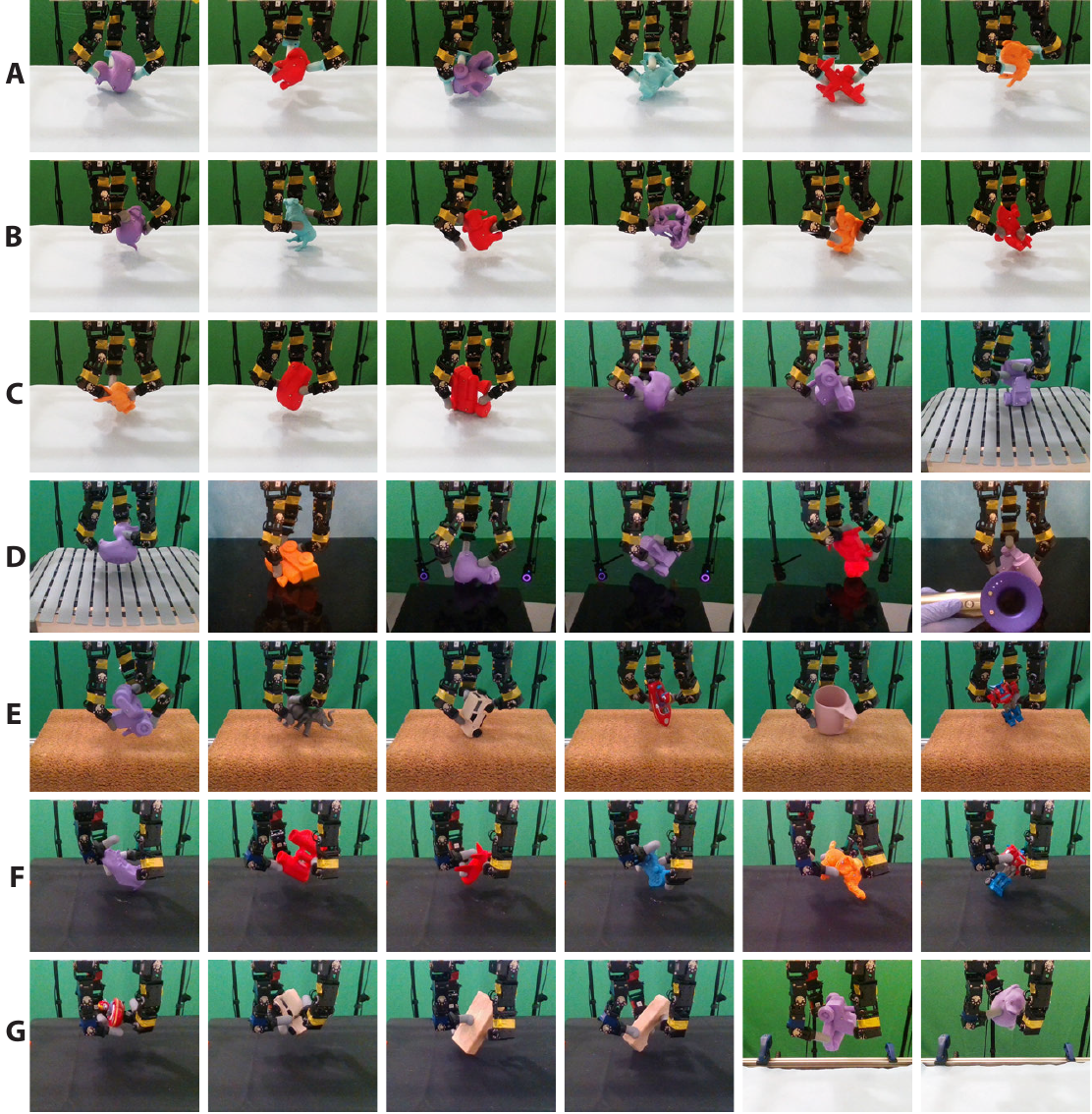}
    \caption{\textbf{Different testing scenarios}. We test our controller on objects with diverse shapes and reorientation conditions such as using different supporting surfaces such as a tablecloth, an uneven door mat, a slippery acrylic sheet, and a perforated bath mat. We also evaluate performance using fingertips with different softness: rigid 3D-printed (row (\textbf{A})), and soft elastomer fingertips (rows (\textbf{B}) to (\textbf{G})). Row (\textbf{A}) to (\textbf{E}) use a three-fingered robot hand. And row (\textbf{F}) to (\textbf{G}) use a four-fingered robot hand. Our policy can reorient real household objects (rows (\textbf{E},\textbf{G})) and can operate without the need for a supporting surface (in the air) as shown in row (\textbf{G}).
    }
    \label{fig:diff_scene}
\end{figure}

\subsubsection*{Reorientation using a three-fingered manipulator with rigid and soft fingertips} 
With table support, we found three fingers to suffice for the reorientation task. The error distribution for different objects, when tested on a table surface covered with a white cloth (material M1 in Figure 2E), is shown in Figure 2B using a violin plot \cite{hintze1998violin}. Although the overall error distribution is more informative, for ease of comparison, in Table 1, following the success threshold used in previous work~\cite{andrychowicz2020learning}, we report summary statistics of success rate measured as the percentage of tests with error within $0.4$  or $0.8$ radians. The seven train objects can be reoriented within an error of $0.4$ radians $81\%$ of the time. On the five OOD test objects, the success rate is lower at $45\%$. As expected, the performance is better with a relaxed error threshold of $0.8$ radians and worse at stricter thresholds.

\begin{table}[t!]
\centering
\caption{\textbf{Statistics of the orientation error when the hand reorients objects on a table}. \textbf{CI} stands for bias-corrected and accelerated (BCa) bootstrap confidence interval. \textbf{Train} stands for testing on the seven objects (Figure 2A) from the training dataset $\bigdataset$. \textbf{Test} stands for testing on the five objects from the testing dataset $\smalldataset$.}
\label{tbl:success_rate_table}
\resizebox{\columnwidth}{!}{
\begin{tabular}{c|cccccc} 
\hline
\multirow{2}{*}{}             & \multicolumn{2}{c}{\begin{tabular}[c]{@{}c@{}}with rigid fingertips \\(real)\end{tabular}} & \multicolumn{2}{c}{\begin{tabular}[c]{@{}c@{}}with soft fingertips\\(real)\end{tabular}} & \multicolumn{2}{c}{\begin{tabular}[c]{@{}c@{}}in simulation\end{tabular}}  \\ 
\cline{2-7}
                              & Train                                 & Test                                                                & Train                                 & Test                                                               & Train                                 & Test                                                               \\ 
\hline
$\leq 0.4$ radians  ($22.9^\circ$)               & $81\%$                             & $45\%$                                                             & $79\%$                             & $55\%$                                                            & $96\%$                             & $85\%$                                                            \\
{\color{gray}$95\%$ CI}  & \multicolumn{1}{l}{{\color{gray}$[73\%, 90\%]$}} & \multicolumn{1}{l}{{\color{gray}$[32\%, 58\%]$}}                                 & \multicolumn{1}{l}{{\color{gray}$[71\%, 86\%]$}} & \multicolumn{1}{l}{{\color{gray}$[44\%, 62\%]$}}                                & \multicolumn{1}{l}{{\color{gray}$[94\%, 97\%]$}} & \multicolumn{1}{l}{{\color{gray}$[82\%, 88\%]$}}                            \\
\hline
$\leq 0.8$ radians   ($45.8^\circ$)              & $95\%$                             & $75\%$                                                             & $98\%$                             & $86\%$                                                            & $98\%$                             & $87\%$                                                            \\
{\color{gray}$95\%$ CI}  & \multicolumn{1}{l}{{\color{gray}$[88\%, 98\%]$}} & \multicolumn{1}{l}{{\color{gray}$[46\%, 91\%]$}} & \multicolumn{1}{l}{{\color{gray}$[96\%, 99\%]$}} & \multicolumn{1}{l}{{\color{gray}$[58\%, 96\%]$}} & \multicolumn{1}{l}{{\color{gray}$[97\%, 99\%]$}} & \multicolumn{1}{l}{{\color{gray}$[84\%, 90\%]$}}    \\
\hline
\begin{tabular}[c]{@{}c@{}}$95\%$ CI of the median \\of orientation errors (radian)\end{tabular} & $[0.20, 0.27]$   & $[0.29, 0.46]$  & $[0.21, 0.28]$  & $[0.33, 0.42]$ & $[0.12, 0.13]$  & $[0.15, 0.18]$                                                    \\
\hline
\end{tabular}
}
\end{table}

Qualitatively observing the robot behavior revealed that some causes of failure were the object overshooting the target orientation or the finger slipping across the object, especially for OOD objects. One explanation is that rigid hemispherical fingertips contact the object in a very small area (close to making a point contact), which makes small errors in the action commands more pronounced. Further, we found that the fingertip material had low friction resulting in slips which made manipulation harder. To mitigate these issues, we designed and fabricated soft fingertips that cover the rigid 3D-printed skeleton with a soft elastomer (see Figure S2c in the supplementary material). Soft fingertips provide higher friction and deform when contact happens (compliance), increasing the contact area between the finger and the object. The error distribution in Figure 2C shows using soft fingers doesn't affect performance on train objects but improves generalization to OOD objects. Results in Table 1 confirm the findings -- success rate on OOD objects increases from $45\%$ to $55\%$ when switching from rigid to soft fingertips. Qualitatively, we noticed that soft fingertips behave less aggressively than rigid fingertips resulting in smoother object motion. We, therefore, use soft fingertips in the rest of the experiments. It's worth noting that although the controller was trained using a rigid-body simulator, its performance does not degrade when applied to soft fingertips.

The reorientation error can result from imperfect training, sim-to-real gap, generalization gap, or failures at detecting if the object is at the target orientation, which triggers the controller to stop. In Figure 2D, we report the error distribution in simulation. Although the trained controller is not perfect in simulation, the errors in simulation follow the same trend as in the real world (Figure 2C) but are lower, indicating some sim-to-real gap. As shown in Table 1, the performance gap between the simulation and the real world is smaller with a relaxed error threshold of 0.8 radians than with a threshold of 0.4 radians, illustrating the difficulty in precise reorientation. For some objects ($\#1, \#12$), the error distribution is bi-modal both in simulation and the real world. The test runs with high errors largely result from incorrect detection of when to stop. For instance, object $\#12$ appears nearly symmetric in the point cloud representation, which often leads to errors close to $180^\circ$. Although it is hard to quantitatively disentangle errors originating from incorrect action prediction and the stopping criterion, based on our experience with the system, we hypothesize that the latter contributes more which is supported by the analysis in Supplementary Discussion (see Discussion on precise manipulation).

\subsubsection*{Object reorientation on different supporting materials}
Changing the table surface changes the dynamics of object motion. We tested if our controller is robust to a diverse set of materials: a rough cloth (M1), a smooth cloth (M2), a slippery acrylic sheet (M3), a bathtub mat with perforations resulting in non-stationary object dynamics depending on the object's position on the mat (M4), and a door mat with uneven texture (M5). The materials have different surface structures, roughness, and friction, leading to different system dynamics. We evaluate with one in-distribution object (object $\#5$) and one out-of-distribution object (object $\#10$). Figure 2F and Figure 2G show that our controller performs similarly on different supporting materials, demonstrating its robustness.

\subsection*{Towards object reorientation in air}
As the controllers discussed above were trained with a supporting surface, when the supporting surface was removed, the manipulator consistently dropped the object resulting in failures. Prior work used a specialized training procedure of configuring the object in a good pose at the start of each training episode and a manually designed gravity curriculum \cite{chen2021system} to learn in-air (without supporting surface) reorientation controllers. Consequently, it was necessary to train separate controllers for reorientation with a supporting surface and in the air. It is preferable to have a single controller capable of in-air reorientation and use the supporting surface, if available, to recover from any dropping failures. We achieved this desideratum by employing a four-fingered hand and designing a reward function that penalizes contact between the object and the supporting surface to discourage the controller from using external support for reorientation. When the controller is trained on a supporting surface with the proposed reward function, in-air reorientation emerges.

\begin{figure}[!tb]
    \centering
    \includegraphics[width=0.9\linewidth]{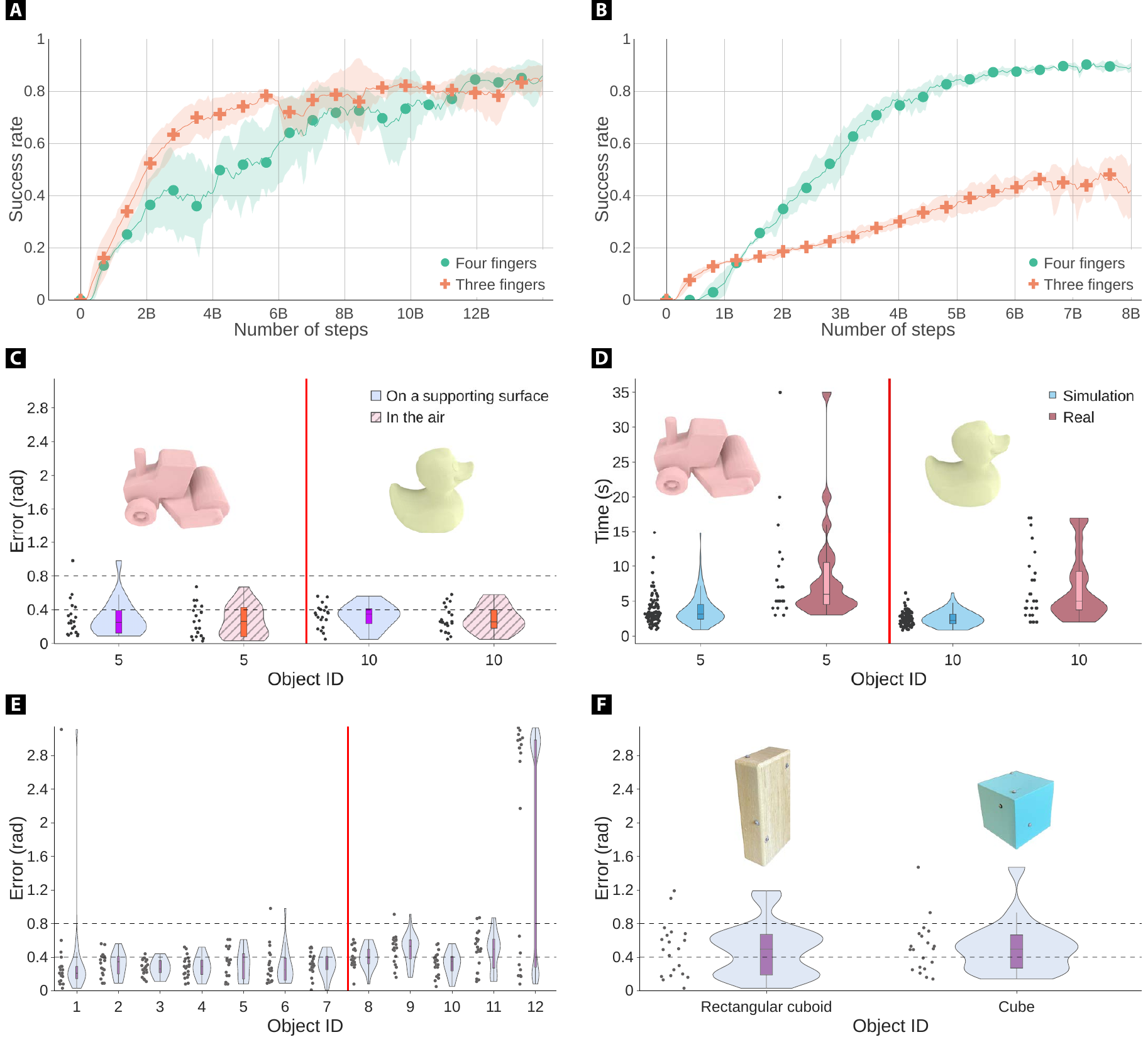}
    \caption{\textbf{Benefit and performance of reorientation with a four-fingered hand}.\textbf{(A)}: When training a controller to reorient objects with a supporting surface, the three-fingered and four-fingered hands achieve similar learning performance. \textbf{(B)}: However, when we incentivize the hands to lift the object during reorientation, the four-fingered hand outperforms the three-fingered hand substantially. \textbf{(C)}: We tested the controller performance with a four-fingered hand in the air. We collected $20$ non-dropping testing cases for one in-distribution object and one out-of-distribution object. The error distribution is similar to that in the case of table-top reorientation. \textbf{(D)} shows the distribution of the episode time both in simulation and the real world. \textbf{(E)}: We show the same controller's performance on twelve objects with a supporting surface. \textbf{(F)}: We tested the controller on symmetric objects with a supporting surface. The controller behaves reasonably well even though it was never trained with symmetric objects. }
    \label{fig:4_finger_ecdf}
\end{figure}

Although both three and four-fingered hands can reorient objects on a supporting surface (Figure 4A), only the four-fingered hand was capable of in-air reorientation (Figure 4B). We hypothesize this to be the case because, with four fingers, more finger configurations can reorient the object, making it easier for policy optimization to find one solution. Furthermore, we hypothesize that the redundancy in the number of fingers makes the system more robust to errors in action prediction. 

\subsubsection*{SO(3) object reorientation in air}
Figure 1B shows how our controller trained in simulation reorients different real-world objects in the air. In-air reorientation can fail if the object is not accurately reoriented or if the robot drops the object. Because in-air reorientation is more challenging, it is possible that the controller is less accurate at reorienting objects. On evaluation with two objects, we found the distribution of orientation error in trials where the objects are not dropped (Figure 4C) to be similar to reorientation with the supporting surface, indicating that the controller doesn't lose reorientation precision in the more challenging in-air scenario. In simulation analysis, we did not notice any notable correlation between orientation error and the distance between the initial and target orientations (Figure S12b in the supplementary material), indicating that the controller performs similarly in the full SO(3) space.

Our controller performs dynamic reorientation. The median time for manipulation across objects and randomly sampled orientation distances in the full SO(3) space is less than 7s (Figure 4D), which makes it a fast in-air reorientation controller operating in the full SO(3) space. Figure 4D also shows that the reorientation times in the real world are longer than in simulation, which we believe is due to real-world contact dynamics being different from simulation.

Simulation analysis reveals that object dropping is the most notable source of errors (Figure S12c). Dropping rates vary substantially across objects. Real-world results follow the same trend. The dropping rate of a shape used in training, the truck (object $\#5$), was $23\%$, much lower than the dropping rate of $56\%$ for an out-of-distribution duck-shaped object ($\#10$). The dropping rate for the duck object shape in the simulation was around $20\%$ showing a sim-to-real gap. However, it remains unclear if the difference in performance can be attributed to the simulator being an approximate model of the real world or whether the object in the real world is much harder to manipulate. This is because, even though the simulation and real-world experiments used the object with the same shape, properties such as surface friction that are critical in reorientation can be different. If an object is curved and has a smooth surface, which is the case with the duck, small differences in friction can substantially change the task difficulty. We chose to report results on the duck as it was used in prior work~\cite{morgan2022complex} and is among the harder objects to reorient and thus also highlights the limitations of our controller.

If a table is present below the hand (for example, the setup shown in the third row of Figure 1B) and the object is dropped, we notice that our controller picks up the object and continues reorienting -- an instance of recovery from failures. It is possible that the reward term encouraging in-air reorientation might hurt on-table reorientation. However, the error distribution for on-table reorientation with the updated reward function (Equation 6)(Figure 4E) is similar to earlier on-table experiments. Moreover, although our controller is trained using objects with asymmetry or reflective symmetry, which makes learning much easier, we noticed some generalization to symmetric objects (Figure 4F, more discussion in Supplementary Discussion). The in-air, on-table, and dropping recovery results demonstrate that it is possible to build a single controller that works across different scenarios. 

Qualitatively looking at the reorientation behavior, it might appear that the object is not always moving toward the target orientation. One possibility is that the manipulator randomly moves the object until it gets close to the target orientation by chance and then stops. To rule out this possibility, we provide videos in Movie S1 showing that for the same initial but different target orientation, the object motions are different. And for the same initial and target orientation, object motions across trials are similar, which would not be the case if the object was randomly being reoriented.

\subsection*{Generalization to objects in daily life}
In previous experiments, we used 3D-printed objects for quantitative evaluation. However, real-world objects have varying object dynamics due to differences in material properties, non-uniform mass distribution, and other factors that can vary across the object surface. To test the generalization ability of our controller on such objects, we conducted a qualitative evaluation on a few household objects. Since we did not have the CAD (Computer Aided Design) model of these objects to generate point clouds in target orientations, we used a free iPad App called Scaniverse to scan the objects. Note that the scan was only required to specify the target orientation, and the scanned object cloud was imperfect (see Figure 5), resulting in noisy goal specification. Figures 1B and 5 illustrate examples of reorienting such objects. The results illustrate that the controller exhibits a certain degree of robustness against noise in the goal specification and some ability to generalize to new materials and shapes.

\begin{figure}[!htb]
    \centering
    \includegraphics[width=\linewidth]{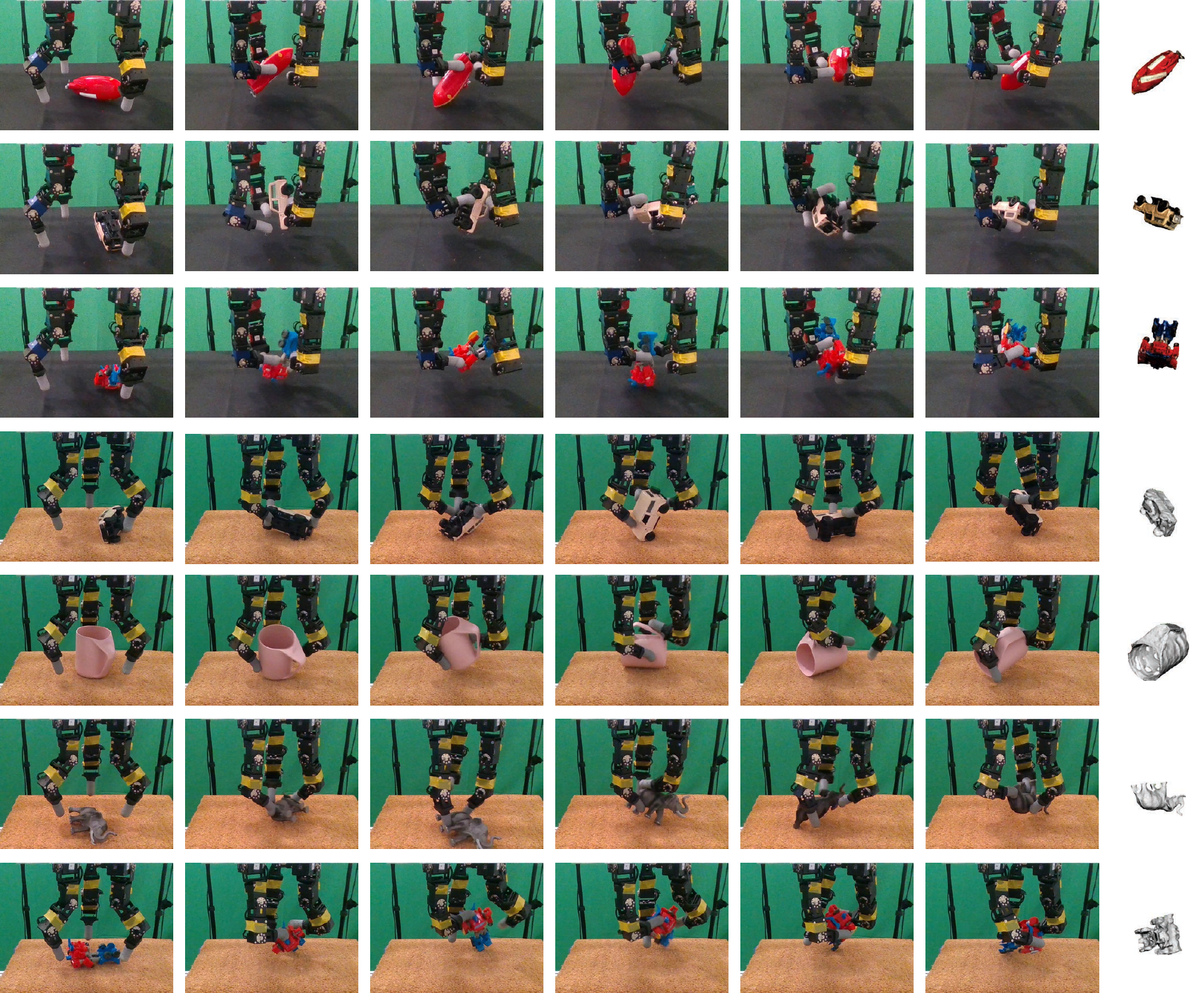}
    \caption{\textbf{Reorientation of real objects}. Examples of reorienting real objects that were not 3D printed using a four-fingered and a three-fingered manipulator.
    }
    \label{fig:real_objects_demo}
\end{figure}

\subsection*{Comparison to prior works}
Unfortunately, a strictly fair comparison with prior work is not possible as we make fewer assumptions (such as no object-specific pose trackers, reorientation in full SO(3) space, and not being quasi-static), and there are substantial differences in hardware/sensing. Nevertheless, to contextualize our research within the existing literature, we present an approximate comparison to the closest work that reported reorientation results on a duck-shaped object with a downward-facing but under-actuated hand of different morphology and mechanical properties~\cite{morgan2022complex}. 
They reported a success rate of $60\%$ (3 out of 5 tests) for reorienting the duck quasi-statically (reorientation time of more than $70s$ compared to $\sim7$s for our controller) to within 0.1 radians, but only in a subset of the SO(3) space (rotation only along two axes). Further, they used a precise object-specific pose tracker (error $<2$ degrees or $0.034$ radians). If we assume perfect stopping criteria (the agent stops reorientation if the object is within $0.1$ radians of the target), then for the duck-shaped object, we achieve a success rate of $71\%$ when dynamically reorienting in the full SO(3) space in simulation. Due to challenges in setting up precise stopping in the real world, we could not run these evaluations in the real world. Even if we did, the differences in material properties between the duck used by us and prior research~\cite{morgan2022complex} would make the comparison unfair. Comparing our simulation and their real-world results is also unfair. However, the results indicate that with more assumptions, such as the precise stopping criterion, the performance of our system improves. Improving the precision of our system without any additional assumptions is an exciting avenue for future research.  

The differences in experimental setups with other prior works~\cite{andrychowicz2020learning,akkaya2019solving,sievers2022learning,abondance2020dexterous} and concurrent work~\cite{handa2023dextreme} are even larger. For instance, OpenAI's work~\cite{andrychowicz2020learning} reported results on reorientation with a single object (no generalization), with a simple shape (cube), an upward-facing hand, and an extensive sensing system consisting of three RGB cameras, a motion capture system, and a different hand. Moreover, their success criterion was the number of times an object passes through a target pose, and they never trained their controller to stop the object at the target pose, which we experimentally found harder to learn. In the broader context of manipulation, the ability to stop at the target pose is vital: If the robot uses a tool, it must reorient it to the desired pose and hold the tool in that pose. 

The focus of our work is not to increase the reorientation performance on a single object; rather, our work expands the scope of object reorientation to operate in more general and pragmatic settings. The result is a single controller for reorienting multiple objects, evidence of some generalization to new objects, and dynamic reorientation in the air without a highly specialized perception system. At the same time, there remains ample scope for improving performance, and we hope that our conscious use of open-source hardware, commodity sensing, computing, and fast-learning framework (Figure 6 and Figure 7) will facilitate future research in enhancing performance and comparing results.

\section*{Discussion}
Solving contact-rich tasks typically requires optimizing the location at which the robotic manipulator contacts the object~\cite{chen2021trajectotree, mordatch2012contact,pang2022global}. One would assume predicting the contact location requires knowledge of the object's shape. However, inputs to the teacher policy have no information about object shape, yet it could reorient diverse and new objects. One possibility is that the agent gathers shape information by integrating information across the sequence of touches made by the fingers. However, the teacher policy is not recurrent, ruling out this possibility. The surprising observation of reorientation without knowledge of shape was made by earlier work in the context of a reorientation system in simulation~\cite{chen2021system}. However, because real-world results were not demonstrated, it remained unclear if such an observation was an artifact of the simulator or the property of the reorientation problem. With real-world evaluation, we have more confidence that shape information may not be as critical to object reorientation as one might apriori think. However, this is not to suggest that shape is not useful at all. The results show that one can go quite far without shape information, but the performance, especially on precise manipulation and in generalization to new shapes, can likely be improved by incorporating shape features into the teacher policy, an exciting direction for future research.   

Typically, having more fingers introduces more optimization variables, making the optimization problem harder in the conventional view. However, we have some evidence to the contrary (Figure 4B). Having more fingers can make it easier for deep reinforcement learning to find a solution, especially in challenging manipulation scenarios such as in the air, similar to how over-parameterized deep networks find better solutions (a conjecture). We conjecture that over-parameterized hardware results in a larger pool of good solutions (more ways to reorient an object with more fingers), making it easier for current optimizers in deep learning to find a good solution.

In designing the proposed system, we either devised or made several technical choices: two-stage student training, representing both the camera recordings and proprioceptive readings as a point cloud, sparse convolution neural network for real-time control, limited range of domain randomization due to system identification, system identification using parallel GPU simulation, use of soft material on fingertips, using a larger number of fingers instead of the conventional wisdom of using fewer fingers. These choices, however, are not specific to in-hand reorientation but can be applied to a broad spectrum of vision-based manipulation tasks involving rigid bodies. We hope that the knowledge of these choices, along with a low-cost platform, can further the goal of democratizing research in dexterous manipulation. 

\paragraph{Limitations and Possible Extensions}
Object reorientation with a downward-facing hand has notable room for improving precision and reducing the drop rate. We hypothesize that one possible cause for dropping objects is that the control frequency of $12$Hz is not fast enough. The robot dynamically manipulates the object, and it takes a fraction of a second to lose control. It might be challenging to determine when the object is slipping from the fingers in real-time using visual feedback at 12Hz. Feedback control at a higher frequency may mitigate such failures but either requires more efficient neural network architectures or more processing power. 

Another hypothesis for object dropping is missing information regarding whether the finger is in contact with the object, if the object is slipping, or how much force is being applied. We conjecture that explicit knowledge of contact, contact force, and other signals such as slip can substantially improve performance. Currently, the robot relies purely on occluded vision observations to infer contacts. Augmenting the robot's observation with touch sensors is therefore an exciting direction for future investigation.  

We also found that inaccurate prediction of rotational distance is another cause for imprecise object reorientation. The prediction of rotational distance is less accurate when the actual rotational distance is less than 0.4 radians (see Discussion on precise manipulation in Supplementary Discussion).

We hypothesize that generalization and precision can be improved by training on a larger object dataset, investigating RGB sensing to complement depth sensing to capture fine geometric structures and reduce noise, and integrating visual and tactile sensing to obtain more complete point clouds. Further, there remains a sim-to-real gap that future research should investigate. 

We used \dclaw manipulators in this work as it is open-source and low-cost. However, many aspects of the D'Claw, such as the finger design and the number of fingers, are sub-optimal. For instance, although we observed some robustness to the softness of fingertips, different softness and skeleton designs can notably affect the longevity of fingertips. We manually iterated over many soft fingertip designs, which was time-consuming. Similarly, the fingertips have a hemispherical shape, quite different from humans and presumably not optimal. The performance of the task can be improved by better hardware design: the shape of fingers, the degrees of actuation on each finger, the placement of fingers, and the choice of materials. Manually iterating over these choices is infeasible. A promising future direction is to utilize a computational approach for automatically designing the hand for specific tasks~\cite{xu2021end}.

In summary, we presented a real-time controller that can dynamically reorient complex and new objects by any desired amount using a single depth camera. The system is both simple and affordable, which aligns with the objective of making dexterous manipulation research accessible to a wider audience.

\section*{Materials and Method}
Given a random object in a random initial pose, the robot is tasked to reorient the object to a user-provided target orientation in SO(3) space. We train a single vision-based object reorientation controller (or policy) in simulation to reorient hundreds of objects. The controller trained in simulation is directly deployed in the real world (zero-shot transfer). The choices in our experimental setup have been made to support future deployment of reorientation in service of tool use and on a mobile manipulator. 

\paragraph{Object datasets} We use two object datasets in this work: \textbf{Big dataset} ($\bigdataset$) and \textbf{Small dataset} ($\smalldataset$). $\bigdataset$ contains $150$ objects from internet sources. $\smalldataset$ contains $12$ objects from the ContactDB \cite{Brahmbhatt_2019_CVPR} dataset. These two datasets do not have overlapped shapes. More details on the object dataset are in Supplementary Methods.

\paragraph{Simulation setup} We use Isaac Gym \cite{makoviychuk2021isaac} as the rigid body physics simulator. We train all the policies on a table-top setup: hands face downward with a supporting table.

\paragraph{Success criteria}
During training, the success criterion for reorienting an object acts as both a reward signal and a criterion for success to end the episode. A straightforward success criterion is judging whether an object's orientation is close to the target orientation (orientation criterion). However, a controller trained using this criterion tends to cause the object to oscillate around the target orientation. To address this issue, the success criterion is expanded to explicitly penalize finger and object movements. For further details on how we designed the success criteria for training, please refer to Supplementary Methods.

\subsection*{Training the visuomotor policy}
We model the problem of learning the controller, $\pi$, as a finite-horizon discrete-time decision process with horizon length $T$. The policy $\pi$ takes as input sensory observations ($\obs$) and outputs action commands ($\act$) at every time step $t$. Learning $\pi$ using RL is data inefficient when the observation ($\obs$) is high-dimensional (for example, point clouds). The reason is that the policy needs to simultaneously learn which features to extract from visual observations and what are the high-rewarding actions. The problem would be simplified if one of these factors were known: learning a policy via RL from sufficient state information would be much easier than direct learning from sensory observations. Similarly, apriori knowledge of high-rewarding actions would reduce the data requirements of learning from visual observations.  

Prior work has employed this intuition to ease policy learning by decomposing the learning process into two steps~\cite{chen2021system,margolis2021learning,chen2020learning,lee2020learning}. In the first step, a teacher policy is trained in simulation with RL using low-dimensional state space that includes privileged information. In the case of in-hand object reorientation, privileged information includes quantities such as fingertip velocity, object pose, and object velocity that can be directly accessed from the simulator but can be challenging to measure in the real world. Because the teacher policy operates from a low-dimensional state space, it can be more efficiently trained using RL. Next, to enable operation in the real world, one can either train a perception system to predict the privileged information~\cite{andrychowicz2020learning,allshire2022transferring} or train a second student policy to predict high-rewarding teacher actions from raw sensory observations via supervised learning~\cite{chen2020learning,lee2020learning,chen2021system,margolis2021learning}. 

An underlying assumption of the two-stage training paradigm is that a low-dimensional state for learning a teacher policy can be identified. Because there are no tools available to theoretically analyze if a particular choice of state space is sufficient for policy learning, selecting the state inputs for the teacher policy is a manual process based on human intuition. At first, object reorientation might seem to require knowledge of object shape since the controller must reason about where to make contact. If object shape is necessary, then it will not be possible to reduce depth observations into a low-dimensional state. However, past work found that even without any shape information, it is possible to train RL policies to achieve good reorientation performance on a diverse set of objects in simulation~\cite{chen2021system}. Therefore, teacher-student training can be leveraged to simplify the learning of object reorientation.

To deploy the policy in the real world, some prior works train a perception system to predict the object pose~\cite{akkaya2019solving,andrychowicz2020learning}. However, object pose is only defined with respect to a particular reference frame. Choosing a common frame of reference across different objects is not possible. As a consequence, pose estimators cannot generalize across objects. Therefore, we choose to train an end-to-end student policy that takes as input the raw sensory observations and is optimized to match the actions predicted by the teacher policy via supervised learning~\cite{ross2011reduction}. Because supervised learning is considerably more data efficient than RL, such an approach solves the hard problem of learning a policy from raw sensory observations.

The teacher-student training paradigm has been used to learn object reorientation policy in simulation from visual and proprioceptive observations~\cite{chen2021system}. However, a separate policy was trained per object. Secondly, it required more than a week to train the student vision policy for a single object on an NVIDIA V100 GPU. We developed a two-stage student training (Teacher-student$^2$) framework (Figure 6) that substantially speeds up the vision student policy learning. Using this framework, we were able to learn a vision policy that operates across a diverse set of objects and generalizes to objects with different shapes and physical parameters. 

\begin{figure}[!htb]
    \centering
    \includegraphics[width=0.82\linewidth]{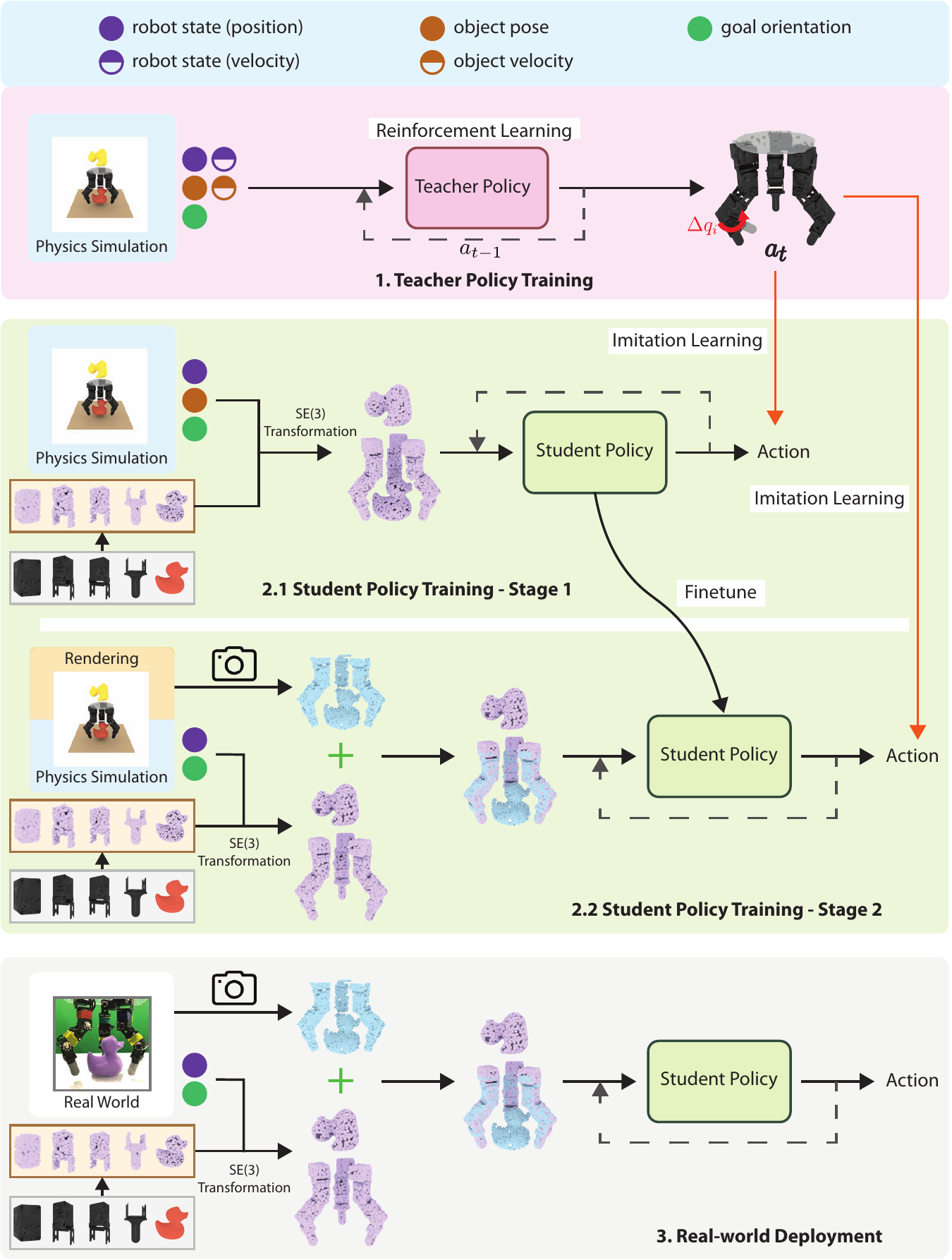}
    \caption{\textbf{Teacher and two-stage student training framework}. First, a teacher policy is trained using reinforcement learning with privileged state information. Then, a student policy is trained to imitate the teacher using synthetic and complete point clouds as input. The student policy is further fine-tuned using rendered point clouds. During deployment, the student policy can be directly used to control real robots.}
    \label{fig:ts2}
\end{figure}

\subsubsection*{Teacher policy: reinforcement learning with privileged information}
\label{subsec:teacher-policy}
The learning of teacher policy ($\expert$) is formulated as a reinforcement learning problem where the robot observes the current observation ($\bm{o}_t^\expsymbol$), takes an action ($\bm{a}_t$), and receives a reward ($r_t$) afterward. A single policy ($\expert$) is trained across multiple objects using proximal policy optimization (PPO)~\cite{schulman2017proximal} to maximize the expected discounted episodic return: $\pi^{\expsymbol^*}=\arg\max_{\expert}\mathbb{E}\left[\sum_{t=0}^{T-1}\gamma^tr_t\right]$. Since the observation $\bm{o}_t$ at a single time step $t$ does not convey the full state information such as the geometric shape of an object, our setup is an instance of Partially Observable Markov Decision Process (POMDP). However, for the sake of simplicity and based on the finding that knowledge of object shape may not be critical as discussed above, we chose to model the policy as a Markov Decision Process (MDP): $\act = \pi^{\expsymbol^*}(\obs; \bm{a}_{t-1})$. The policy also takes as input the previous action ($\bm{a}_{t-1}$) to encourage smooth control.

\paragraph{Observation space} The inputs to the teacher policy, $\bm{o}_t$, include proprioceptive state information, object state, and target orientation. Details are shown in Supplementary Methods.

\paragraph{Action space} We use position controllers to actuate the robot joints at a frequency of $12$Hz. The policy outputs the relative joint position changes $\bm{a}_t\in\mathbb{R}^{3G}$. Instead of directly using $\act$, we use the exponential moving average of actions $\bar{\bm{a}}_t=\alpha \bm{a}_t + (1 - \alpha) \bar{\bm{a}}_{t-1}$ for smooth control, where  $\alpha\in[0, 1]$ is a smoothing coefficient. 
In our experiments, we set $\alpha=0.8$. Given the smoothed action $\bar{\bm{a}}_t$, the target joint position at the next time step is: $\bm{q}^{tgt}_{t+1}=\bm{q}_t + \bar{\bm{a}}_t$. 

\paragraph{Reward} We first describe the reward function for the hand to reorient objects on a table. The first term in the reward function (Equation 1) is the success criteria for the task. However, since this only provides sparse reward supervision, the criteria by itself is insufficient for successful learning. Therefore we add additional reward shaping \cite{ng1999policy} terms to encourage reorientation. We use a dense reward term that encourages minimization of the distance ($\Delta \theta_t$) between the agent's current and target orientation (Equation 2). We penalize the agent for moving fingertips far away from the object (Equation 3). Without this term, fingers barely made any contact with the object during training. We also penalize the agent for expending energy (Equation 4) and for pushing the object too far from the robot's hand (Equation 5) in which case the episode is also terminated. The reward terms are mathematically expressed as:   

\begin{align}
\label{eqn:rot_reward_sparse}
    r_{1t} =& c_1 \mathds{1}(\text{Task successful}) &\text{sparse task reward} \\
    \label{eqn:rot_reward_dense}+& c_2\frac{1}{|\Delta \theta_t|+\epsilon_\theta} &\text{dense task reward}
    \\ \label{eqn:rot_reward_fingertip}+&c_3\sum_{i=1}^G\left\Vert \bm{p}_t^{f_i}-\bm{p}_t^o\right\Vert_2^2 &\text{keep fingertip close to the object}
    \\ \label{eqn:rot_reward_energy} +& c_4|\dot{\bm{q}}_t|^T|\bm{\tau}_t|  &\text{energy reward}
    \\ \label{eqn:rot_reward_reset} +& c_5 \mathds{1}(\left\Vert\bm{p}_t^o\right\Vert_2^2 > \Bar{p})  & \text{penalty for pushing the object away}
\end{align}

where $c_1, c_2>0$, and $c_3, c_4, c_5 <0$ are coefficients, $\mathds{1}$ is an indicator function, $\epsilon_\theta$ and $\Bar{p}$ are constants, $\bm{p}_t^{f_i}$ is the fingertip position of $i^{th}$ finger, $\bm{p}_t^o$ is the object center position, $\bm{\tau}_t$ is the vector of the joint torques. 

Using the aforementioned reward function, we were able to train reorientation policies that used the support of the table. Next, to enable the more challenging behavior of reorienting objects in the air, we added a penalty for the contact between the object and table (Equation 7) and a penalty for using the penultimate joint instead of the fingertip for reorientation (Equation 8). Although the term in Equation 8 is not critical, it results in more natural-looking behaviors. The overall reward function is:   

\begin{align}
\label{eqn:rot_reward_air}
    r_{2t} = & r_{1t} \\
     + \label{eqn:reward_table_penalty} &c_6\mathds{1}(\text{object contacts with the table}) \\
     + \label{eqn:reward_finger_use} &c_7\sum_{i=1}^{N}\mathds{1}(p_{t,z}^{f_i}>\bar{p}_z)
\end{align}
where $c_6, c_7 <0$ are coefficients.

\subsubsection*{Student policy - imitation learning from depth observations}
The student policy ($\student$) is trained in simulation with the purpose of being deployed in the real world. Since the sim-to-real gap for depth data is less pronounced than RGB data, we only use the depth images provided by the camera along with readings from joint encoders. We represent the depth data as a point cloud in the robot's base link frame. To enable the neural network representing $\student$ to model the spatial relationship between the fingers and the object, we express the robot's current configuration by showing the policy a point cloud representing points sampled on the surface of the fingers. We concatenate the point cloud obtained from the camera along with the generated point cloud of the hand. We denote this scene point cloud as $\bm{P}^s_t$.

\begin{figure}[!htb]
    \centering
    \includegraphics[width=0.9\linewidth]{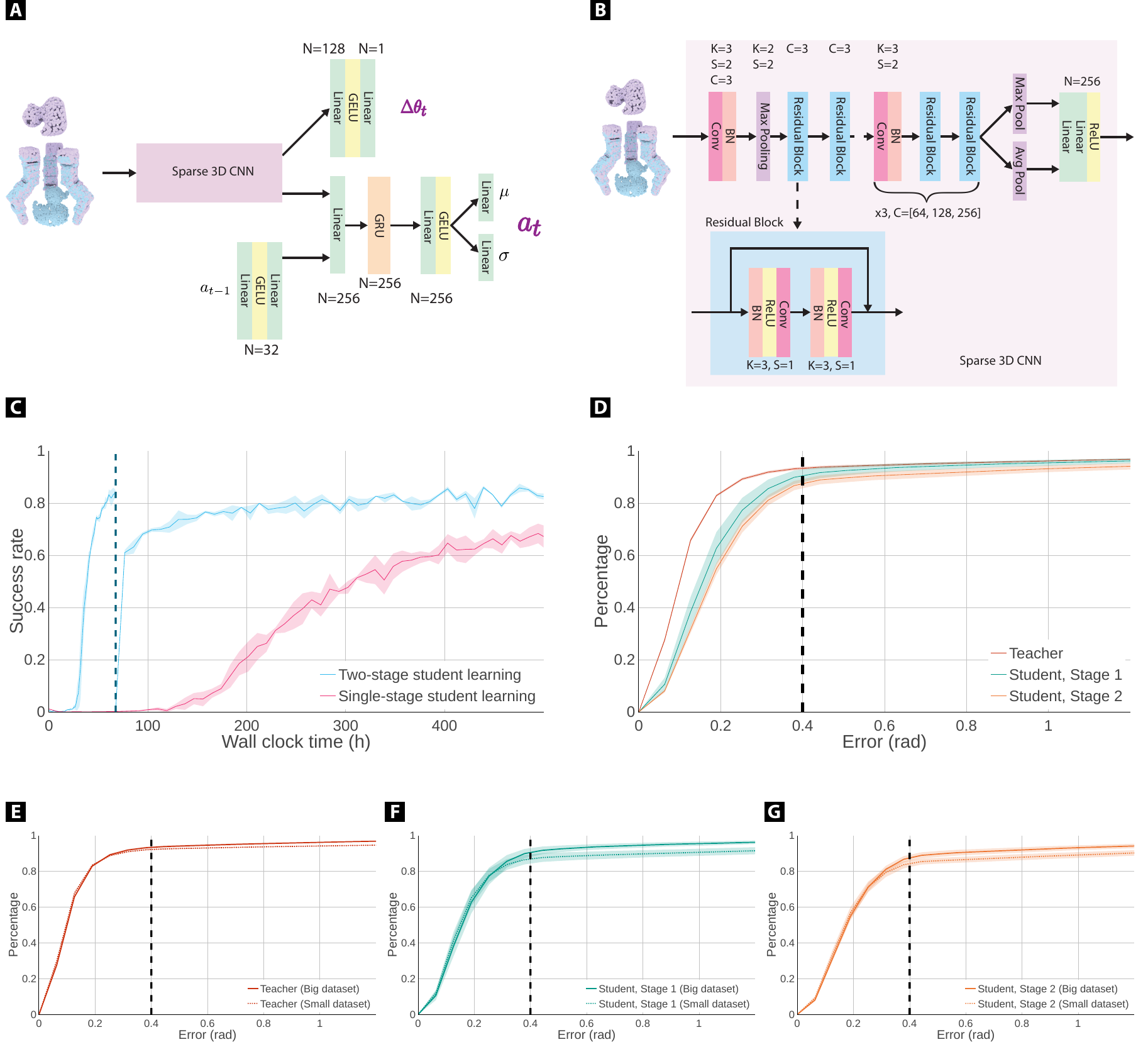}
    \caption{\textbf{Student policy learning}. \textbf{(A)}: Student vision policy network architecture. \textbf{(B)}: Sparse 3D CNN (Convolutional Neural Network) component of the policy network. \textbf{(C)}: Proposed two-stage student learning learns faster than single-stage student learning. The dashed vertical line denotes the transition from the first to the second stage of student learning. The performance dip happens due to a change in the distribution of point cloud inputs from being unoccluded in the first stage to being occluded in the second. \textbf{(D)}: Post-training evaluation of teacher and student policies on the training dataset $\bigdataset$. For each object, the initial and target orientations are randomly sampled $50$ times, resulting in $7500$ samples. The empirical cumulative distribution function (ECDF) of the orientation error is plotted. The results show that the students are close to the teacher's performance. \textbf{(E)}, \textbf{(F)}, \textbf{(G)}: Comparing the ECDFs of the policies being evaluated on dataset $\bigdataset$ and dataset $\smalldataset$ reveals small generalization gap for all the policies.
    }
    \label{fig:ts2_curves}
\end{figure}

\paragraph{Goal representation} 

Instead of providing the goal orientation as a pose which has generalization issues discussed above, the goal is represented as the object's point cloud in the target orientation $\bm{P}^g$. In other words, the policy sees how the object should look in the end (see the top left of Figure 7A).

\paragraph{Observation space} The input to $\student$ is the point cloud $\bm{P}_t = \bm{P}^s_t\cup \bm{P}^g$ (see Figure 7A). We also did an ablation study on different ways to process the goal point cloud in Supplementary Discussion S5.4. The results show that merging $\bm{P}^s_t$ and $\bm{P}^g$ before they are input to the network leads to faster learning.

\paragraph{Architecture} The critical requirement for the vision policy is to run at a high enough frequency to enable real-time control. For fast computation, we designed a sparse convolutional neural network to process point cloud ($\bm{P}_t$) using the Minkowski Engine \cite{choy20194d} (see Figure 7A). Compared to the architecture used in \cite{chen2021system}, our convolutional network has a higher capacity to make it possible to learn the reorientation of multiple objects. Without direct access to object velocity, it is necessary to integrate temporal information in $\student$, for which we use the gated recurrent unit~\cite{cho2014learning} in the network.

\paragraph{Optimization} The student policy $\student$ is trained using DAGGER~\cite{ross2011reduction} to imitate the teacher policy $\expert$. 

\paragraph{Need for two-stage student learning} We found training a vision policy in simulation to be slow, consuming 20+ days on an NVIDIA V100 GPU (Figure 7C). The main reason for slow training is that the simulator performs rendering to generate a point cloud which consumes a substantial amount of time and GPU memory. To reduce training time, we generated synthetic point clouds by uniformly sampling points on the object and robot meshes used by the simulator. The synthetic point cloud is also complete (no occlusions), which makes training easier. The vision policy ($\student_1$) can be trained with synthetic point cloud in less than three days, which is a $7\times$ speedup (\textbf{stage 1}; see Figure 7C). However, the policy, $\student_1$, cannot be deployed in the real world because it operates on an idealized point cloud (no occlusions). Therefore, once the student reaches high performance, we initiate \textbf{stage 2}, where the policy is finetuned with the rendered point cloud. Such finetuning is quick in wall-clock time (around one day), and the resulting policy ($\student_2$ ) performs better than training from scratch with rendered point clouds (see Figure 7C). It is possible to further reduce the training time of the student policy by employing visual pre-training with passive data that we discuss in Supplementary Discussion S5.5. An additional benefit of the two-stage student policy training is that $\student_1$ is agnostic to the camera pose. Therefore a policy from a new viewpoint ($\student_2$) can be quickly obtained by finetuning using rendered point clouds from that camera pose. Training the vision policy from scratch is not necessary. 

\paragraph{Stage 1: details of synthetic point cloud} In stage 1, the simulation is not used for rendering but only for physics simulation. We generate the point cloud for each link on the manipulator and object by sampling $K$ points on their meshes in the following way: let the point cloud of link $l_j$ in the local coordinate frame of the link be denoted as $\bm{P}^{l_j}\in\mathbb{R}^{K\times 3}$.  Given link orientation ($\bm{R}_t^{l_j}\in\mathbb{R}^{3\times 3})$ and position ($\bm{p}_t^{l_j}\in\mathbb{R}^{3\times 1}$) at time step $t$, the point cloud can be computed in the global frame, $\bm{P}_t^{l_j} = \bm{P}^{l_j}(\bm{R}_t^{l_j})^T + (\bm{p}_t^{l_j})^T$. The point cloud representation of the entire scene is the union of point clouds of all the links, the object being manipulated, and the object in the goal orientation: $\bm{P}_t^s = \bigcup_{j=1}^{j=M} \bm{P}_t^{l_j}$ where $M$ is the total number of links (bodies) in the environment. The point cloud $\bm{P}_t^s$ can be efficiently generated using matrix multiplication. 

\paragraph{Stage 2: details of rendered point cloud}
In stage 2, at each time step, we acquire depth images from the simulator and convert them into point clouds (which we call exteroceptive point cloud) using the camera's intrinsic and extrinsic matrices. Note that such a point cloud is incomplete due to occlusions. We also convert the joint angle information into poses of the links on the robot hand via forward kinematics and then generate the complete point cloud of the robot (which we call proprioceptive point cloud). Note that such a proprioceptive point cloud of a robot can be easily obtained in the real world in real-time from the joint position readings. The policy input is the union of the exteroceptive and the proprioceptive point cloud.

\subsection*{Reducing the simulation to reality gap}
There are two main sources of the gap between simulation and reality. The first one is dynamics gap that arises from differences in the robot dynamics, the approximation in the simulator's contact model, and differences in object dynamics that depends on material properties such as friction. The other source is perception gap caused by differences in statistics of sensor readings and/or noise. One way to reduce these gaps is to train a single policy across many different settings of the simulation parameters (domain randomization~\cite{tobin2017domain}). The success of domain randomization hinges on the hope that the real world is well approximated by one of the many simulation parameter settings used during training. The chances of such a match increase by randomizing parameters over a larger range. However, excessive randomization may result in an overly conservative policy with low performance~\cite{tan2018sim}. Therefore, we make design choices that reduce the need for domain randomization and use it only when needed.   

The perception gap is reduced by using only depth readings, which is more similar between simulation and reality than RGB. To account for noisy depth sensing, we add noise to the simulated point cloud. The dynamics gap can be reduced by identifying simulation parameters closest to the real world. While such identification is possible for the robotic manipulator, it is infeasible for object dynamics that vary in material and mass distribution. Therefore, we perform system identification on the robot dynamics and use only small randomization to account for unmodeled errors. We use a larger range of domain randomization on the object and environment dynamics. To make the policy more robust to unmodeled real-world physics, we apply random forces on the object during training which pressures the policy to reorient objects while being robust to external disturbance. Lastly, to increase compliance and friction between the object and the manipulator, we use soft fingertips. Such a choice makes the system more tolerant of errors in control commands. Empirically we noticed that soft fingertips make the robot less aggressive and reduce overshoot. 

\subsubsection*{Identification of robot dynamics}
We build the Unified Robot Description Format (URDF) model for the manipulator using its CAD model, which provides accurate kinematics parameters, but the dynamics parameters, such as joint damping and stiffness, must be estimated. One way of identifying dynamics parameters is to leverage the equations of motion (or the dynamics model) and solve for the unknown variables using a dataset of motion trajectories. The Isaac Gym simulator has a built-in dynamics model. But because the simulator's code is not open-source, we do not have access to the precise dynamics model nor the gradients of dynamics parameters. We, therefore, used a black-box approach that leverages the ability of Isaac Gym to perform massively parallel simulations. We spawn many simulations with different dynamics parameters and use the one that has the closest match to the real robot's motion. 

Let $\bm{\lambda}_i \in \bm{\Lambda}$ denote the dynamics parameter of the $i^{th}$ simulated robot ($C^{\bm{\lambda}_i}$), where $\bm{\Lambda}$ denotes the entire set of dynamics parameter values over which search is performed. To evaluate the similarity between the motion of $C^{\bm{\lambda}_i}$ and the real robot ($C^{real}$), we compute the score: $h(\bm{q}_A^{C^{real}}(\cdot), \bm{q}_A^{C^{\bm{\lambda}_i}}(\cdot)) = -\left\Vert\bm{q}_A^{C^{real}}(\cdot) - \bm{q}_A^{C^{\bm{\lambda}_i}}(\cdot)\right\Vert_2^2$, where $\bm{q}_A^{C}(\cdot)$ represents the joint position trajectories of a robot $C$ given action commands $\bm{A}(\cdot)$ which are detailed in Supplementary Methods. The closer the motion of the simulated robot is to that of the real world, the higher the score will be. We use the black-box optimization method of Covariance Matrix Adaptation Evolution Strategy (CMA-ES)~\cite{hansen2003reducing}, an instance of evolutionary search algorithms, to determine the optimal dynamics parameter:  $\bm{\lambda}^*=\arg\max_{\bm{\lambda} \in \bm{\Lambda}} h(\bm{q}_A^{C^{real}}(\cdot), \bm{q}_A^{C^{\bm{\lambda}}}(\cdot))$. Note that it might be impossible to find a simulated robot that exactly matches the real robot due to the approximate parameterization of real-world dynamics in simulation and the stochasticity in the real-world resulting from actuation/sensing noise.  More details on the identification are in Supplementary Methods.

\subsection*{Real-world deployment}
\paragraph{Real-world observation} It includes the joint positions of each motor in the manipulator and the depth image from a RealSense camera. Details how the joint positions and depth image are converted into a unified point cloud input can be found in Supplementary Methods.

\paragraph{Stopping criteria} To automatically stop the robot, we train a predictor that re-uses features from the policy network to predict $|\Delta \theta_t|$ (see Figure 7A). The robot is stopped when $\Delta\theta_t^{pred}<\bar{\theta}$ and $||\act||<\bar{a}$. 

More details on the stopping criteria, the real-world experimental setup, and the procedure for quantitative evaluation are in Supplementary Methods.

\clearpage

\section*{List of Supplementary Materials}

The supplementary PDF file includes:

\indent Supplementary Methods\\
\indent Supplementary Discussion\\
\indent Figs. S1 to S14\\
\indent Tables S1 to S3

\noindent Other Supplementary Materials for this manuscript include the following:

\indent Movies S1 to S2

\clearpage

\section*{Acknowledgments}
We thank the members of the Improbable AI lab for the helpful discussions and feedback on the paper. We are grateful to MIT Supercloud and the Lincoln Laboratory Supercomputing Center for providing HPC resources. \textbf{Funding}: Toyota Research Institute, DARPA Machine Common Sense, MIT-IBM Watson AI Lab, and MIT-Airforce AI Accelerator provided funds to Improbable AI lab to support this work. M.T. was supported by the National Science Foundation Graduate Research Fellowship. The views and conclusions contained in this document are those of the authors and should not be interpreted as representing the official policies, either expressed or implied, of the DARPA or the United States Air Force, or the U.S. Government. The U.S. Government is authorized to reproduce and distribute reprints for Government purposes notwithstanding any copyright notation herein. \textbf{Author contributions}: T.C. and P.A. jointly conceived the project. T.C. formulated the main idea of the training and control methods, set up the simulation and training code, trained the controllers, designed and built the real-world hardware platforms, developed the software for controlling the real hands, designed and conducted experiments in simulation and in the real world, and led the manuscript writing. M.T. designed and fabricated the fingertips of the robot hand, a majority of objects used for real-world evaluation and built the four-finger hand. S.W. experimented with vision network pre-training (Stage 0 in the paper) and performed ablations quantifying the effect of each pre-training task. V.K. provided feedback on the manuscript and on experimental results. E.A. provided advice and support for hardware fabrication. P.A. was responsible for overall project supervision, contributed to research discussions, provided advice on experimental design and setup, and played a substantial role in manuscript writing. \textbf{Competing interests}: The authors declare that they have no competing interests. \textbf{Advisory Affiliations.} P.A. is an advisor to Tutor Intelligence Inc., Common Sense Machines Inc., and Lab0 Inc. \textbf{Data and materials availability}: All (other) data needed to evaluate the conclusions in the paper are present in the paper or the Supplementary Materials. The code is available at \url{https://zenodo.org/records/10039109} and \url{https://github.com/Improbable-AI/dexenv}. \textbf{Patents:} A provisional patent application is filed covering some aspects of work.

\newpage

\begin{appendices}
\clearpage
\pagenumbering{gobble}
\setcounter{figure}{0}
\setcounter{table}{0}
\setcounter{footnote}{0}
\renewcommand{\thesection}{S\arabic{section}}
\renewcommand{\thefigure}{S\arabic{figure}}
\renewcommand{\thetable}{S\arabic{table}}
\renewcommand{\thesubfigure}{\Alph{subfigure}}
\newpage
{
\centering

{\Huge\bfseries Visual Dexterity: In-Hand Reorientation of Novel and Complex Object Shapes}

\author
{Tao Chen$^{1,2}$, Megha Tippur$^{2}$, Siyang Wu$^{3}$, Vikash Kumar$^{4}$, \\ Edward Adelson$^{2}$, Pulkit Agrawal$^{\ast 1,2,5}$\\
\ \\
\normalsize{$^{1}$Improbable AI Laboratory, Massachusetts Institute of Technology}\\
\normalsize{Cambridge, MA 02139, USA}\\
\normalsize{$^{2}$Computer Science and Artificial Intelligence Laboratory (CSAIL)},\\ \normalsize{Massachusetts Institute of Technology,}\\
\normalsize{Cambridge, MA 02139, USA}\\
\normalsize{$^{3}$Institute for Interdisciplinary Information Sciences, }\\
\normalsize{Tsinghua University, Beijing, 100084, China} \\
\normalsize{$^{4}$Meta AI, Pittsburgh, PA 15213, USA} \\
\normalsize{$^{5}$Institute of Artificial Intelligence and Advanced Interactions (IAIFI)}\\
\normalsize{Massachusetts Institute of Technology,}\\
\normalsize{Cambridge, MA 02139, USA}\\
\normalsize{$^\ast$To whom correspondence should be addressed; E-mail: pulkitag@mit.edu.}
}

}
\newpage

\section*{Supplementary Methods}

\subsection*{Nomenclature}

\begin{itemize}[noitemsep]
  \item[$\mathbb{B}$] \hspace{1cm} the Big dataset
  \item[$\mathbb{S}$] \hspace{1cm} the Small dataset 
  \item[$\bm{o}$] \hspace{1cm} observation      \item[$\bm{a}$] \hspace{1cm} action command
  \item[$\bar{\bm{a}}$] \hspace{1cm} smoothed action command 
  \item[$\pi$] \hspace{1cm} policy
  \item[$r$] \hspace{1cm} reward
  \item[$\expsymbol$] \hspace{1cm} expert
  \item[$\stusymbol$] \hspace{1cm} student
  \item[$\gamma$] \hspace{1cm} discount factor
  \item[$\alpha$] \hspace{1cm} smoothing factor for action
  \item[$\bm{q}$] \hspace{1cm} joint positions
  \item[$\Delta\theta$] \hspace{1cm} the distance between the object's current orientation and goal orientation
  \item[$\dot{\bm{q}}$] \hspace{1cm}   joint velocities
  \item[$\bm{v}^o$] \hspace{1cm}  object's linear velocity
  \item[$\bm{\omega}^o$] \hspace{1cm} object's angular velocity
  \item[$\bm{z}$] \hspace{1cm}  embedding vector 
  \item[$\bm{A}(\cdot)$] \hspace{1cm} a sequence of action commands 
  \item[$\bm{\Lambda}$] \hspace{1cm} the entire space of possible dynamics parameters  
\item[$\bm{\lambda}$] \hspace{1cm} dynamics parameters of a robot
\item[$f$] \hspace{1cm} frequency  
\item[$F^o_d$] \hspace{1cm} disturbance force on the object  
\item[$m^o$] \hspace{1cm} object's mass  
\item[$\Bar{\theta}$] \hspace{1cm}threshold value for orientation distance 
 \item[$\bar{\dot{q}}$] \hspace{1cm} threshold value for joint velocity norm
 \item[$\bar{a}$] \hspace{1cm} threshold value for action norm  
 \item[$\bar{v}$] \hspace{1cm} threshold value for linear velocity norm
 \item[$\bar{\omega}$] \hspace{1cm} threshold value for angular velocity norm 
 \item[$\Bar{p}$] \hspace{1cm} threshold value for object's distance from the hand
 \item[$\epsilon_\theta$] \hspace{1cm} a constant in reward function  
 \item[$\bm{P}$] \hspace{1cm} point cloud 
  \item[$l_j$] \hspace{1cm} $j^{th}$link on the hand 
  \item[$G$] \hspace{1cm} number of fingers 
  \item[$\bm{p}^{f_i}$] \hspace{1cm} fingertip position of $i^{th}$ finger 
  \item[$\bm{\tau}_t$] \hspace{1cm} joint torques
  \item[$\bm{p}^o$] \hspace{1cm} object center position
  \item[$\bm{R}$] \hspace{1cm} rotation matrix 
  \item[$\bm{p}$] \hspace{1cm} position
  \item[$M$] \hspace{1cm} number of links on a hand
  \item[$C$] \hspace{1cm} robot
  \item[$h$] \hspace{1cm} score function for trajectory similarity
  \item[$\gaussian()$] \hspace{1cm} Gaussian distribution
  \item[$\uniform()$] \hspace{1cm} uniform distribution
\end{itemize}

\subsection*{Experiment details}

\subsubsection*{Object datasets} 
\label{app_subsec:object_dataset}

\begin{figure}[!htb]
\centering
\includegraphics[width=0.98\linewidth]{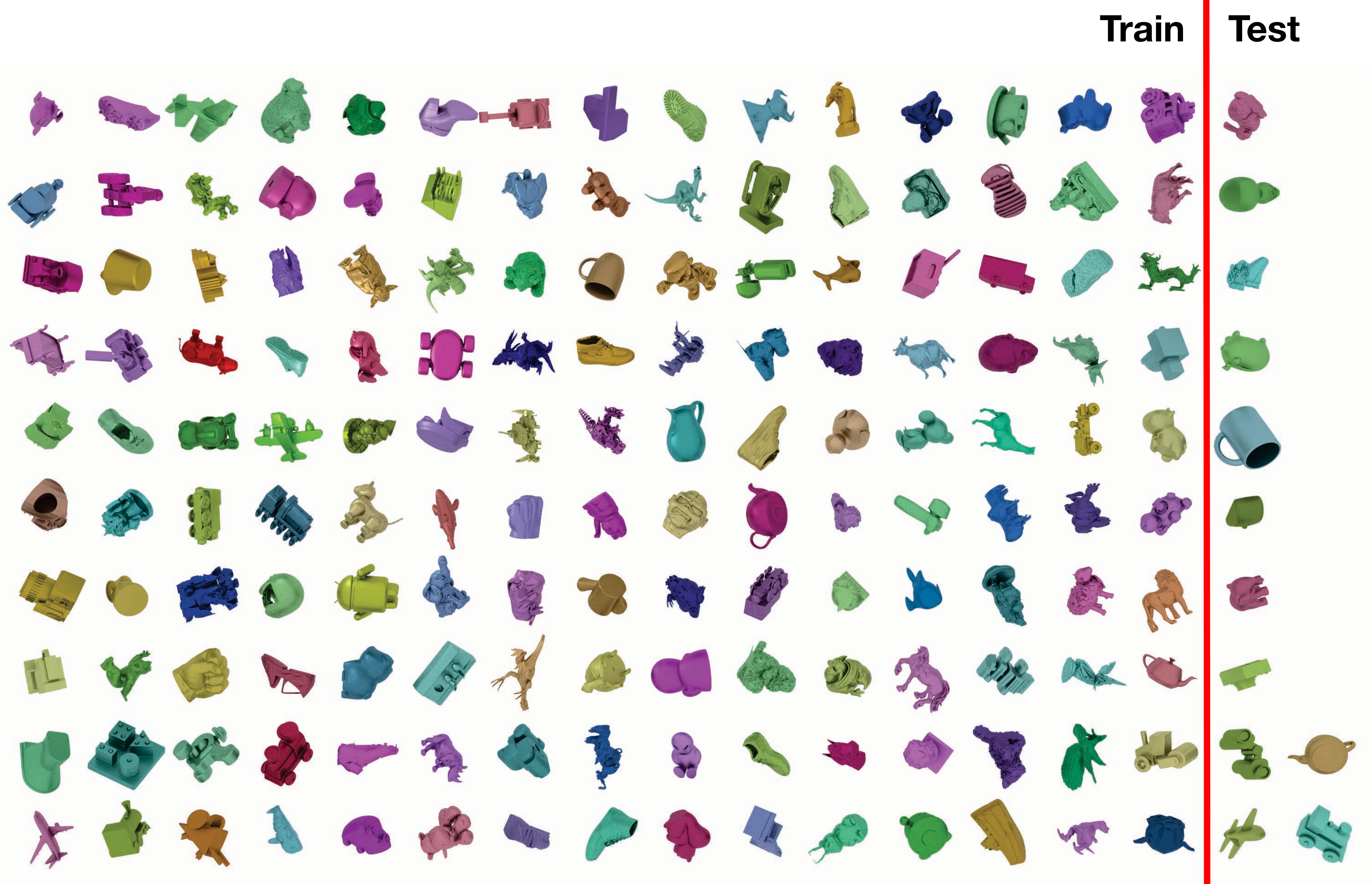}
\caption{\textbf{Object dataset}. On the left of the red line, we show the dataset $\bigdataset$ (the training dataset). And on the right of the red line, we show the dataset $\smalldataset$ (the testing dataset in simulation).
}
\label{fig:merged_dataset}
\end{figure}

We use the following object sets: \textbf{Big dataset} ($\bigdataset$) and \textbf{Small dataset} ($\smalldataset$). $\bigdataset$ is used for training policies. It is a collection of $150$ objects from internet sources such as Google Scanned Objects \cite{downs2022google}. The chosen objects cover a wide range of complex and non-convex shapes, such as cars, shoes, and animals (see \figref{fig:merged_dataset}). We only choose objects that are asymmetric or only reflective symmetric, which mitigates the multi-modality issue in defining object poses \cite{chen2021system}. However, such a choice for training does not restrict our system from reorienting symmetric objects, a claim we empirically evaluate. $\smalldataset$ is used for evaluating generalization performance. It contains $12$ objects from the ContactDB \cite{Brahmbhatt_2019_CVPR} dataset with no overlapping object shape with dataset $\bigdataset$. We use a subset of $5$ objects from $\smalldataset$ to evaluate out-of-distribution (OOD) performance in the real world.

\paragraph{Object dataset preprocessing}

In order to make the objects manipulable by the robot hands, we need to scale their meshes to proper sizes. In simulation, we center each mesh and manually scale each object mesh to a proper size compared to the robot hand size. Overall, the longest side of the objects' bounding boxes lies in the range of $[0.095, 0.165]$m. The mass of each object is randomly sampled from $[0.03, 0.18]$kg. The mass of the objects used in the real-world tests is shown in \tblref{tbl:object_mass}.

\paragraph{Convex Decomposition} We use approximate convex decomposition (V-HACD \cite{mamou2016volumetric}) to perform an
approximate convex decomposition on the object and the robot hand meshes for fast collision detection in the simulator. The decomposition resolution is $100,000$.

\subsubsection*{Training setup}
\label{app_subsec:training_setup}

\paragraph{Observation space for the teacher policy} The inputs to the teacher policy, $\bm{o}_t^\expsymbol\in\mathbb{R}^{19G+21}$, include joint positions ($\mathbb{R}^{3G}$) and velocities ($\mathbb{R}^{3G}$), fingertip pose ($\mathbb{R}^{7G}$) and velocities ($\mathbb{R}^{6G}$), object pose ($\mathbb{R}^{7}$) and velocity ($\mathbb{R}^{6}$), target orientation expressed as a quaternion ($\mathbb{R}^{4}$), and the rotation difference between current and target object orientation expressed as a quaternion ($\mathbb{R}^{4}$), where $G$ represents the number of fingers. The pose of each finger is represented by a position ($\mathbb{R}^3$) and an orientation component (quaternion; $\mathbb{R}^4$). 

\paragraph{Teacher policy architecture} Our teacher policy is an MLP (Multilayer Perceptron) network consisting of three hidden layers (512, 256, 256 neurons) and ELU (Exponential Linear Unit) activation functions \cite{clevert2015fast}. We use Adam \cite{kingma2014adam} to optimize the networks.

\paragraph{GPU hardware}
In our experiments, we use one NVIDIA GeForce RTX 3090 for training the teacher policies, and one NVIDIA Tesla V100 for training the student (vision) policies.

\paragraph{Hyper-parameters}

\tblref{tbl:hyper_params} lists the hyper-parameters used in the experiments.

\paragraph{Point cloud voxelization}
The point cloud input to the policy network has no color information and is voxelized in a resolution of $0.005$m.

\paragraph{Success criteria}
The success criterion defines when the agent has accurately reoriented the object in the target configuration. Its purpose is two-fold: a reward signal during training and a criterion that signals success to stop the reorientation policy and thereby end the episode during training. A straightforward success criterion is judging whether an object's orientation is close to the target orientation (orientation criterion). The controller learned using the criterion when evaluated in simulation results in a behavior wherein the robotic fingers stabilize the object when its orientation is close to the target. However, the same controller, when evaluated in the real world, often does not result in fingers stopping when the object orientation is close to the target leading to overshooting. Consequently, instead of stopping, the object oscillates around the target orientation. We believe this is a result of sim-to-real gaps, including the control latency, observation noise, and the difference in dynamics. We ameliorate this issue by expanding the definition of success criterion to penalize finger and object motions explicitly. The task is considered completed successfully in the simulation if all the following three criteria are satisfied. First, the orientation criterion is satisfied when $|\Delta \theta_t| < \Bar{\theta}$ where $\Delta \theta_t$ is the distance between the object's current and target orientation. Second, the finger motion criterion requires the joint motion of the robot to be small and is satisfied when $||\dot{\bm{q}}_t||<\bar{\dot{q}}$ and $||\act||<\bar{a}$ where $\dot{\bm{q}}_t$ is the joint velocities at time step $t$, $\act$ is the policy output.
Third, the object motion criterion requires the object's velocity to be small. It is satisfied when $||{\bm{v}}^o_t||<\bar{v}$ and $||{\bm{\omega}}_t^o||<\bar{\omega}$ where $\bm{v}_t^o, \bm{\omega}_t^o$ denote object's linear and angular velocities respectively. 

$\Bar{\theta}, \bar{\dot{q}}, \bar{a}, \bar{v}, \bar{\omega}$ are manually defined thresholds. The finger and object motion criteria act as regularizers to explicitly encourage the policies to slow down the motion near the end.

\paragraph{Compute cost} 
Energy Cost: We used a single NVIDIA V100 (32GB memory) GPU to train the policy. The total training time, including the teacher and two student stages, was less than 400 hours. The GPU has a maximum power consumption of 250W when running at full capacity, but our learning system did not always use the GPU to its fullest extent. Therefore, the maximum power consumption was not always reached, resulting in a GPU power consumption of no more than 100kWh. Based on the average electricity cost in the United States of 15.64 cents/kWh in May 2022, the total cost of GPU computing is less than \$15.6. Additionally, if we consider the energy cost of other workstation components, such as the CPU and fans, the power rate remains under 1000W. Running the entire system at full utilization for 400 hours would cost around \$62.4.

GPU Cost: The V100 GPU model designed for HPC (High-performance Computing) is relatively expensive, costing about \$4K on Amazon. We used it only because it is the default GPU model in our servers. However, our training is not limited to V100 and can also be performed on other GPUs. For example, the training can be done using a 3090 RTX GPU, which costs about \$1.4K. During deployment, we have also successfully run the policy on a workstation with a 2080Ti GPU, which costs only \$700. Therefore, the cost of computing hardware can vary greatly depending on the available GPUs. Our policy network is not very large, so it can be trained or deployed on cheaper GPUs, such as the 3090 RTX.

\subsubsection*{Real-world setup}
\label{appsubsec:real_world_setup}
We used two robot hands in our experiments: a three-fingered hand (\dclaw) and a four-fingered hand. The three and four-fingered hands consist of nine and twelve Dynamixel motors, respectively. The hands are fixed on an 80/20 aluminum frame. Since our goal is to construct a real-world ready reorientation system that can be used on a mobile manipulator in the future, we only use one RealSense D415 camera to observe the robotic hand manipulate the object. The robot and camera are calibrated using the dual quaternions method \cite{daniilidis1999hand}. We only perform camera calibration on one of the robot fingers, and the ArUco marker is attached to the fingertip. Due to the limits on the finger's motion (3 DoFs), it is impossible to span a broad range of positions for the ArUco marker, resulting in noisy calibration with noticeable errors. Empirically we found such errors didn't influence the performance of our system. We use ROS \cite{quigley2009ros} for communication with the Dynamixel motors and use a threading lock to prevent simultaneous reading and writing on the motors as Dynamixel motors use a half-duplex UART (Universal Asynchronous Receiver Transmitter) for communication. Each Dynamixel motor is controlled via position control in $12$Hz. The Dynamixel-specific parameters noted in Dynamixel's control table are set as follows: $P$ is $200$, and the $D$ gain is $10$. The observations available to the controller are the joint positions of each motor and the depth image from the Realsense camera.

\paragraph{Real-world observation} The observation consists of the joint positions of each motor in the robotic hand and the depth image from the RealSense camera. We convert both these sensory inputs into a point cloud. The joint angles are converted into a point cloud as follows: Using the robot's CAD model, we uniformly sample points on each link and cache it. Given a sequence of joint positions, we use forward kinematics to compute the pose of each link and accordingly transform each of the associated pre-cached point clouds. We call the concatenated point cloud of all links as proprioceptive point cloud. The depth image is also converted into a 3D point cloud (exteroceptive point cloud). Both point clouds are merged and used as inputs to our policy. We do not assume access to any other sensory information. 

\paragraph{Stopping criteria} The robot stops when it is deemed successful as per the success criteria described above. Evaluating the success criteria requires knowledge of object motion (object motion criterion), orientation distance (orientation criterion), and fingertip motion (finger motion criterion). Although we can measure the fingertip motion ($\dot{\bm{q}}_t$ and $\act$) in the real world, we cannot directly measure the object's motion and pose. To obtain the orientation error, we train a predictor that re-uses features from the policy network to predict $|\Delta \theta_t|$ (see Figure 7A). Predicting object motion from point clouds is harder, and we found it unnecessary to estimate. We found that satisfying only the orientation criterion and finger motion criterion is sufficient to stop the object at the target orientation successfully. We also found it sufficient only to check $||\act||<\bar{a}$ to detect finger motion during real-world deployment. 

\paragraph{Real-world quantitative evaluation setup}
We set up a motion capture system using six OptiTrack cameras to evaluate the policy performance in the real world quantitatively. We add markers on the surface of evaluation objects. Even though the added markers add little bumps on the object's surface and therefore change the dynamics, we found our policies to be robust enough to deal with these changes. The output of the motion capture system is the object pose. We use the tracked object pose when the stopping criteria are satisfied to compute the error from the target orientation. We only use the motion capture system for quantitative evaluation, and it is not required by our system to reorient objects. Note that the motion capture system can occasionally fail to track the objects when the fingers heavily occluded the markers. When it happens, we discard this test as we cannot get a quantitative error in this case.

\subsubsection*{Computing the orientation error for symmetric objects}
\label{app_subsec:compute_error_symm}
With symmetric objects, it is hard to determine whether the controller completes the task or not unless we know in what way the object is symmetric. We can still use a motion capture system to get the object's pose. However, we cannot directly compute the distance between the orientation from the motion capture system, and the goal orientation as there exist many different orientations that make objects look similar. Therefore, we need to find out all such possible goal orientations. Although scaling this up to a wide variety of objects is challenging, in our experiments, we tested on two symmetric objects (a rectangular cuboid and a cube) whose symmetric axes can be easily enumerated. In other words, on the rectangular cuboid and cube, we can easily identify all possible rotational axes upon which the objects can be rotated by some angle and end up with the same visual appearance. Then we compute the distances between the actual object orientation and all possible goal orientations and find the minimum distance.

\subsection*{Fabrication}
\begin{figure}[!htb]
    \centering
    \begin{subfigure}[t]{0.28\linewidth}
        \centering
        \includegraphics[width=0.95\linewidth]{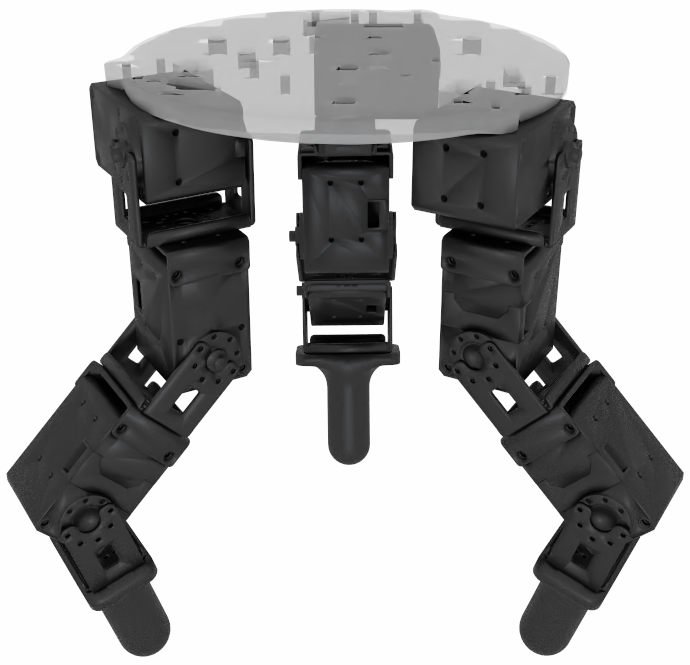}
        \caption{}
        \label{fig:dclaw}
    \end{subfigure}%
        \hfill
    \begin{subfigure}[t]{0.38\linewidth}
        \centering
        \includegraphics[width=0.95\linewidth]{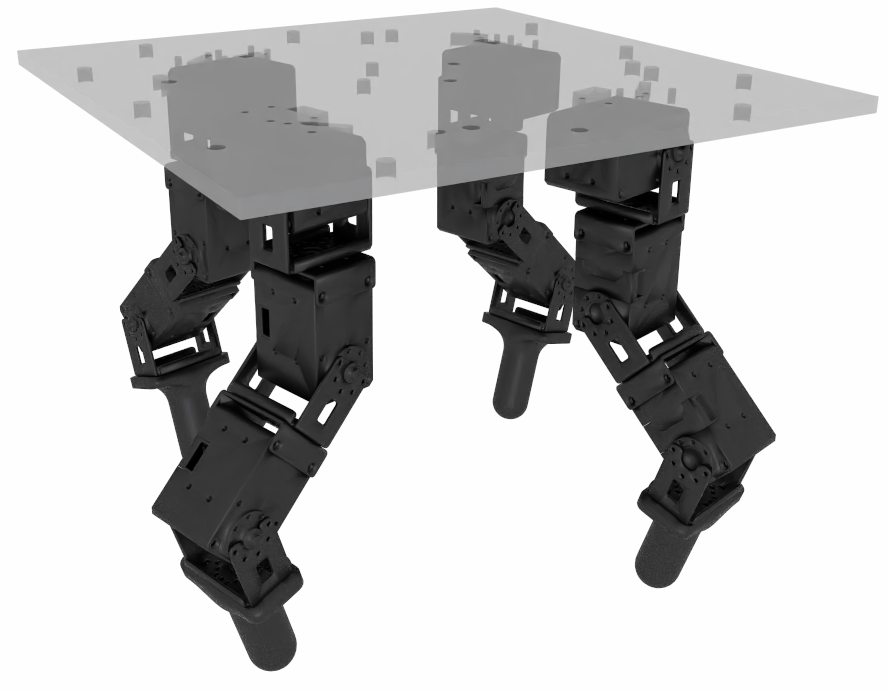}
        \caption{}
        \label{fig:dclaw_4f}
    \end{subfigure}%
        \hfill
    \begin{subfigure}[t]{0.33\linewidth}
        \centering
        \includegraphics[width=0.5\linewidth]{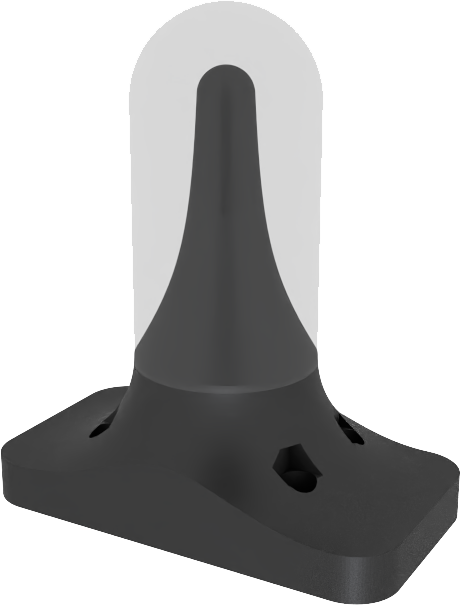}
        \caption{}
        \label{fig:soft_finger_rounded}
    \end{subfigure}%
    \caption{\textbf{3D models for the robot hands}. \textbf{(A)}: three-fingered robot hand. \textbf{(B)}: four-fingered robot hand. \textbf{(C)}: fingertips with a rounded skeleton and the grey shell represents soft elastomer.
 }
    \label{fig:hand_design}
\end{figure}

\paragraph{Fingertip Fabrication} 
We experimented with both rigid and soft fingertips. The rigid fingertips were fabricated using 3D printing. For the soft fingertips, we designed a rigid inner skeleton coated in a soft outer elastomer. The elastomer allows the robot’s fingertips to have increased compliance and friction when interacting with the plastic objects, as the plastic internal skeleton helps maintain the shape of the fingertips without too much deformation, similar to a human finger. The design for the internal skeleton used is shown in \figref{fig:hand_design}.

The fingertip skeletons and molds for the elastomer are 3D printed on the Markforge Onyx One using the Onyx filament. To help improve the adhesion of the silicone to the Onyx material, the skeletons are sanded with 400-grit sandpaper, corona treated, and primed using the Dow DOWSIL P5200 Adhesion Promoter. 

The gel coating for both fingertip designs is made from Smooth-On’s Ecoflex platinum-catalyzed silicone. The elastomer has a shore hardness of 00-10 and exhibits a tacky finish when fully cured. A ratio of 1:0.008:0.005 by weight of the Ecoflex mixture (Parts A and B combined),  Smooth-On Silc Pig White, and Smooth-On Silc Pig Black are combined to provide the gray color of the fingertips. To ensure the surface of the gel elastomer is smooth, XTC-3D is applied to the inside of the 3D-printed molds to smooth out any texture. The uncured Ecoflex mixture is poured into the molds and degassed to eliminate air bubbles from forming on the elastomer’s surface. The skeletons are pushed into the top-down molds and left to cure at room temperature for 4 hours.

\paragraph{Object Fabrication}
We use 3D printing to fabricate objects that are used for quantitative evaluation. Each object was printed with a 0.25mm layer height on the Lulzbot TAZ Pro Dual Extruder using PolyTerra PLA filament in different colors. \tblref{tbl:object_mass} lists the mass of each object used in real-world experiments. Note that to test the transfer of our results, we also included real household objects in our test set.

\subsection*{Overcoming sim-to-real gap}
\subsubsection*{Dynamics Identification}
\label{appsec:dynamics_id}

\paragraph{Action commands} In this work, we send two types of action commands to the robot and collect the joint movement trajectories. The first type of action command is a step command ($\bm{A}(t)=c$ where $c$ is a constant joint position command). We collect the step responses from the motors. The second type of action command is $\sin$-wave action commands in different frequencies. The $\sin$-wave command is $\bm{A}(t) = \sin(2\pi ft)$ where $f \in [0.05, 1.5]$Hz. For both types of action commands, we scale the amplitude proportionally to the joint limit of each joint. 

\begin{figure}[!htb]
    \centering
    \begin{subfigure}[b]{0.3\linewidth}
         \centering
         \includegraphics[width=\linewidth]{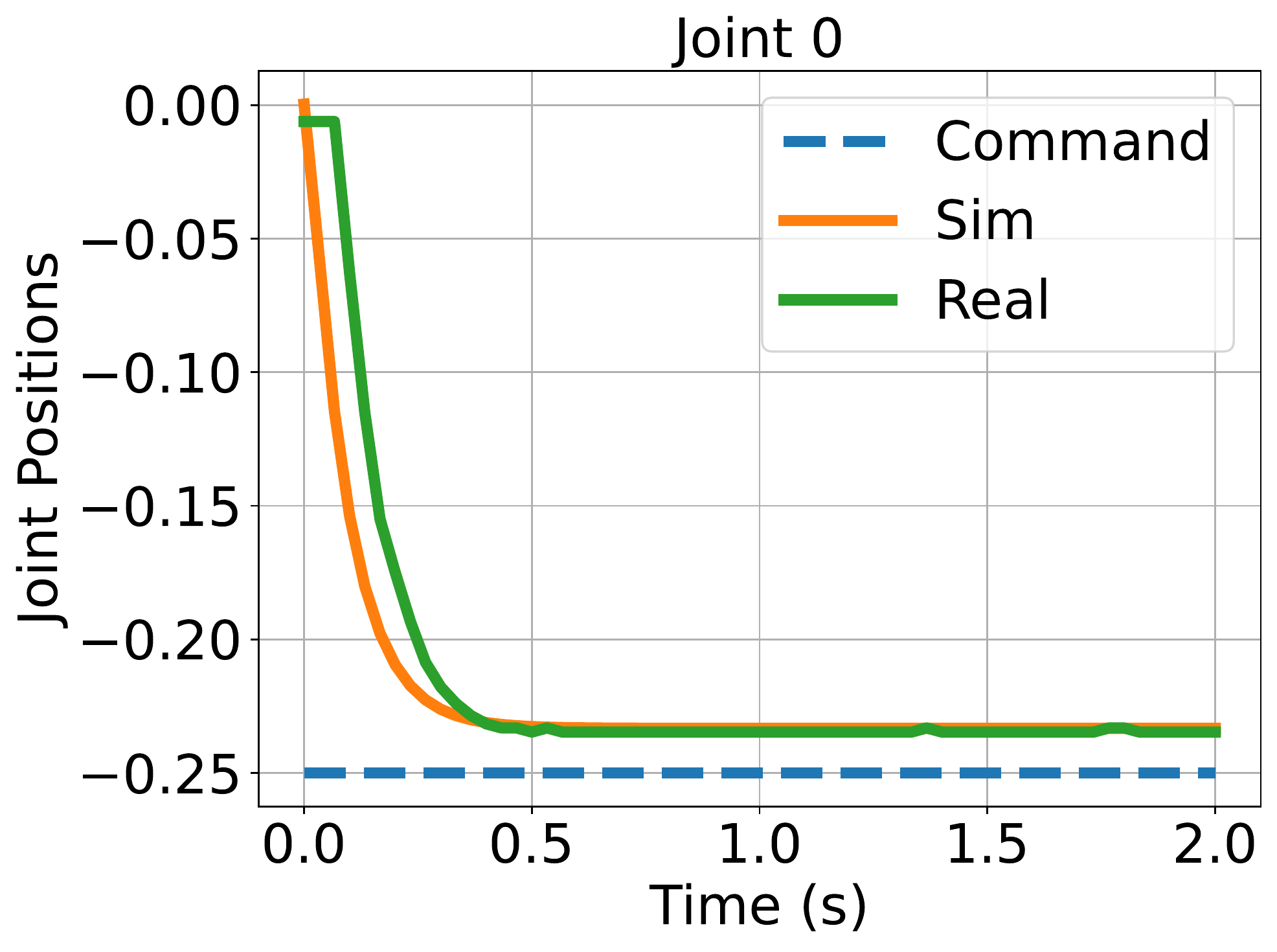}
     \end{subfigure}%
     \quad
     \begin{subfigure}[b]{0.3\linewidth}
         \centering
         \includegraphics[width=\linewidth]{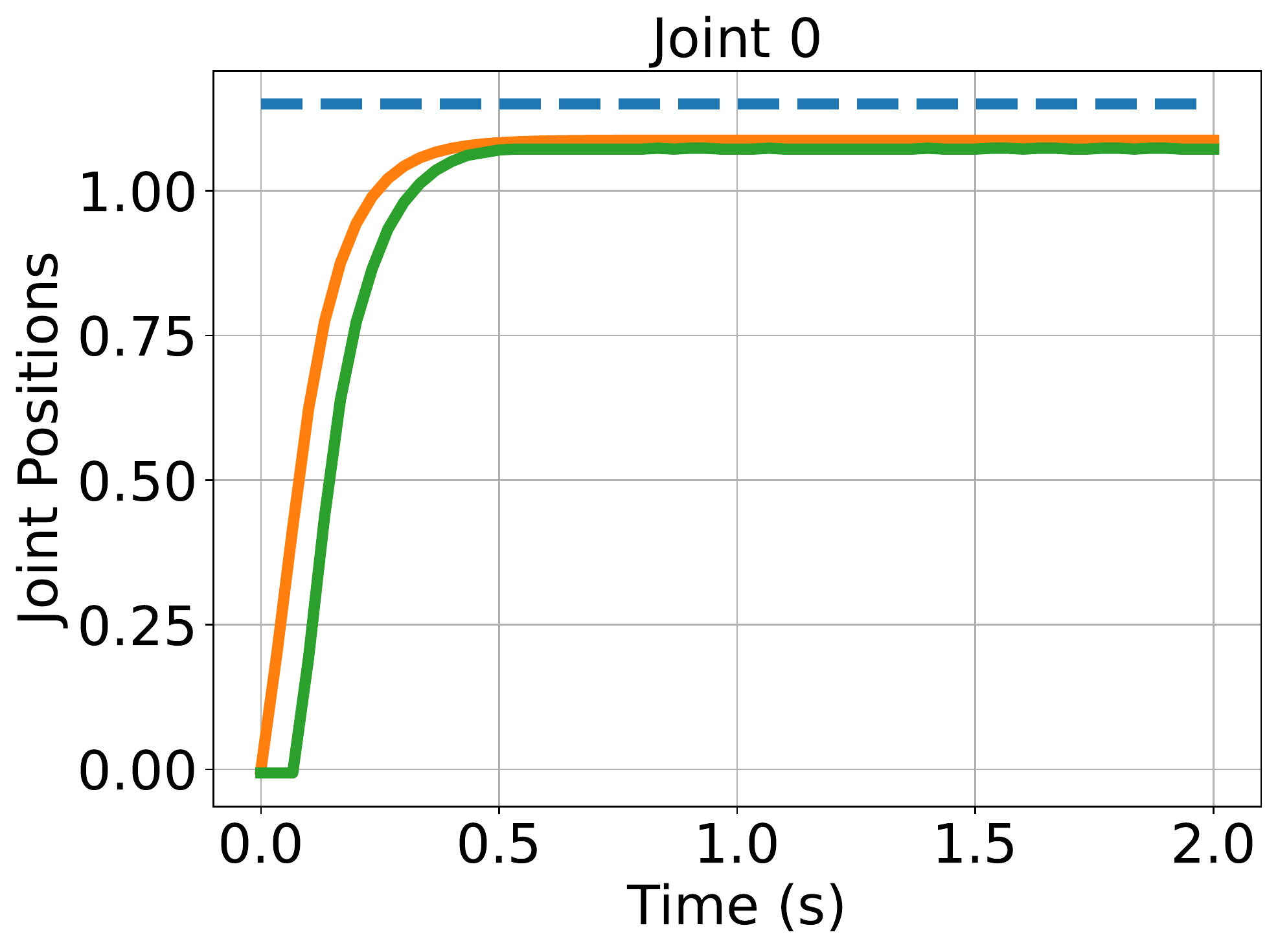}
     \end{subfigure}\\
         \begin{subfigure}[b]{0.3\linewidth}
         \centering
         \includegraphics[width=\linewidth]{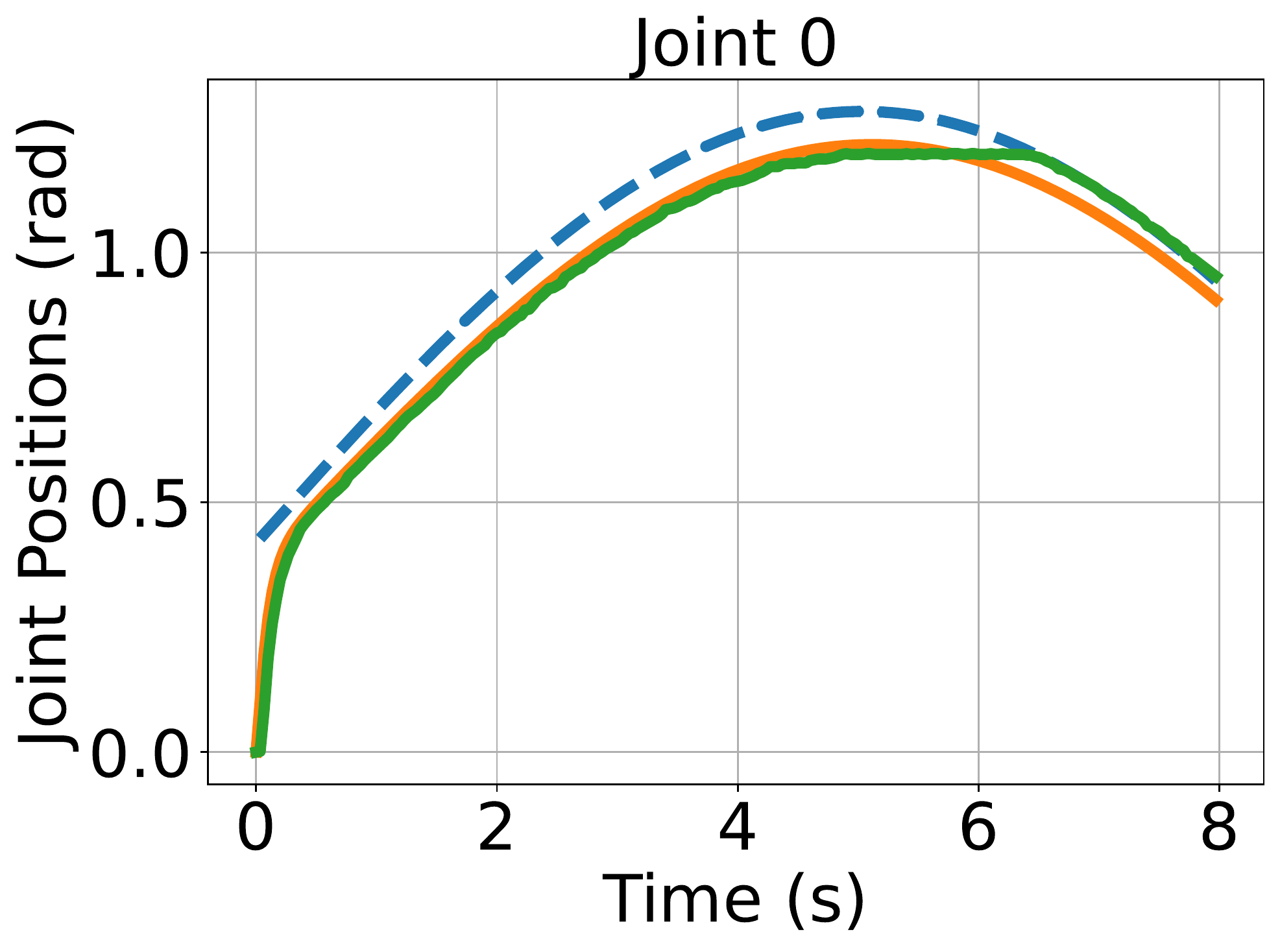}
     \end{subfigure}%
     \quad
     \begin{subfigure}[b]{0.3\linewidth}
         \centering
         \includegraphics[width=\linewidth]{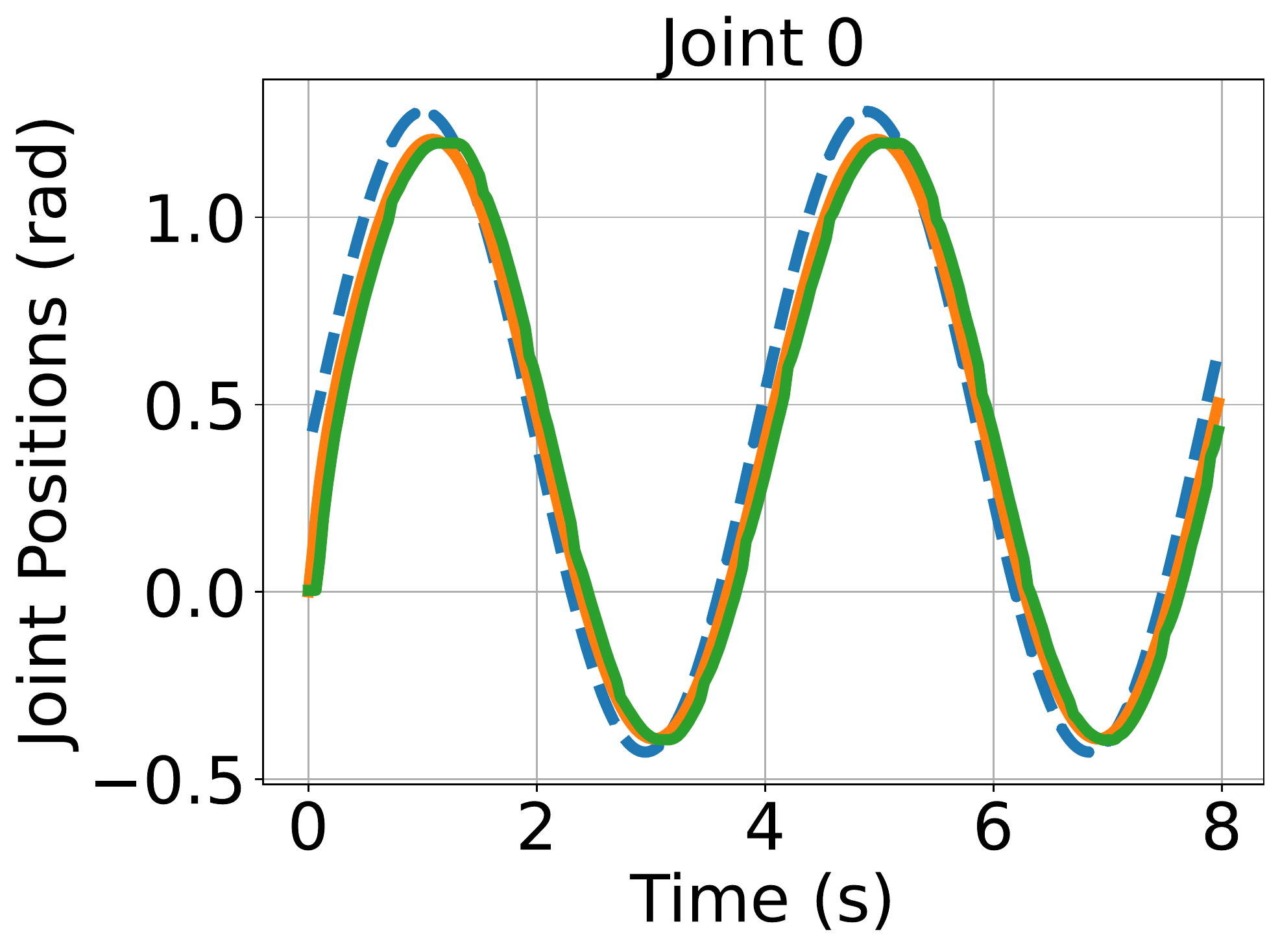}
     \end{subfigure}\\
    \begin{subfigure}[b]{0.3\linewidth}
         \centering
         \includegraphics[width=\linewidth]{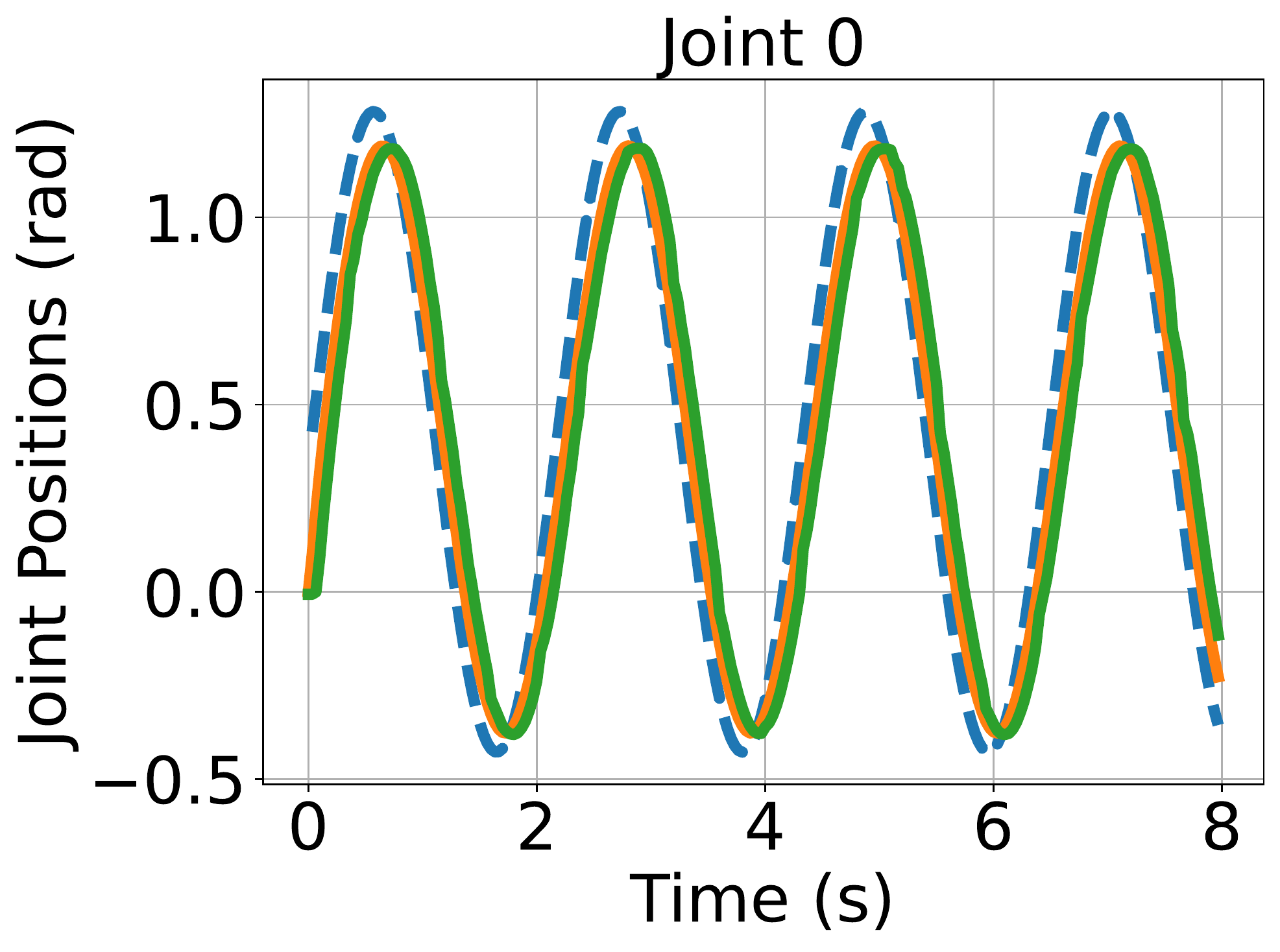}
     \end{subfigure}%
     \quad
    \begin{subfigure}[b]{0.3\linewidth}
         \centering
         \includegraphics[width=\linewidth]{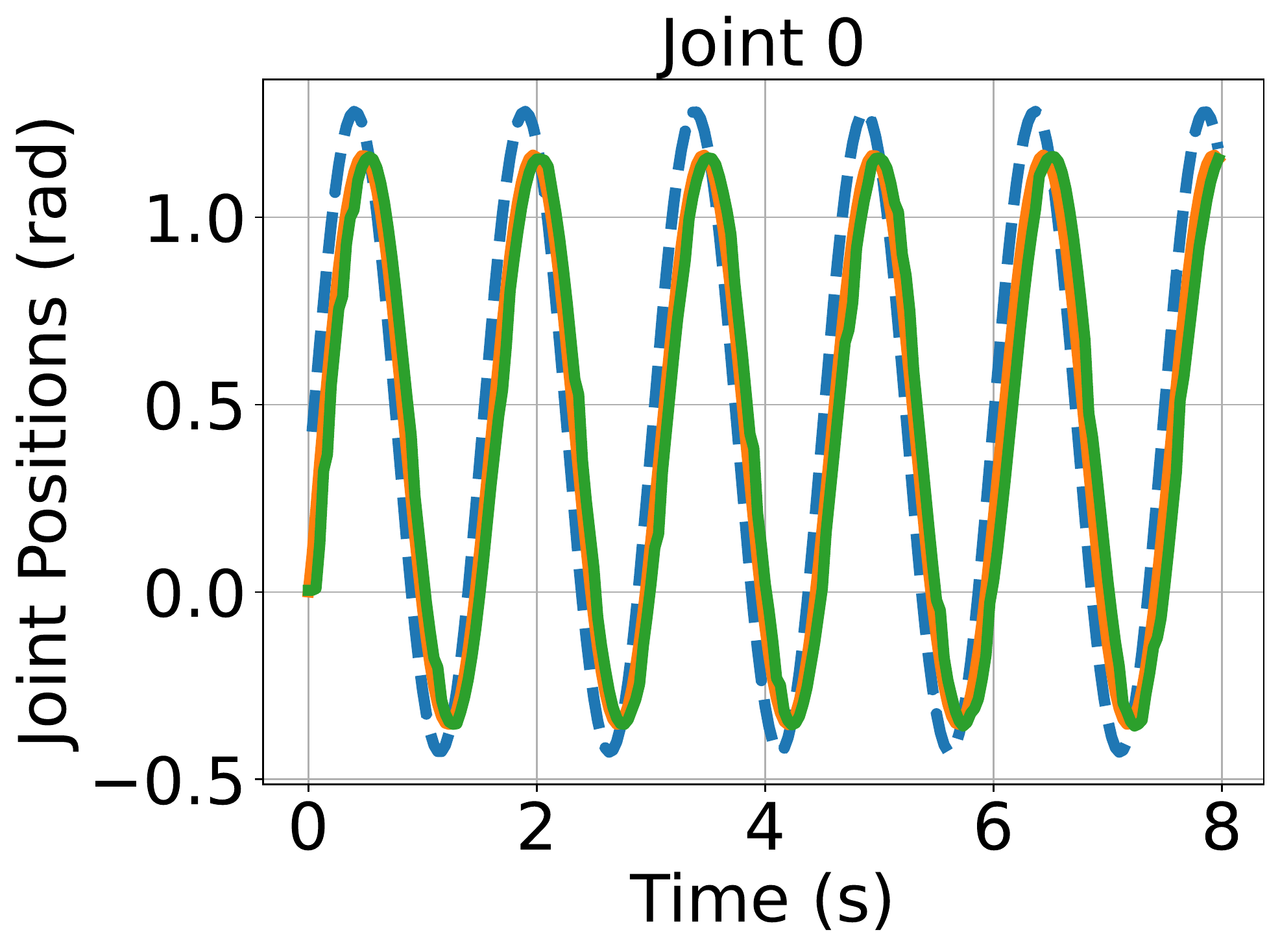}
     \end{subfigure}
\\
    \begin{subfigure}[b]{0.3\linewidth}
         \centering
         \includegraphics[width=\linewidth]{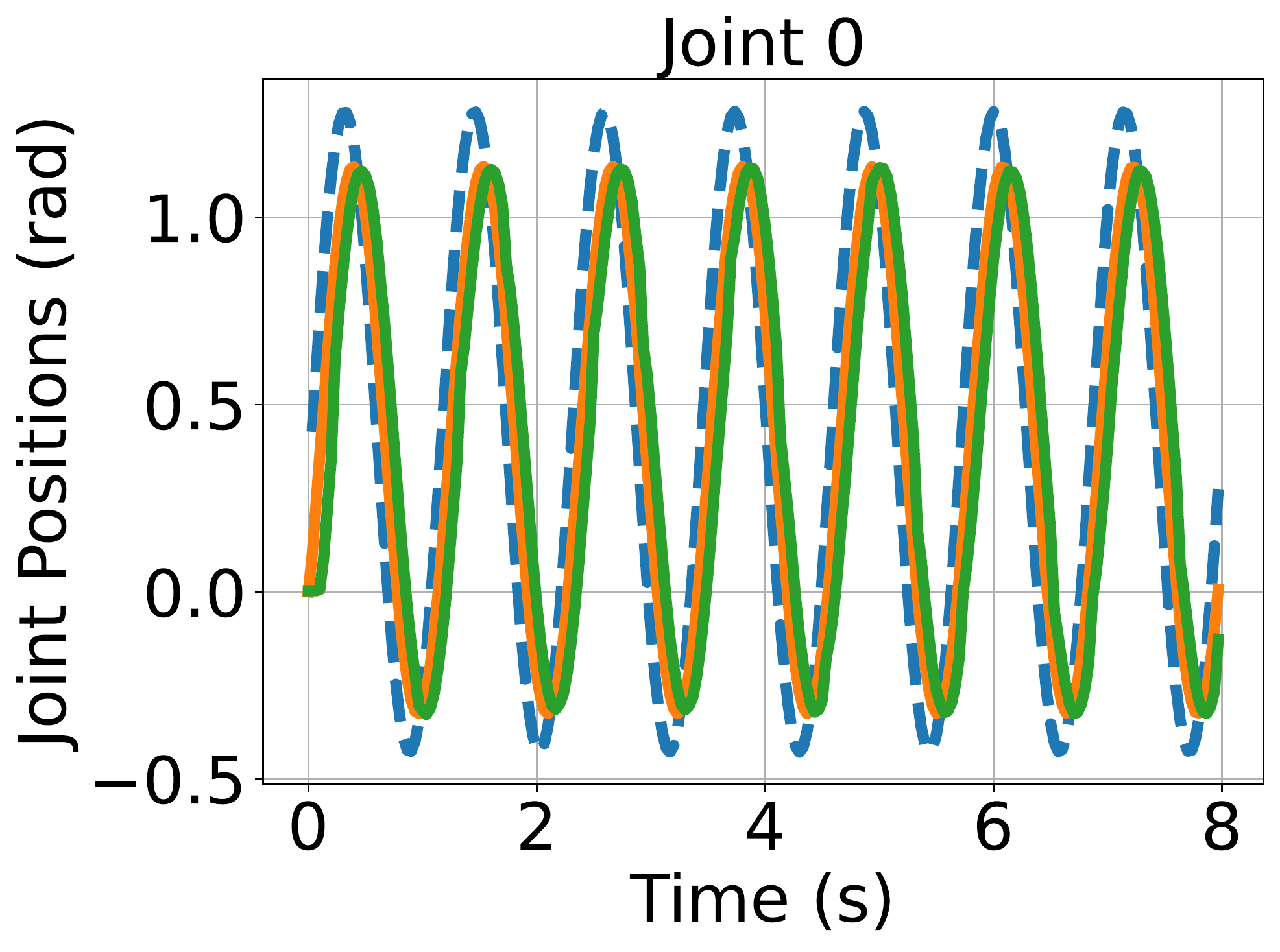}
     \end{subfigure}%
     \quad
    \begin{subfigure}[b]{0.3\linewidth}
         \centering
         \includegraphics[width=\linewidth]{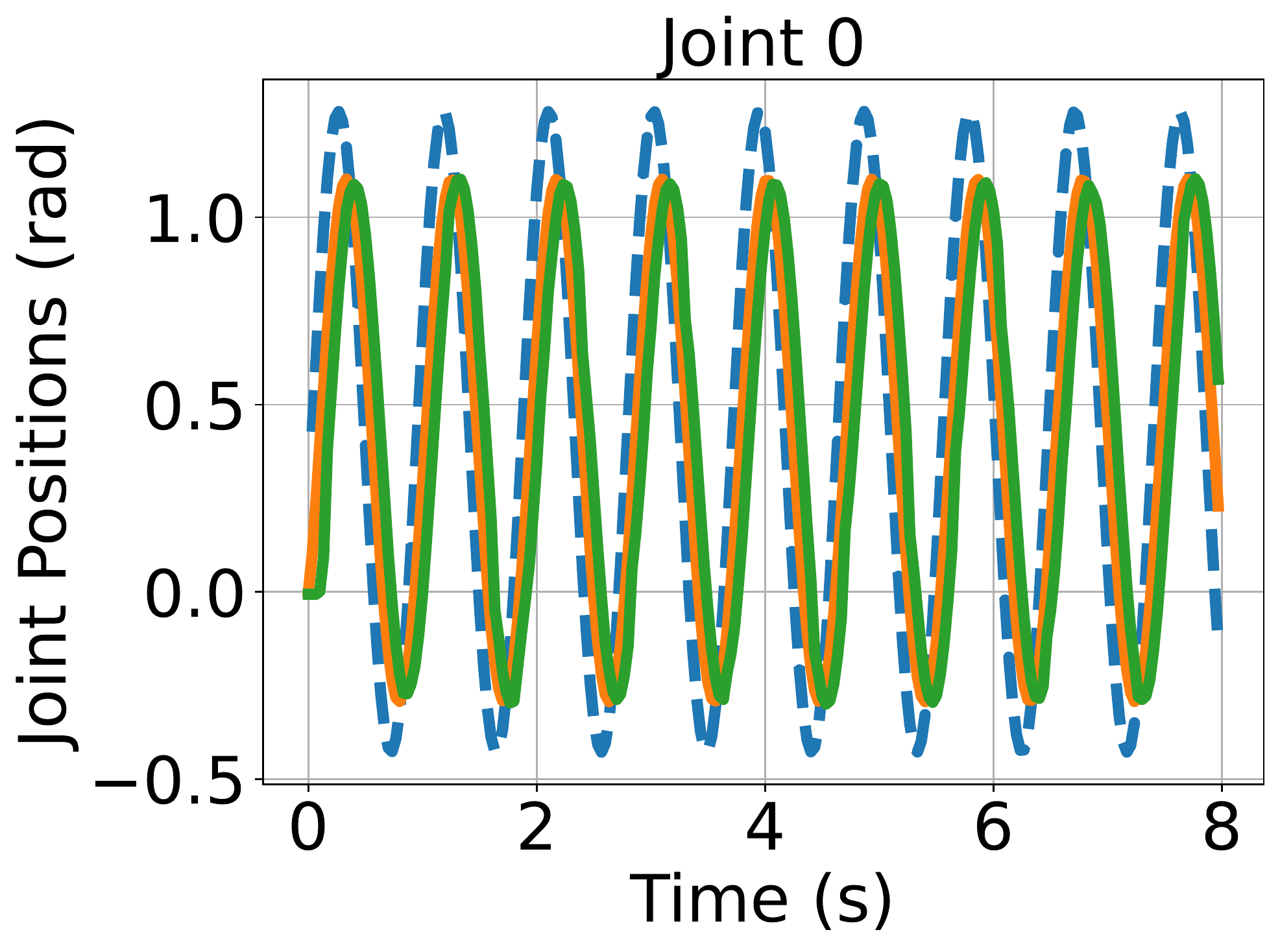}
     \end{subfigure}\\
         \begin{subfigure}[b]{0.3\linewidth}
         \centering
         \includegraphics[width=\linewidth]{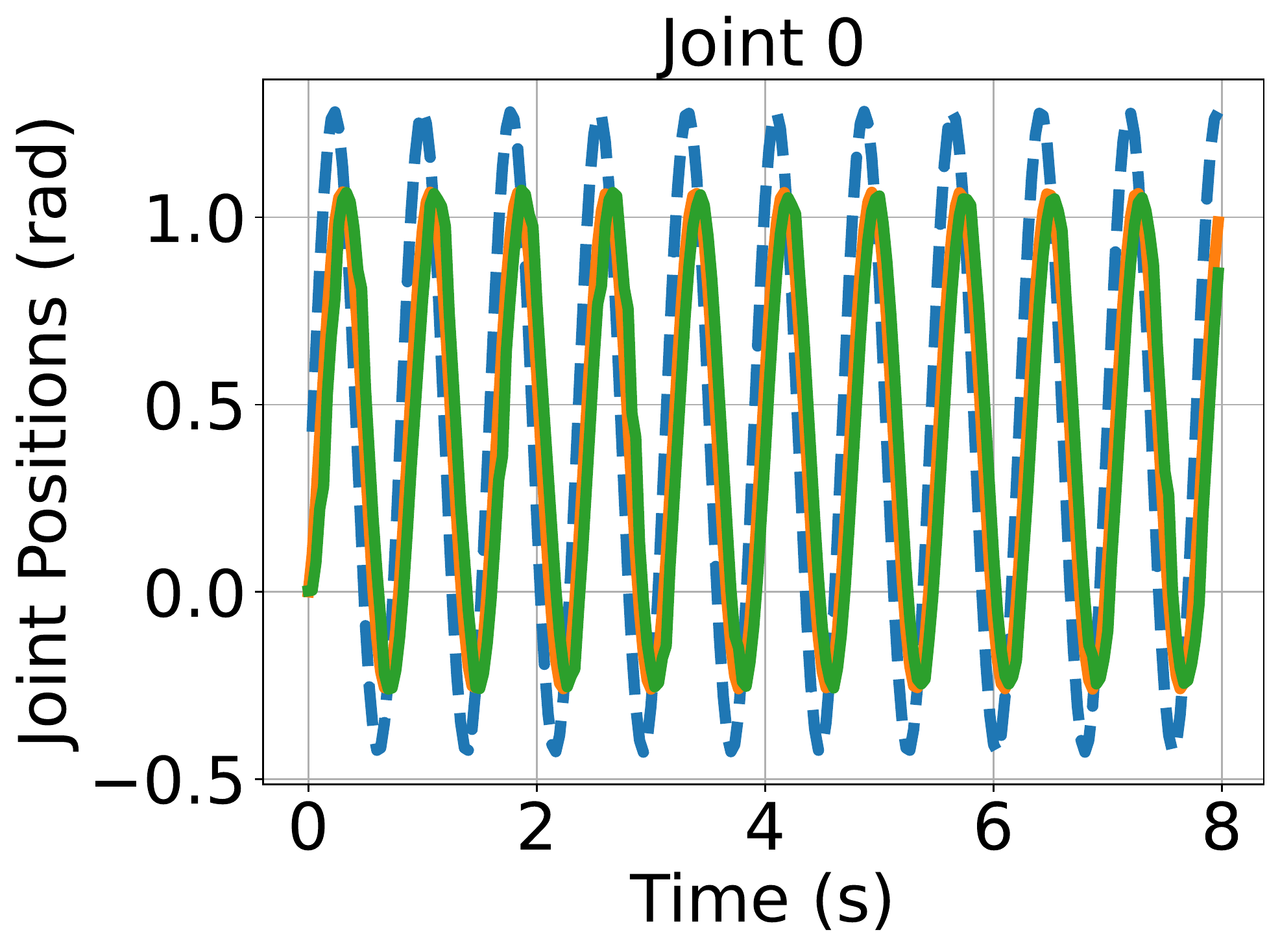}
     \end{subfigure}%
     \quad
     \begin{subfigure}[b]{0.3\linewidth}
         \centering
         \includegraphics[width=\linewidth]{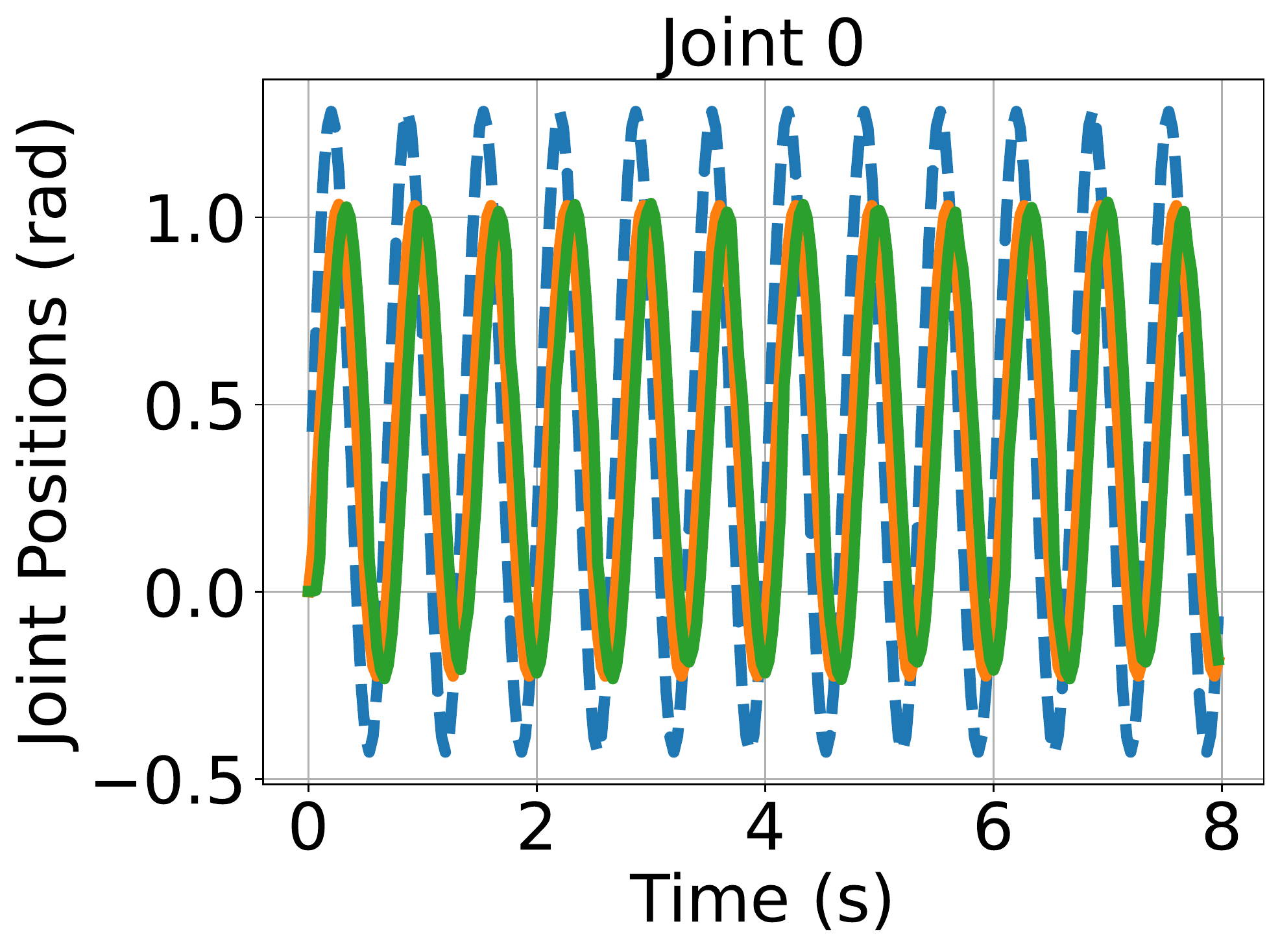}
     \end{subfigure}
    \caption{\textbf{Joint response curves}. We identified the dynamics of a finger joint (the top one) and show the results. Three curves are plotted: (1) the command sent to the joint, (2) the joint's simulated response using the identified dynamics parameters, (3) the joint's real-world response. The identified dynamics parameters allow the simulated joint to move similarly to the real joint. 
    }
    \label{fig:sys_id_jnt0}
\end{figure}

\begin{figure}[!htb]
    \centering
    \begin{subfigure}[b]{0.3\linewidth}
         \centering
         \includegraphics[width=\linewidth]{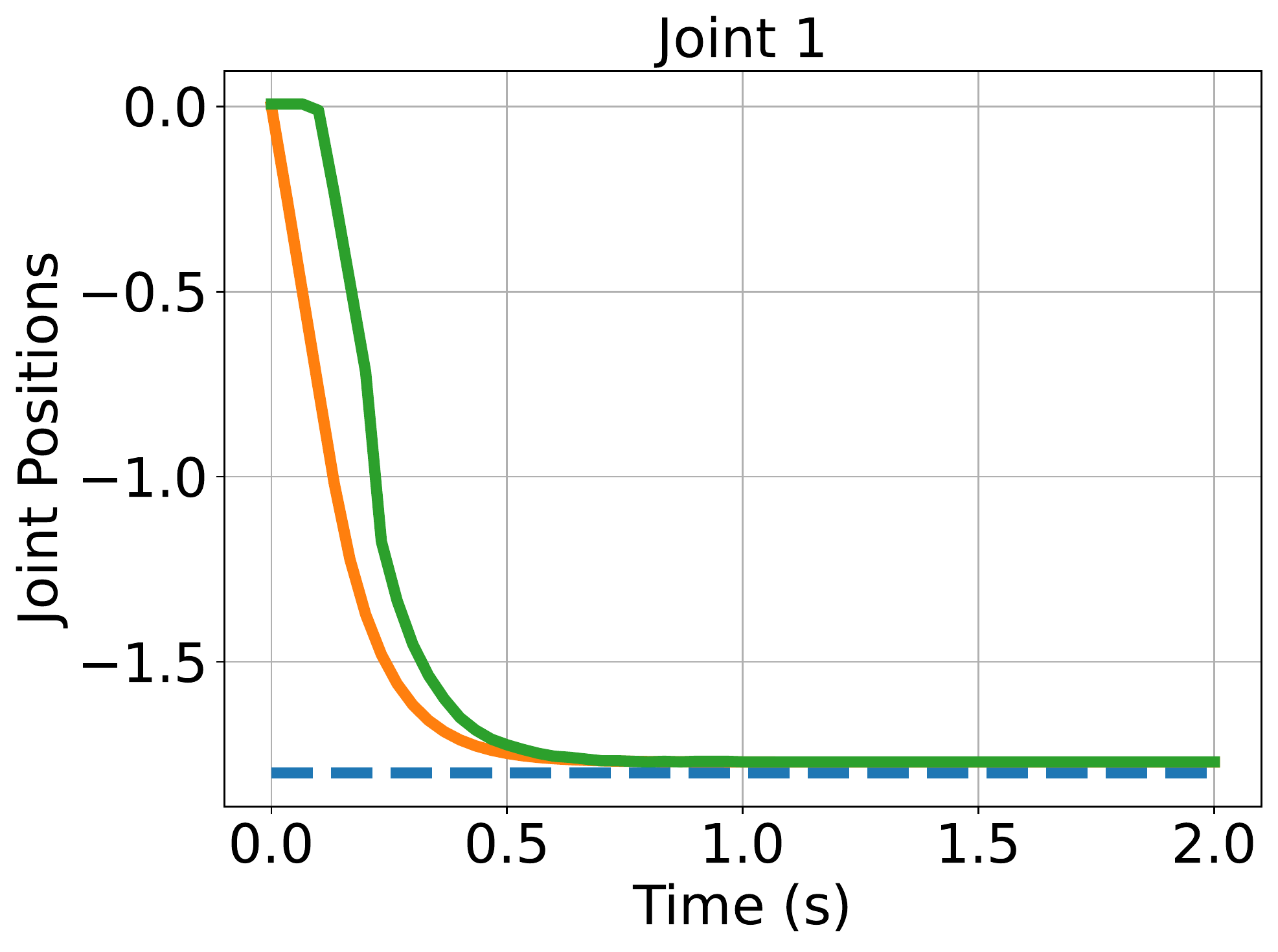}
     \end{subfigure}%
     \quad
     \begin{subfigure}[b]{0.3\linewidth}
         \centering
         \includegraphics[width=\linewidth]{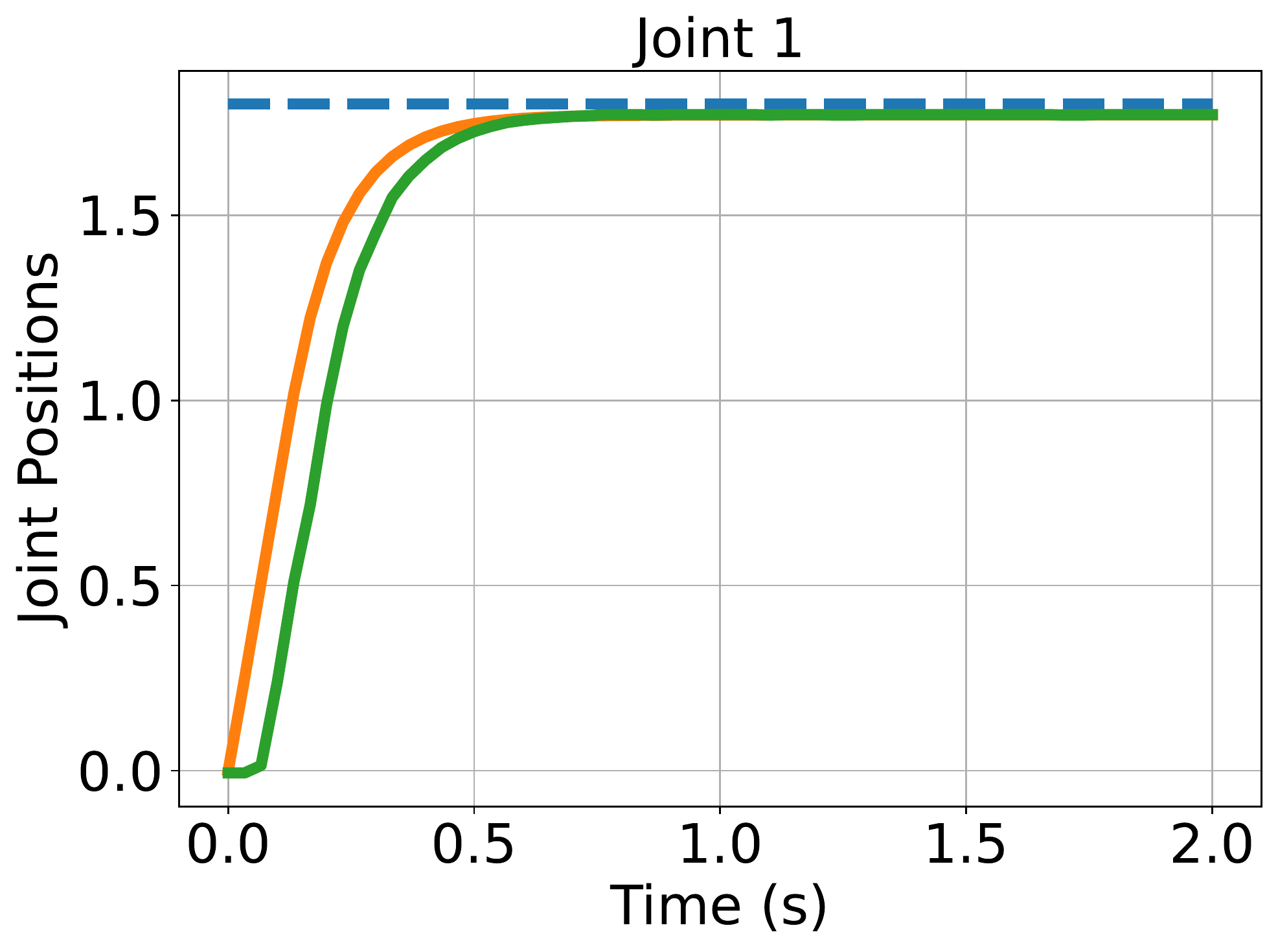}
     \end{subfigure}\\
         \begin{subfigure}[b]{0.3\linewidth}
         \centering
         \includegraphics[width=\linewidth]{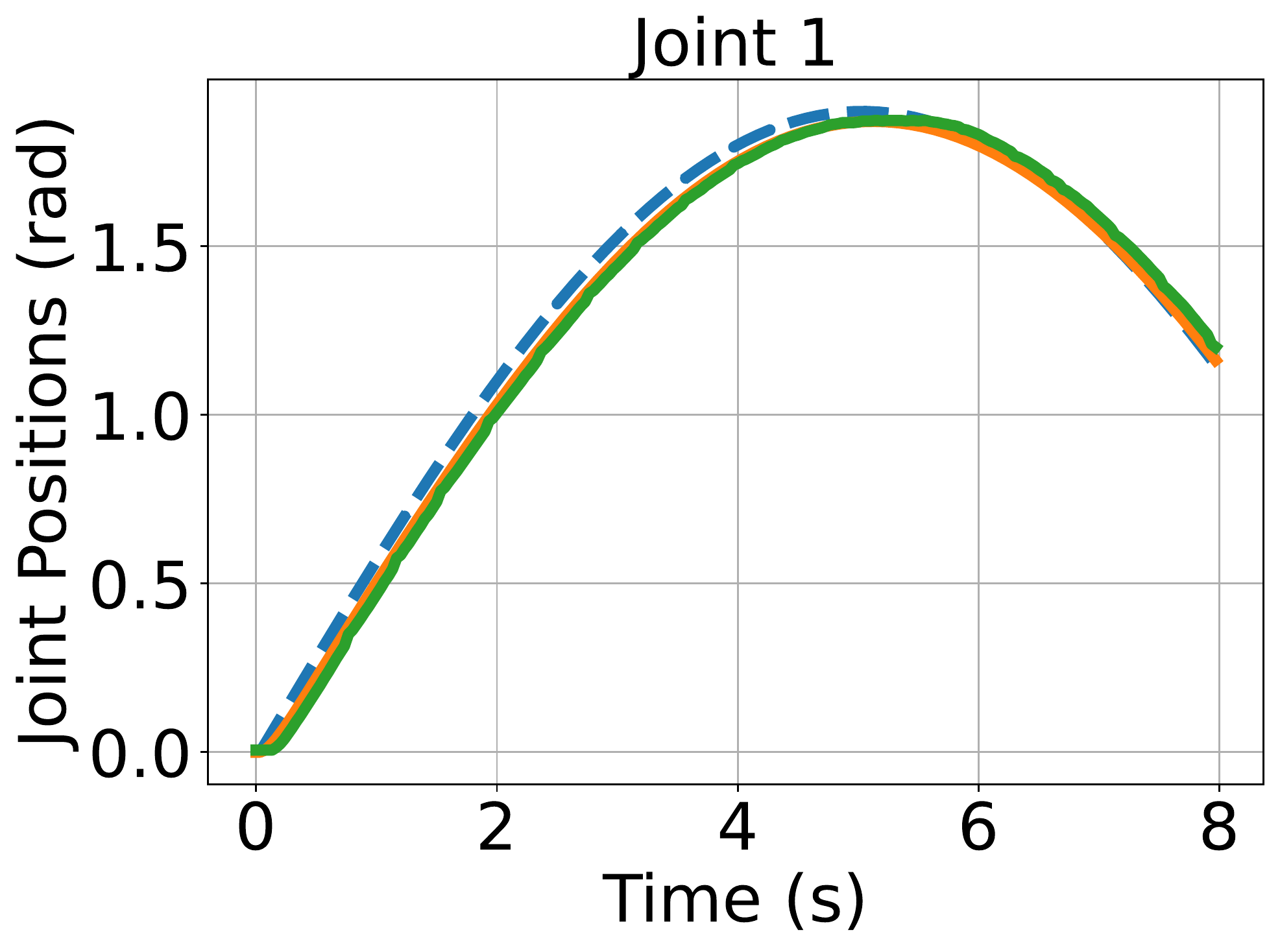}
     \end{subfigure}%
     \quad
     \begin{subfigure}[b]{0.3\linewidth}
         \centering
         \includegraphics[width=\linewidth]{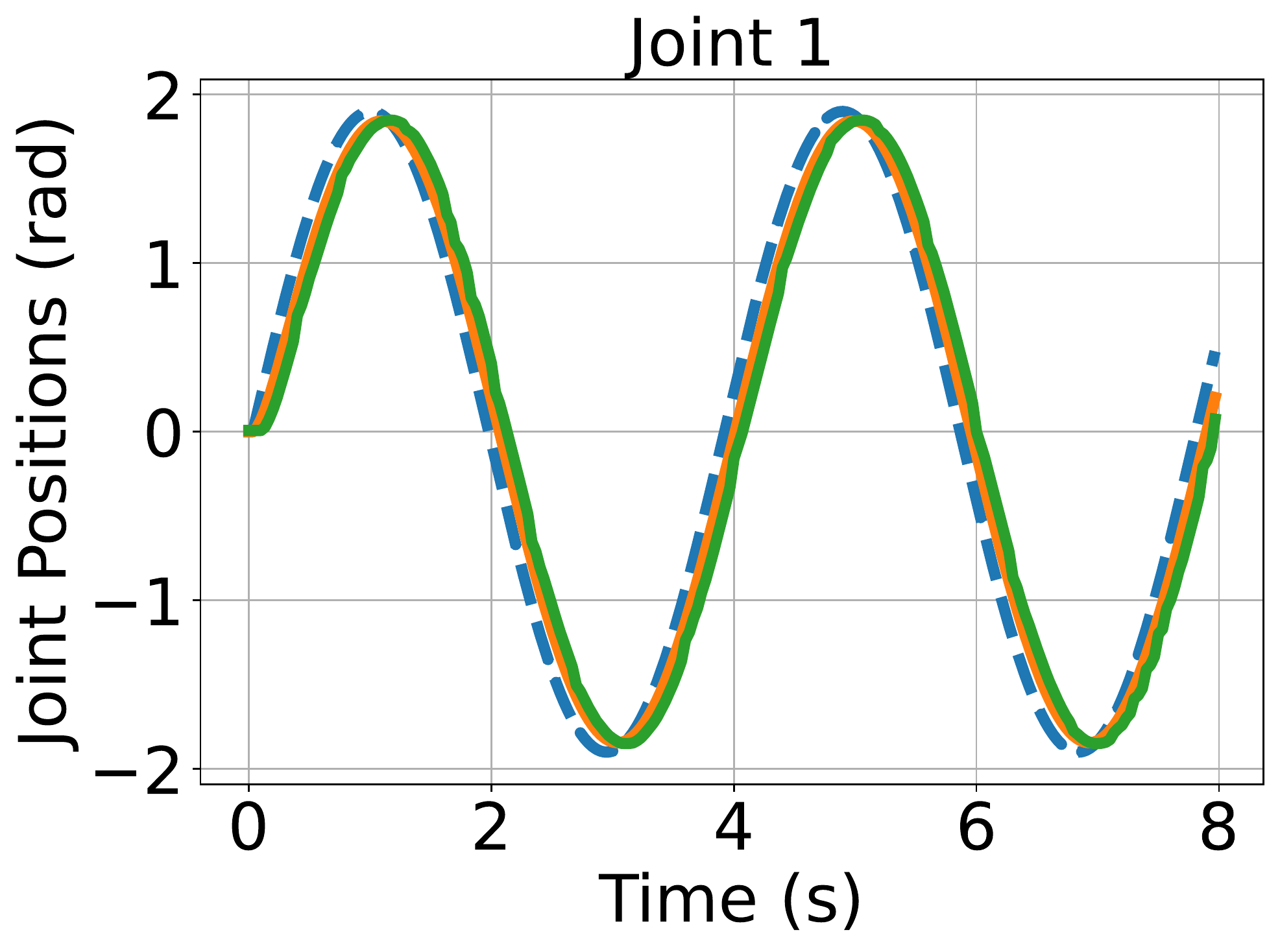}
     \end{subfigure}\\
    \begin{subfigure}[b]{0.3\linewidth}
         \centering
         \includegraphics[width=\linewidth]{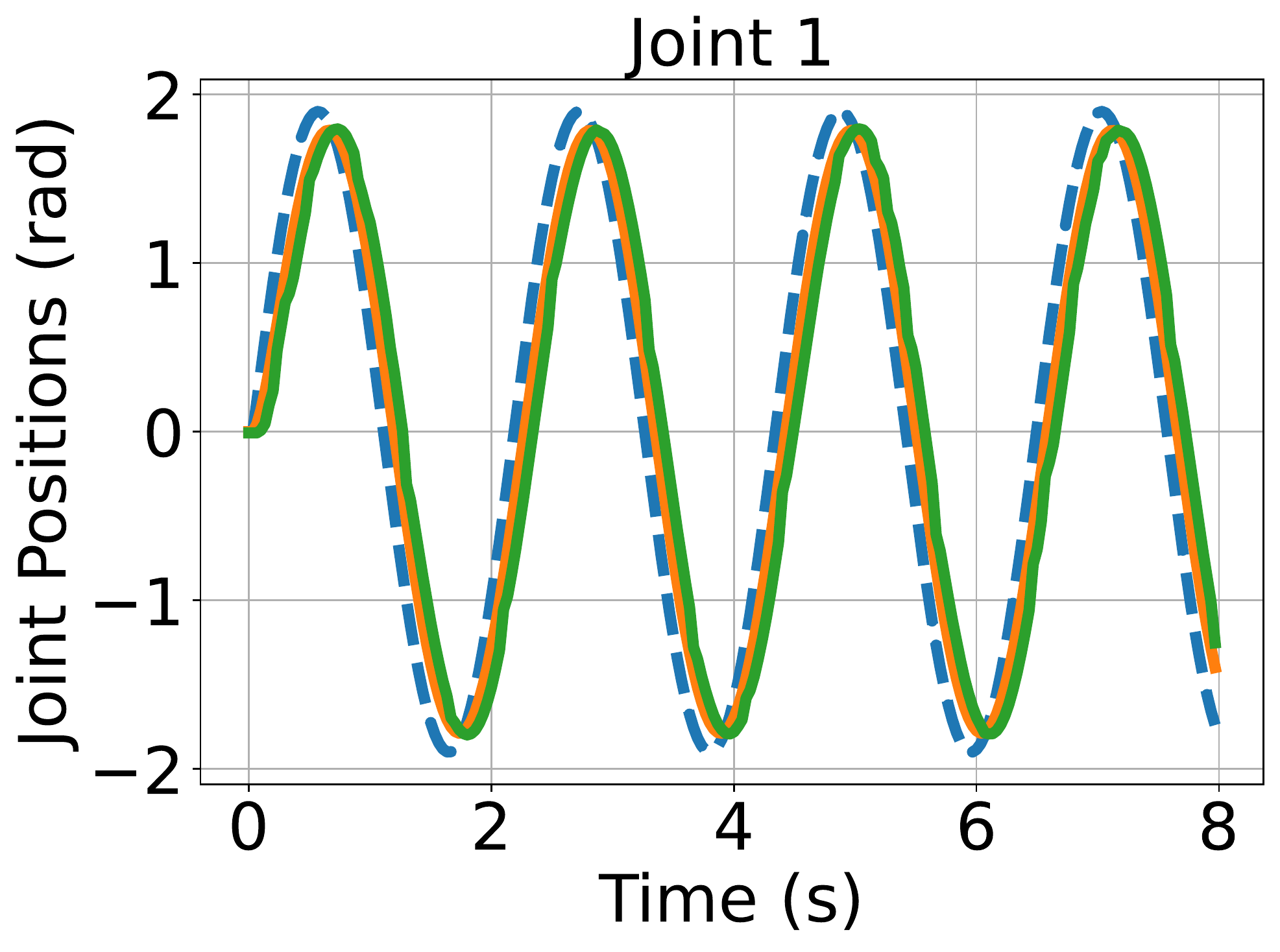}
     \end{subfigure}%
     \quad
    \begin{subfigure}[b]{0.3\linewidth}
         \centering
         \includegraphics[width=\linewidth]{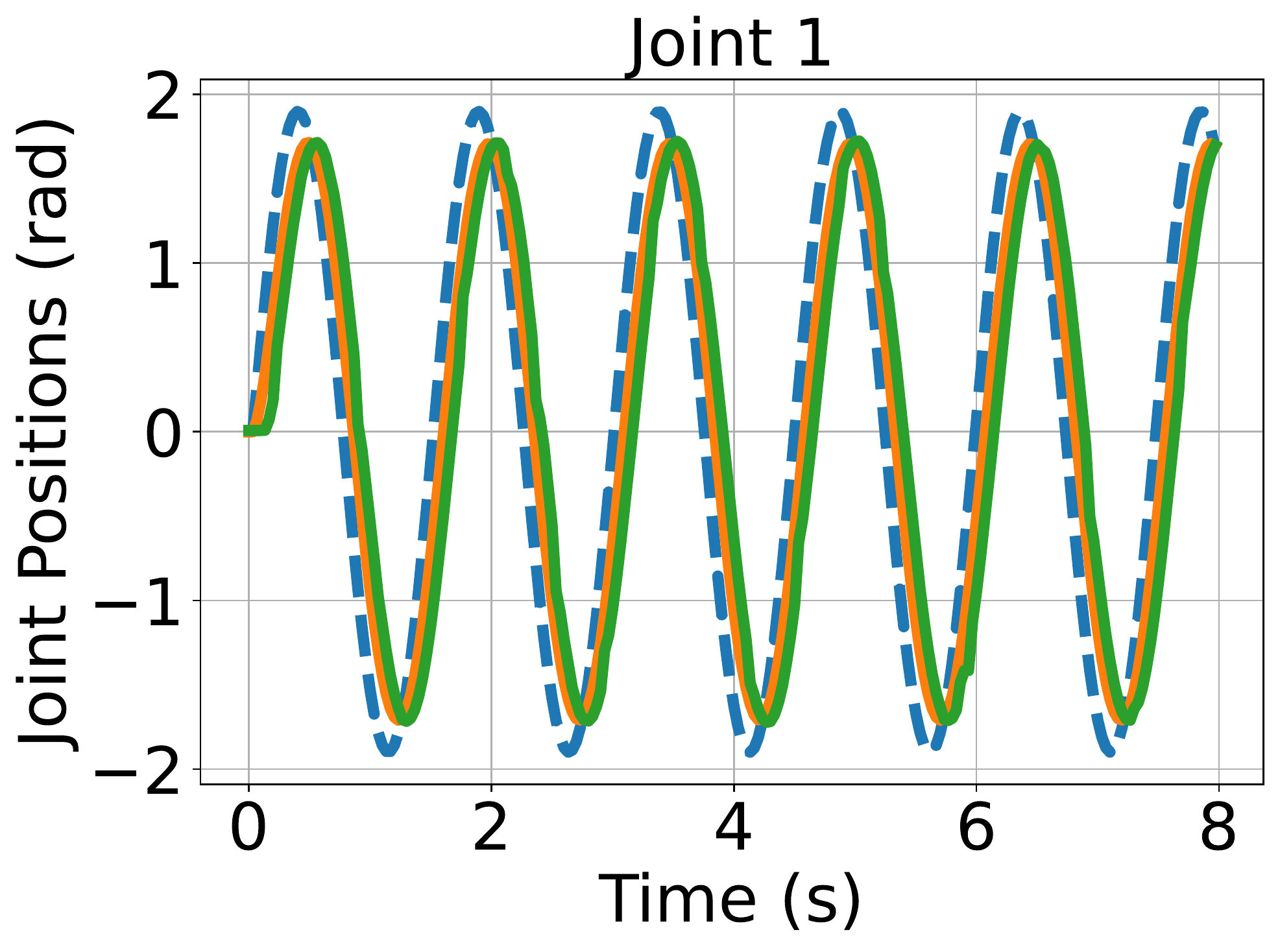}
     \end{subfigure}
\\
    \begin{subfigure}[b]{0.3\linewidth}
         \centering
         \includegraphics[width=\linewidth]{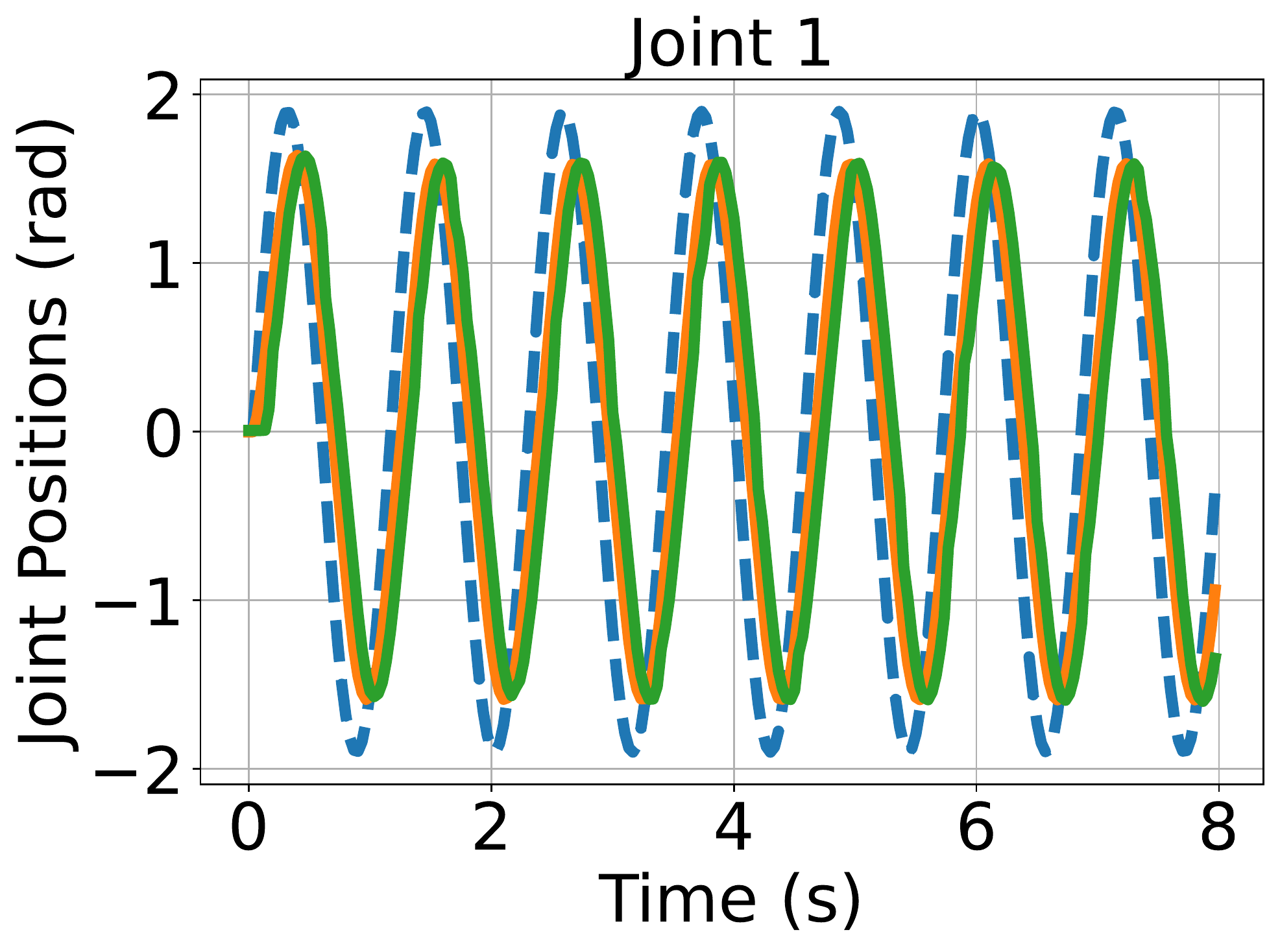}
     \end{subfigure}%
     \quad
    \begin{subfigure}[b]{0.3\linewidth}
         \centering
         \includegraphics[width=\linewidth]{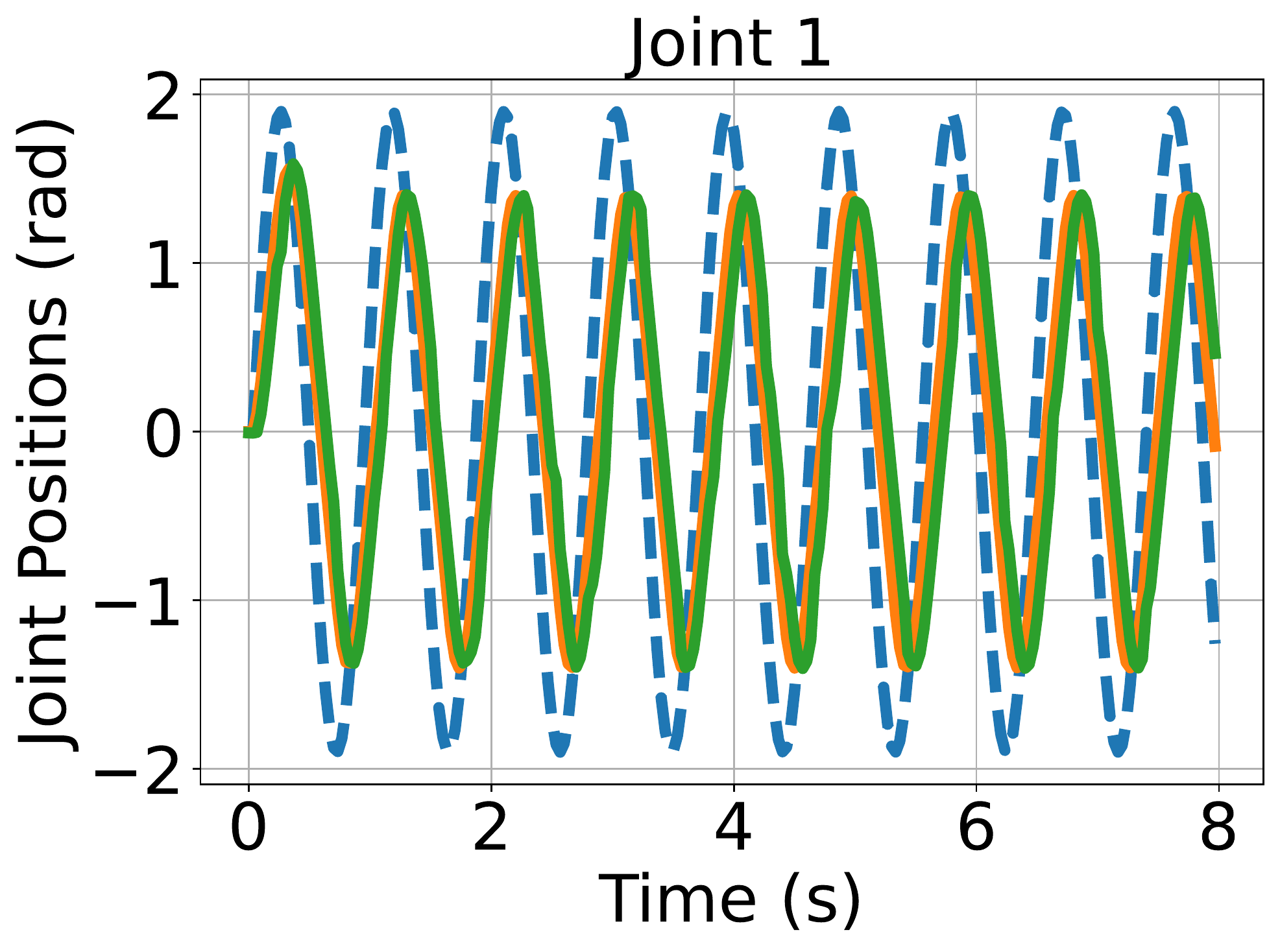}
     \end{subfigure}\\
         \begin{subfigure}[b]{0.3\linewidth}
         \centering
         \includegraphics[width=\linewidth]{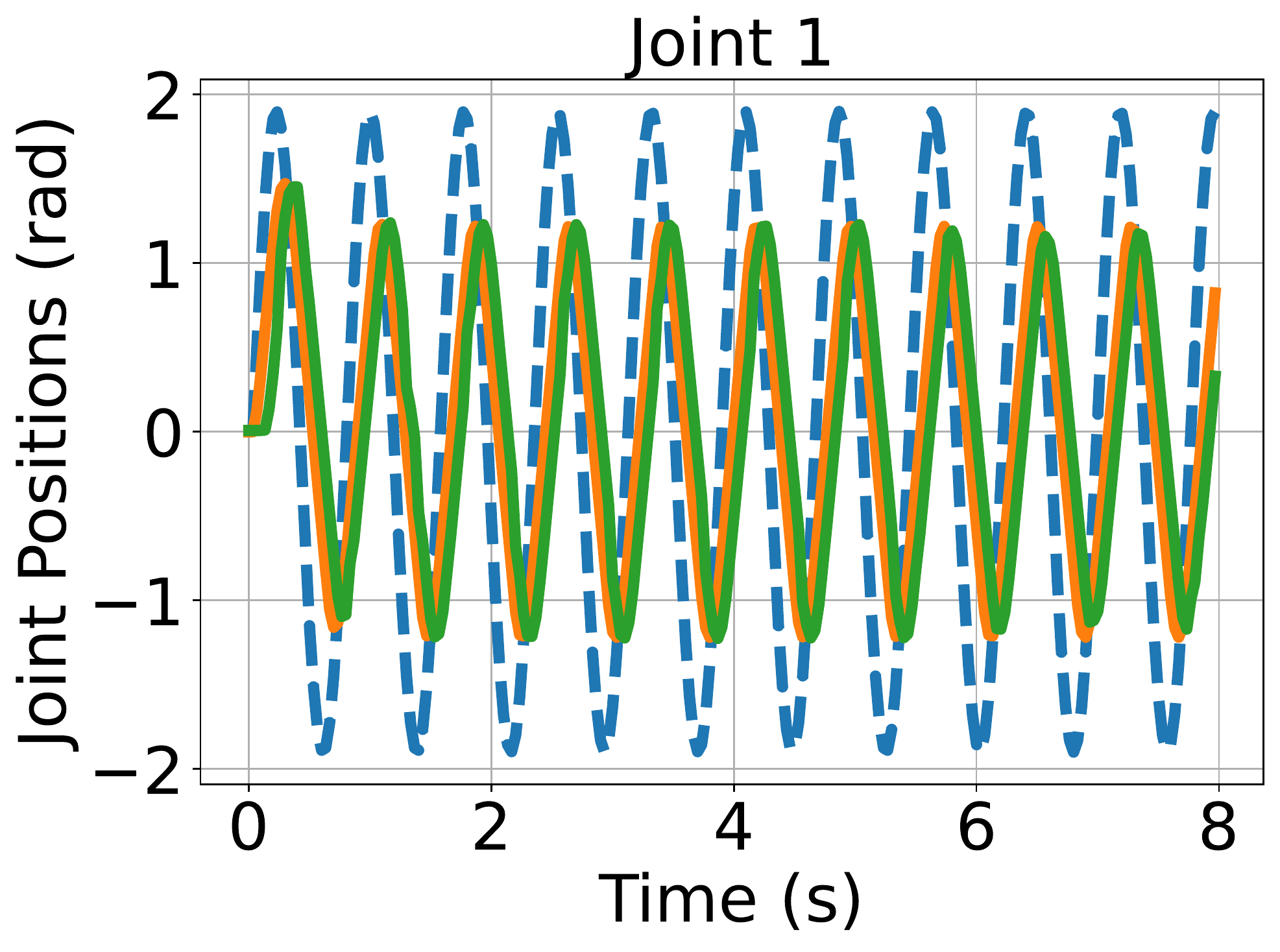}
     \end{subfigure}%
     \quad
     \begin{subfigure}[b]{0.3\linewidth}
         \centering
         \includegraphics[width=\linewidth]{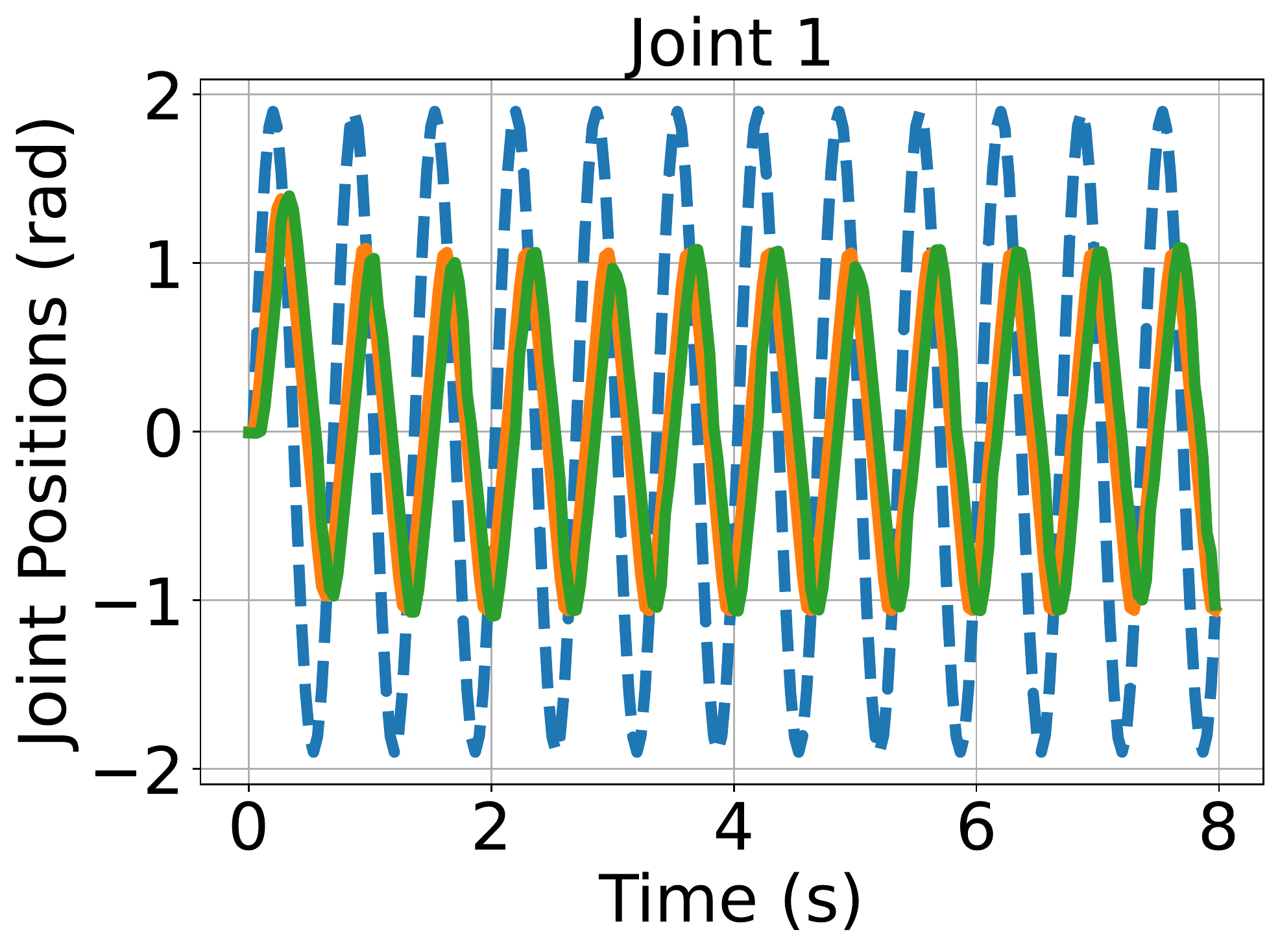}
     \end{subfigure}
    \caption{\textbf{Joint response curves}. Dynamics identification on a middle joint of one finger.}
    \label{fig:sys_id_jnt1}
\end{figure}

\begin{figure}[!htb]
    \centering
    \begin{subfigure}[b]{0.3\linewidth}
         \centering
         \includegraphics[width=\linewidth]{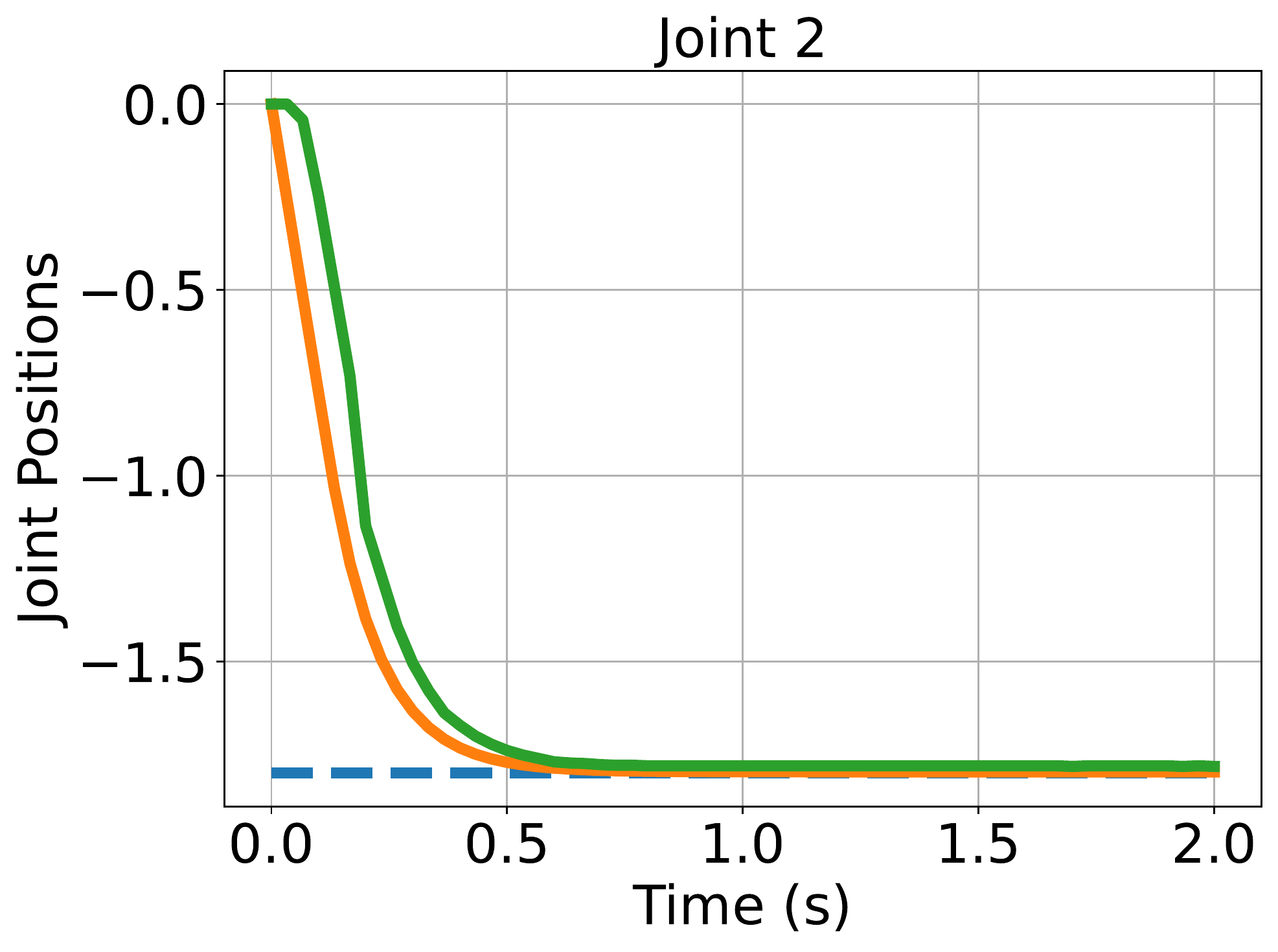}
     \end{subfigure}%
     \quad
     \begin{subfigure}[b]{0.3\linewidth}
         \centering
         \includegraphics[width=\linewidth]{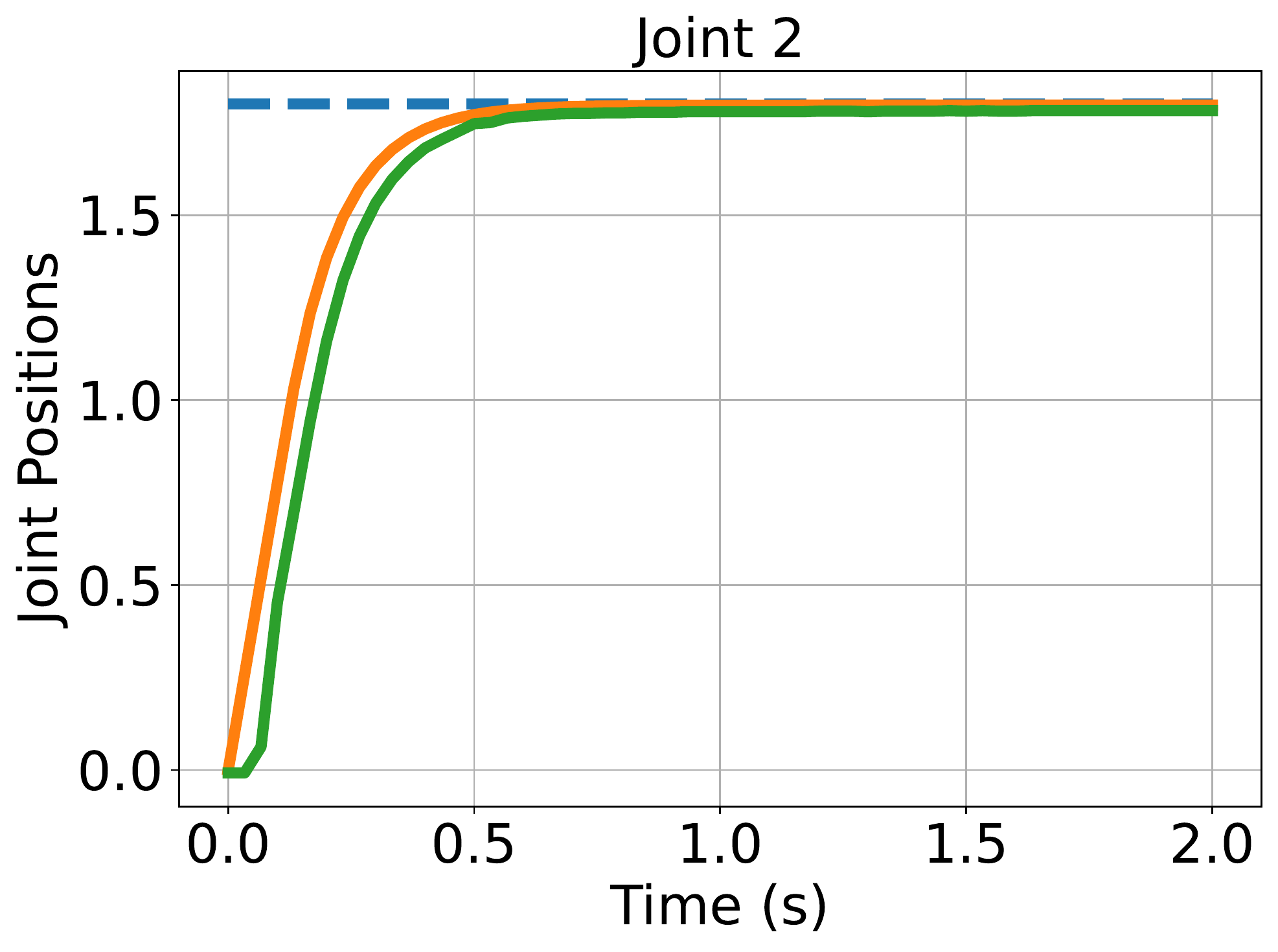}
     \end{subfigure}\\
         \begin{subfigure}[b]{0.3\linewidth}
         \centering
         \includegraphics[width=\linewidth]{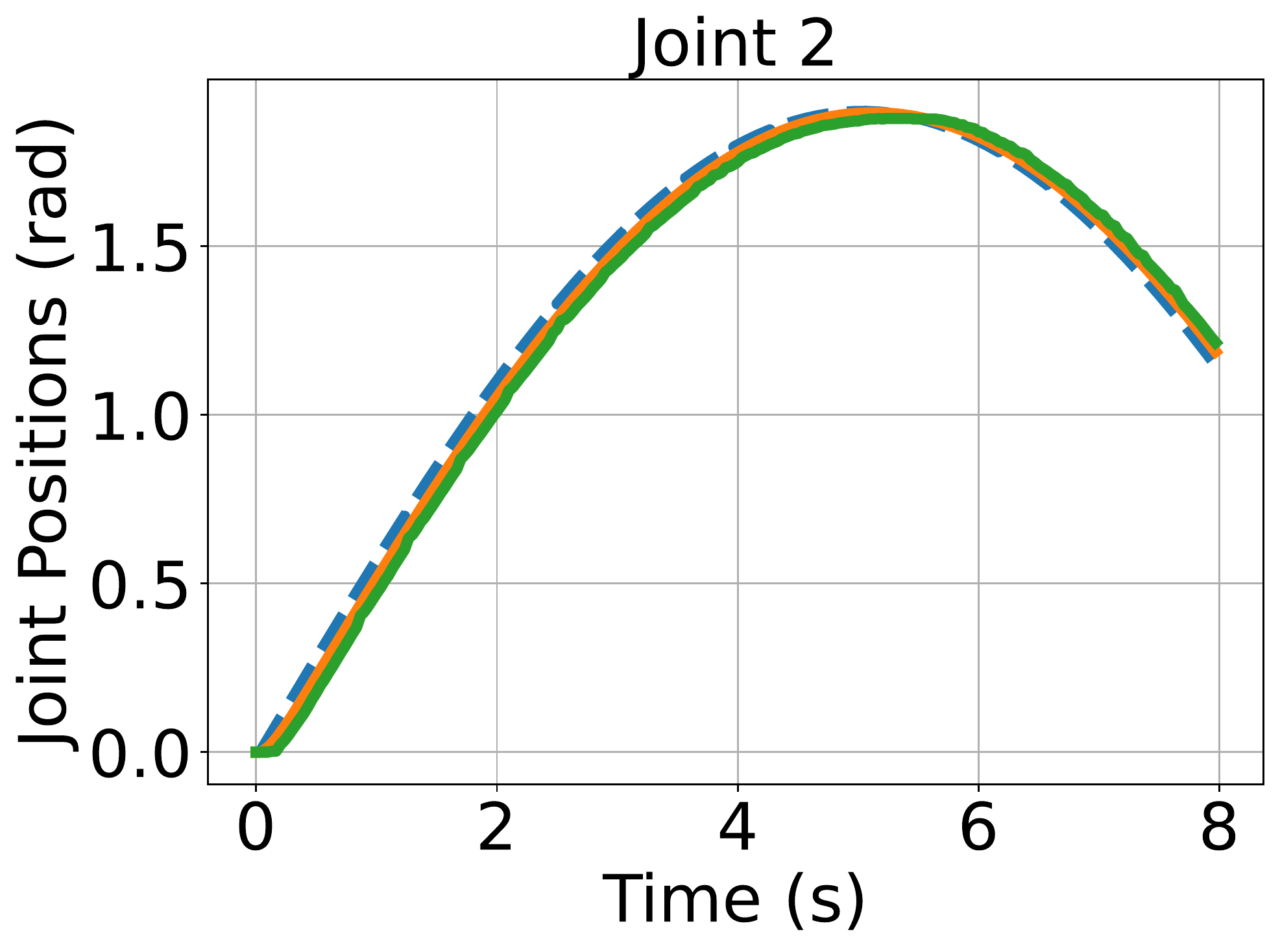}
     \end{subfigure}%
     \quad
     \begin{subfigure}[b]{0.3\linewidth}
         \centering
         \includegraphics[width=\linewidth]{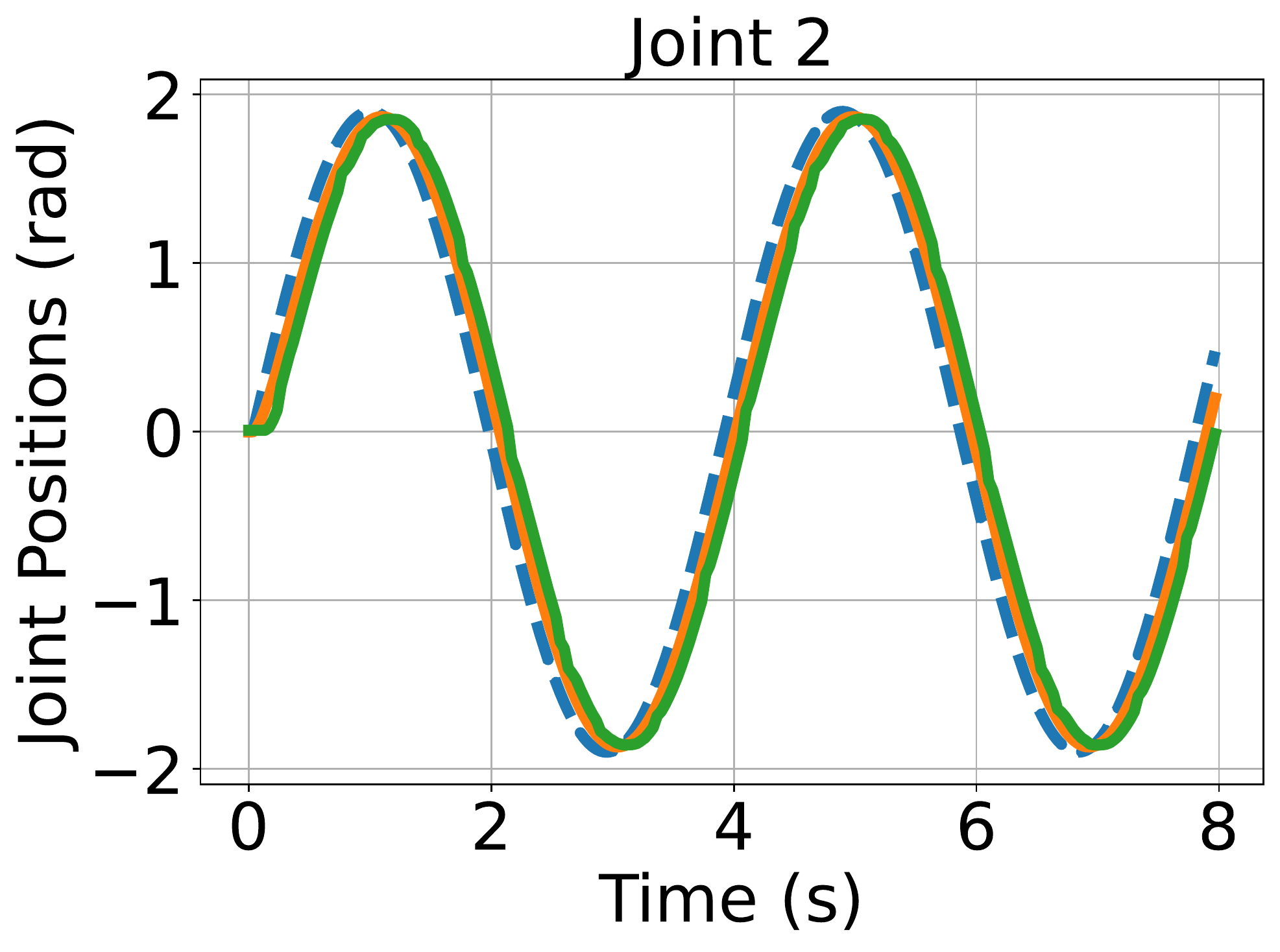}
     \end{subfigure}\\
    \begin{subfigure}[b]{0.3\linewidth}
         \centering
         \includegraphics[width=\linewidth]{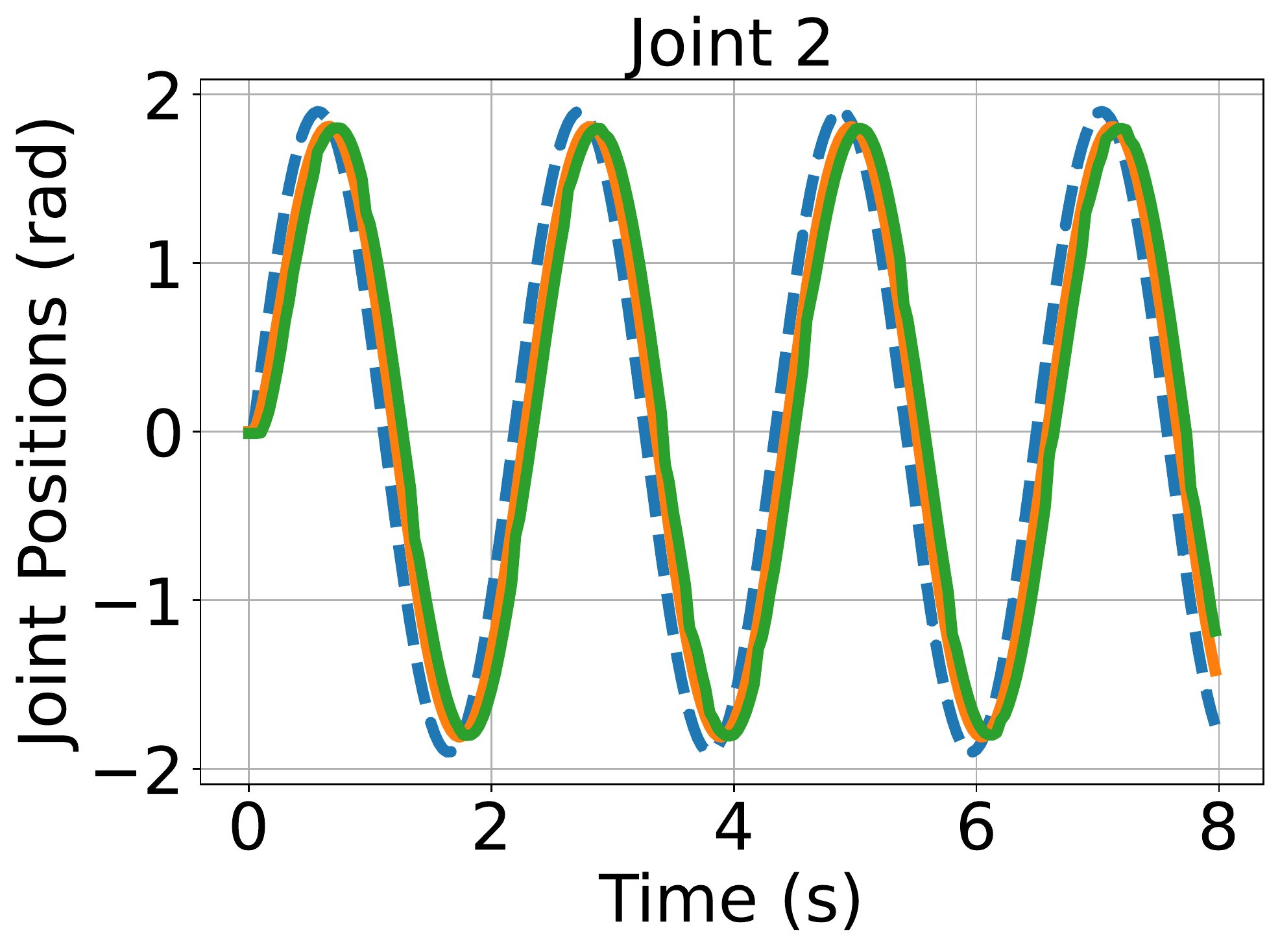}
     \end{subfigure}%
     \quad
    \begin{subfigure}[b]{0.3\linewidth}
         \centering
         \includegraphics[width=\linewidth]{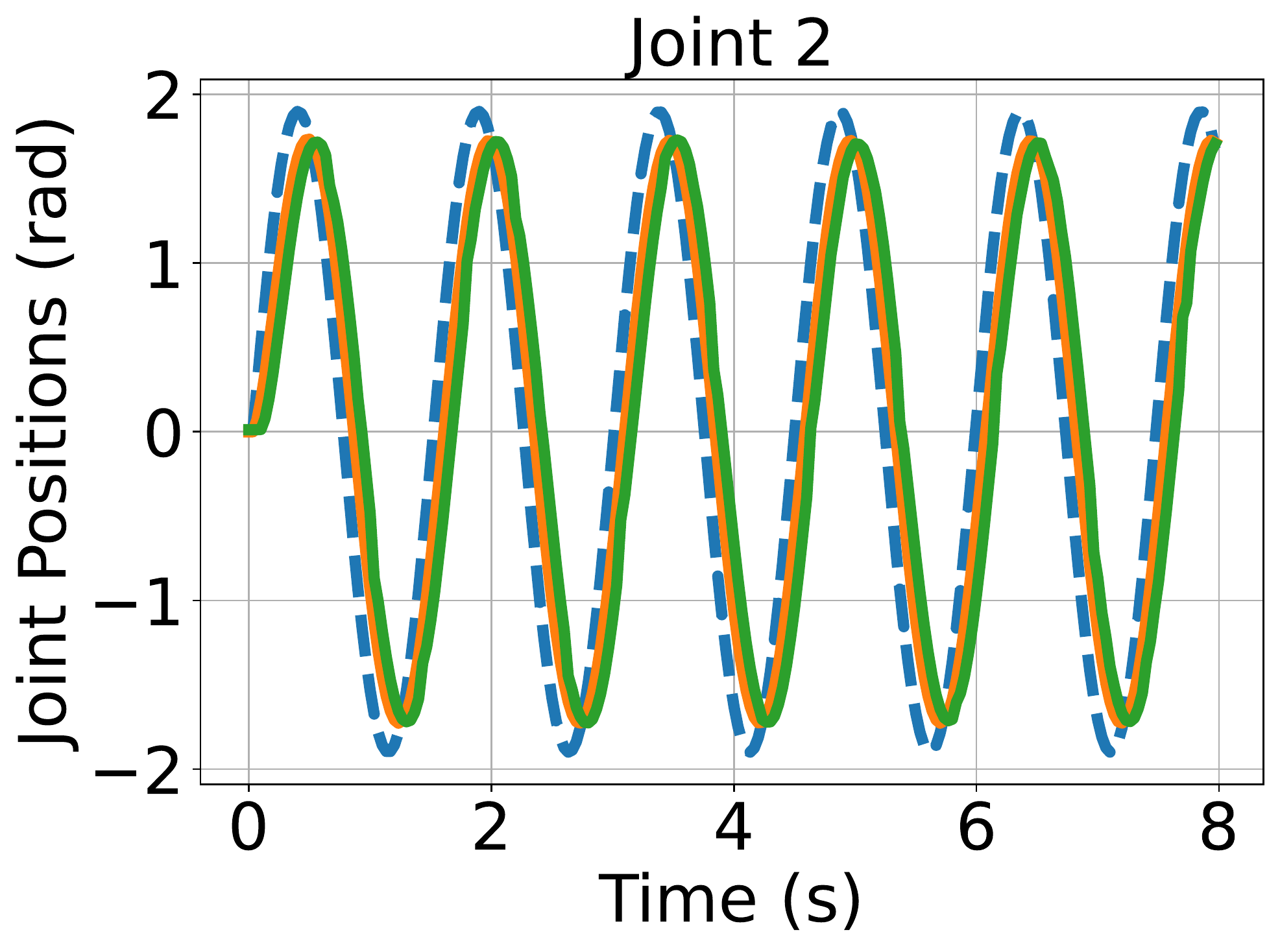}
     \end{subfigure}
\\
    \begin{subfigure}[b]{0.3\linewidth}
         \centering
         \includegraphics[width=\linewidth]{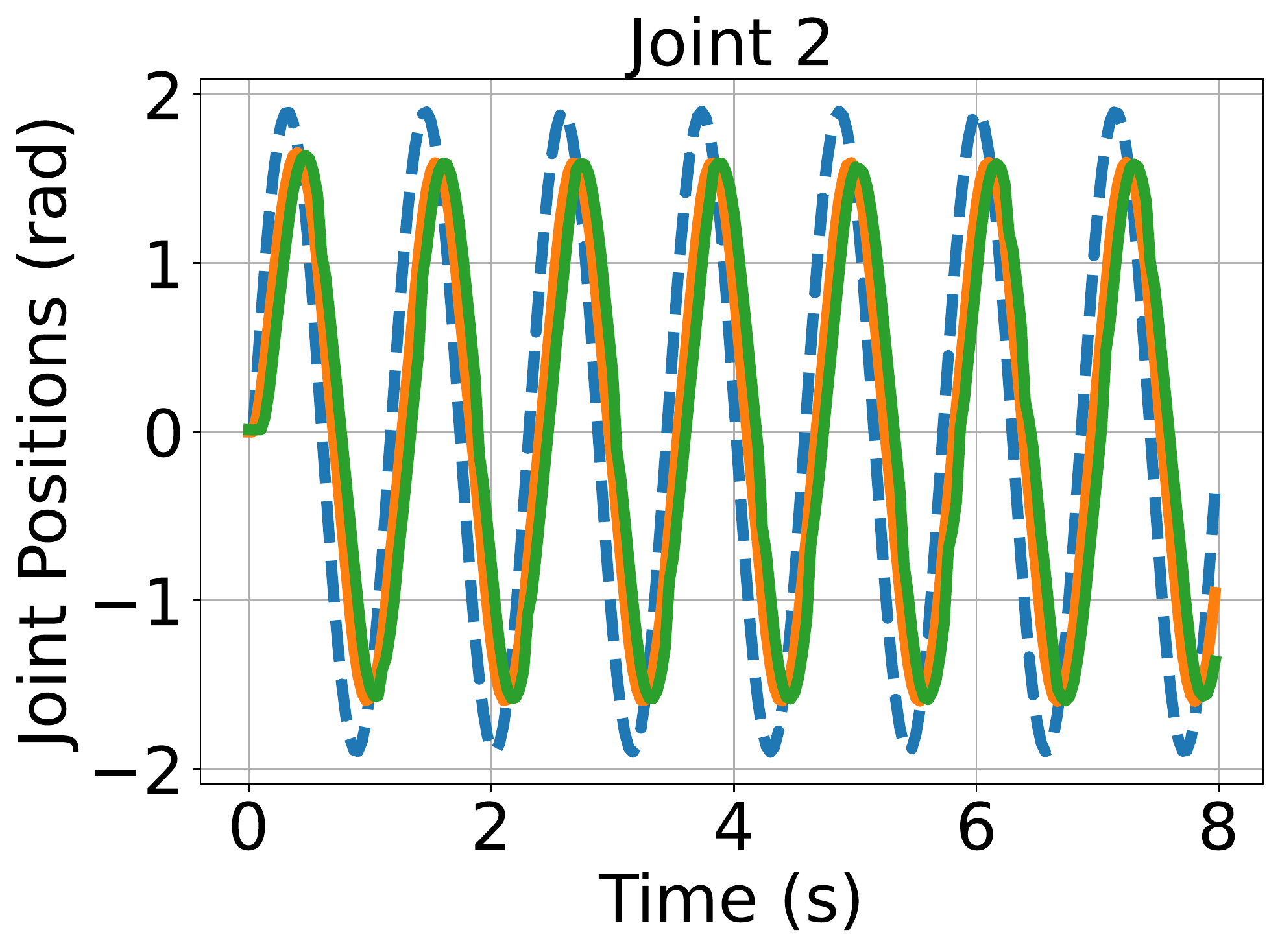}
     \end{subfigure}%
     \quad
    \begin{subfigure}[b]{0.3\linewidth}
         \centering
         \includegraphics[width=\linewidth]{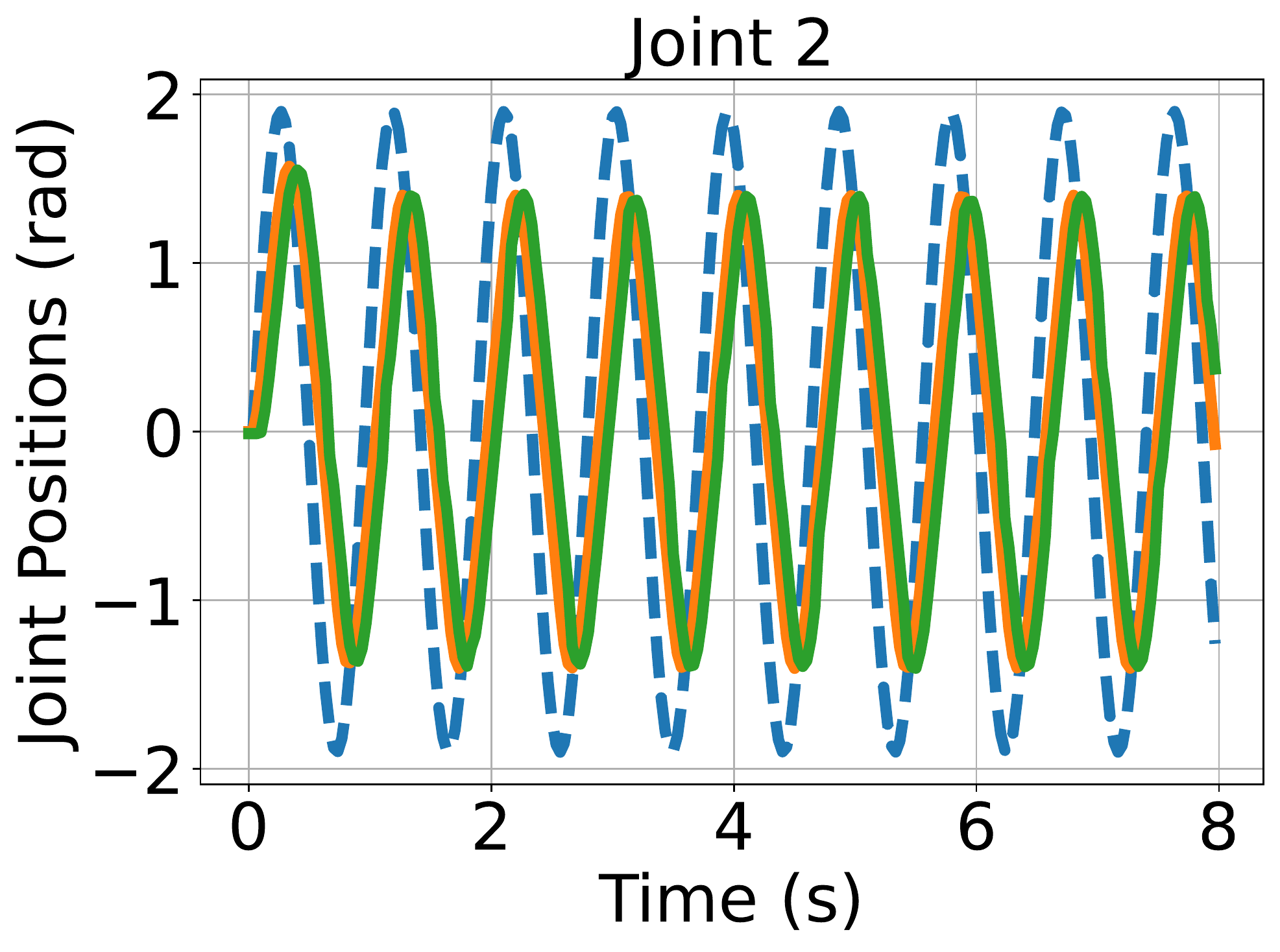}
     \end{subfigure}\\
         \begin{subfigure}[b]{0.3\linewidth}
         \centering
         \includegraphics[width=\linewidth]{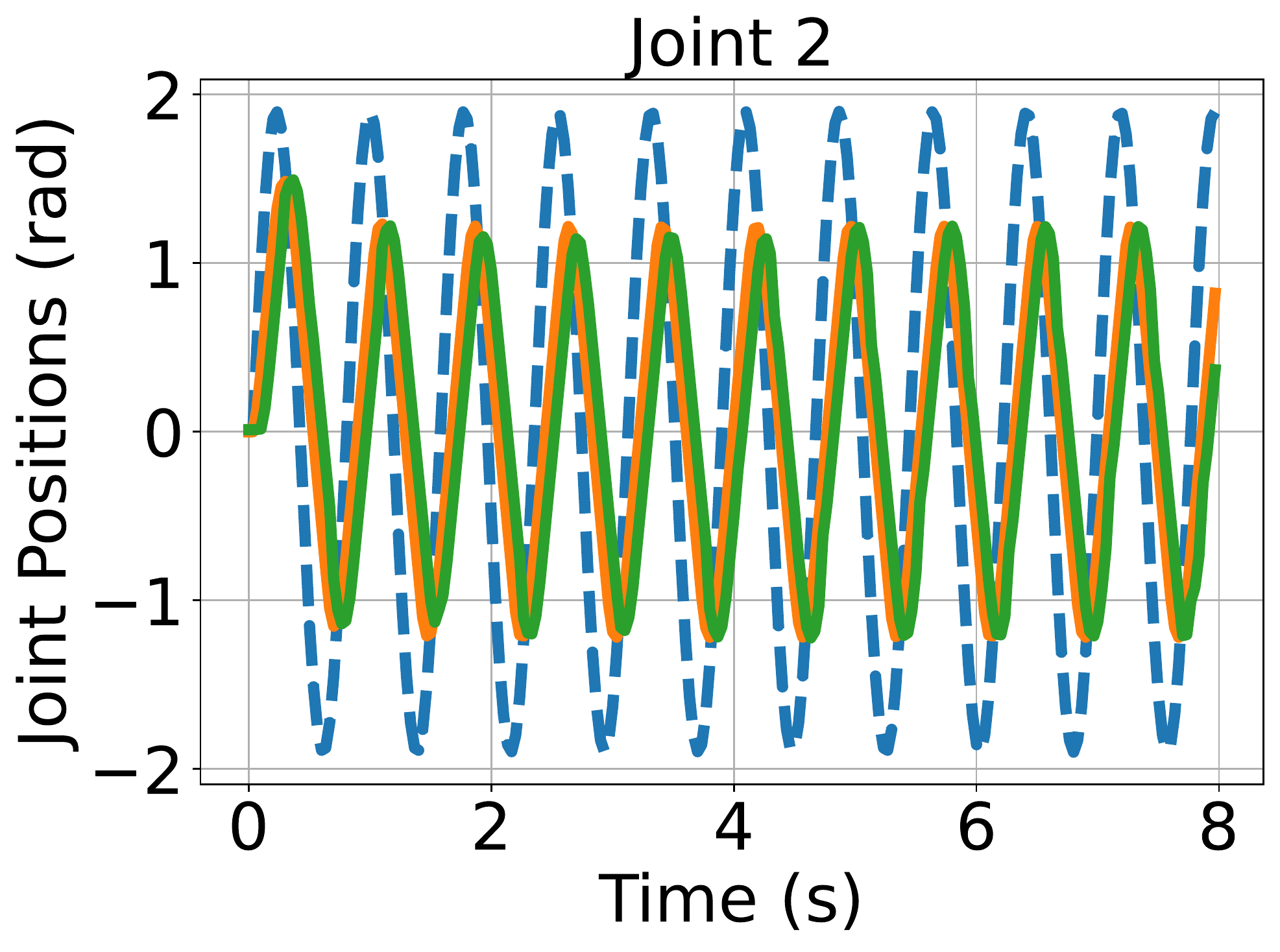}
     \end{subfigure}%
     \quad
     \begin{subfigure}[b]{0.3\linewidth}
         \centering
         \includegraphics[width=\linewidth]{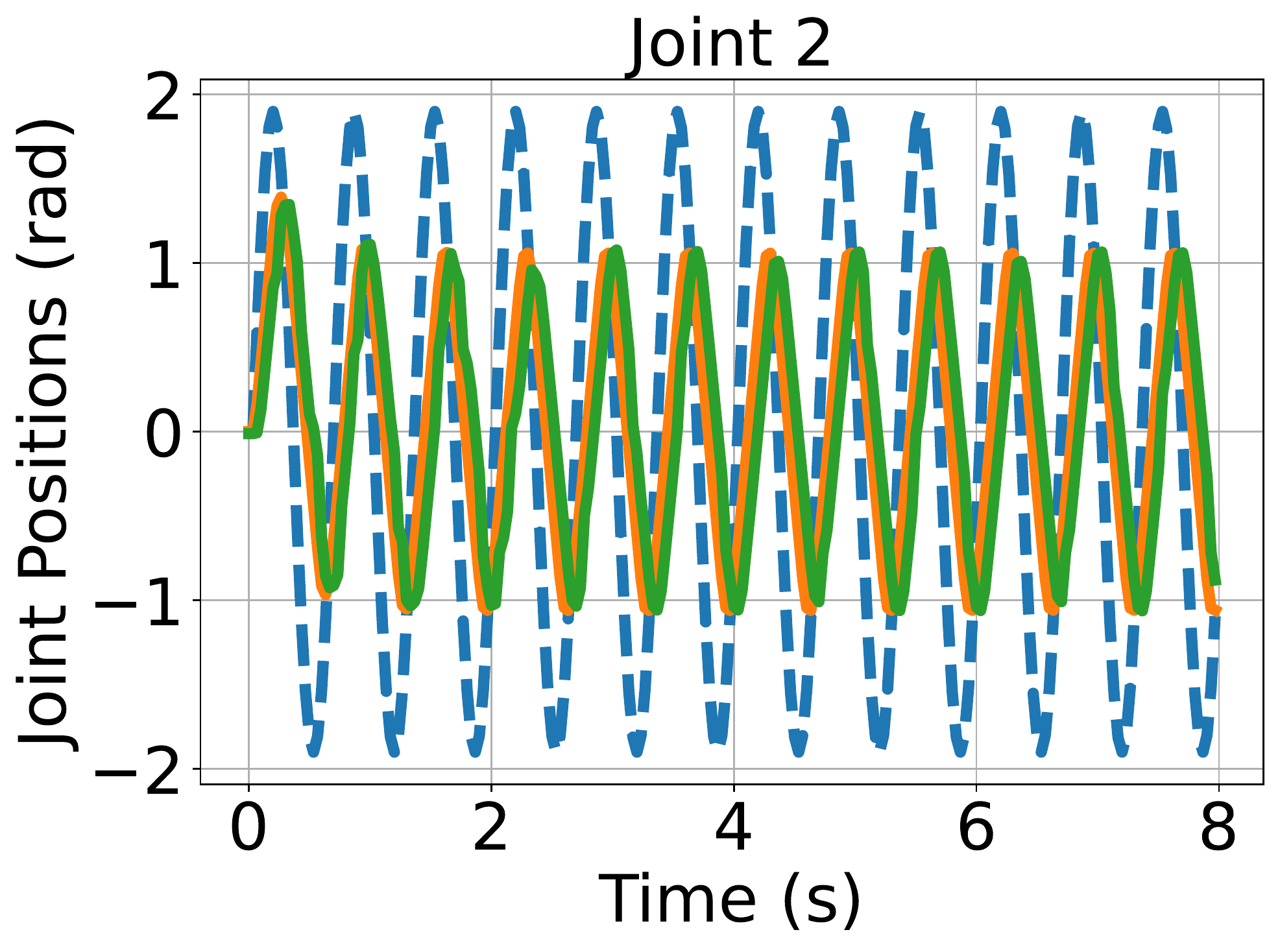}
     \end{subfigure}
    \caption{\textbf{Joint response curves}. Dynamics identification on a bottom joint of one finger.}
    \label{fig:sys_id_jnt2}
\end{figure}

\paragraph{Dynamics parameters to be identified}
We perform dynamics identification on the joint stiffness, damping, and velocity limit. Only one finger on the real robot hand is used to collect the response trajectories. Each joint on the finger is identified individually, and the same group of dynamics parameters of the finger is applied to the remaining fingers in the simulator. \figref{fig:sys_id_jnt0}, \figref{fig:sys_id_jnt1}, and \figref{fig:sys_id_jnt2} show that the simulated joints behave similarly to the real joints given the same control commands.

\paragraph{Response curves} In \figref{fig:sys_id_jnt0}, \figref{fig:sys_id_jnt1}, and \figref{fig:sys_id_jnt2}, we show the response curves of the three joints on a finger given a sequence of action commands both in simulation and in the real world. We can see that after the dynamics identification, the simulated joints can give a similar response as the real joints. We also observe that the real joints (the orange lines) usually have a slightly slower response than the simulated joints (the green lines). This is due to the latency of a real robot hand system. We did not model the latency in simulation and found that our controller still works on the real robot hands. It is possible that including the latency in simulation might further improve the controller's real-world performance, for which we leave the investigation to future work.

\subsubsection*{Robust policy learning}
\label{app_subsec:robust_policy}

\paragraph{Observation and action noise}
We add Gaussian noise to the action commands (\tblref{tbl:dyn_ran_params}) for training all policies. The teacher policy is trained with state noise as detailed in \tblref{tbl:dyn_ran_params}. The vision policies are trained with data augmentation on the point cloud observation. With a probability of $p=0.4$, we add Gaussian noise $\gaussian(0,0.004)$ to the point positions. Independently, with a probability of $p=0.4$, we randomly drop out $q\in[0, 20]$ percent of the points.

\paragraph{Dynamics randomization} 
We train policies with small randomization in the joint dynamic parameters: link mass, joint friction, and joint damping. We add large randomization to the object dynamics parameters such as mass friction, and restitution. \tblref{tbl:dyn_ran_params} lists the amount of randomization we add to the dynamics parameters and the observation/action noise.

\paragraph{Disturbance force on the object}
With a probability $p=0.2$ at each time step, we apply a disturbance force with a magnitude of  $F^o_d=c_dm^o$ where $m^o$ is the object mass, $c_d$ is a coefficient and a random force direction sampled in the $SO(3)$ space.

\section*{Supplementary Discussion}
\subsection*{Student policy closely tracks the performance of teacher policy}
\label{app_subsec:student_track_teacher}
We evaluated our learned policies on the training object dataset ($\bigdataset$) and the testing object dataset ($\smalldataset$), respectively, in simulation. To characterize how well a policy behaves in the testing time, we use the empirical cumulative distribution function (ECDF) as the metric to measure the distribution of the errors ($\Delta\theta$). Figure 7D shows the ECDF curves for the teacher policy and student policies at stages 1 and 2, respectively. Using fully-observable low-dimensional state information, the teacher policy achieves the highest success rate at any error threshold. The student policies are able to track the teacher policy's performance closely.

\subsection*{Symmetric object reorientation}
\label{app_subsec:symmetric}
Learning visual policies to reorient symmetric objects is challenging because objects in different but symmetrical poses appear similar leading to multimodality\cite{chen2021system}: Given a target orientation, many different poses of a symmetric object match the goal configuration visually, leading to different but equally good action sequences. It is challenging to learn a stochastic vision policy that explicitly accounts for multi-modality. An alternative is to use implicit models~\cite{florence2022implicit}. However, these models are computationally inefficient, which prevents their use in a real-time controller. 

The key intuition behind how we overcome this problem is that we need to account for multi-modality only at training time to ensure that a correct action sequence is not incorrectly penalized. However, at deployment, it is not necessary to distinguish between modes, and reorienting the object into any of the equivalent symmetric configurations would suffice. We bypass the problem of accounting for multi-modality at training time by using only unimodal objects -- asymmetric or reflective-symmetric objects. Our hypothesis was that if we are able to successfully learn a controller that operates over a diverse range of asymmetric shapes, it may also generalize to symmetric objects. 

To verify if this hypothesis was true, we tested our controller on two symmetric objects (a rectangular cuboid and a cube) with table support. Figure 4F shows that our controller still works reasonably well. Nonetheless, in this case, the orientation error tends to be higher than non-symmetric objects (Figure 4E). Although it is hard to pinpoint whether this is due to the performance drop in predicting the actions or predicting when the goal orientation is reached, we believe the latter plays a bigger role.

\subsection*{Ablation on the reward terms}
To investigate how reward terms and their coefficients affect policy performance, we conducted an ablation study where we varied the values of $c_1, c_2,$ and $c_3$ in Equation \ref{eqn:rot_reward_sparse}. For the sake of brevity, we trained policies on a single object (object \#10) without domain randomization.

\begin{figure}[!tb]
    \centering
    \begin{subfigure}[t]{0.48\linewidth}
        \centering
        \includegraphics[width=0.99\linewidth]{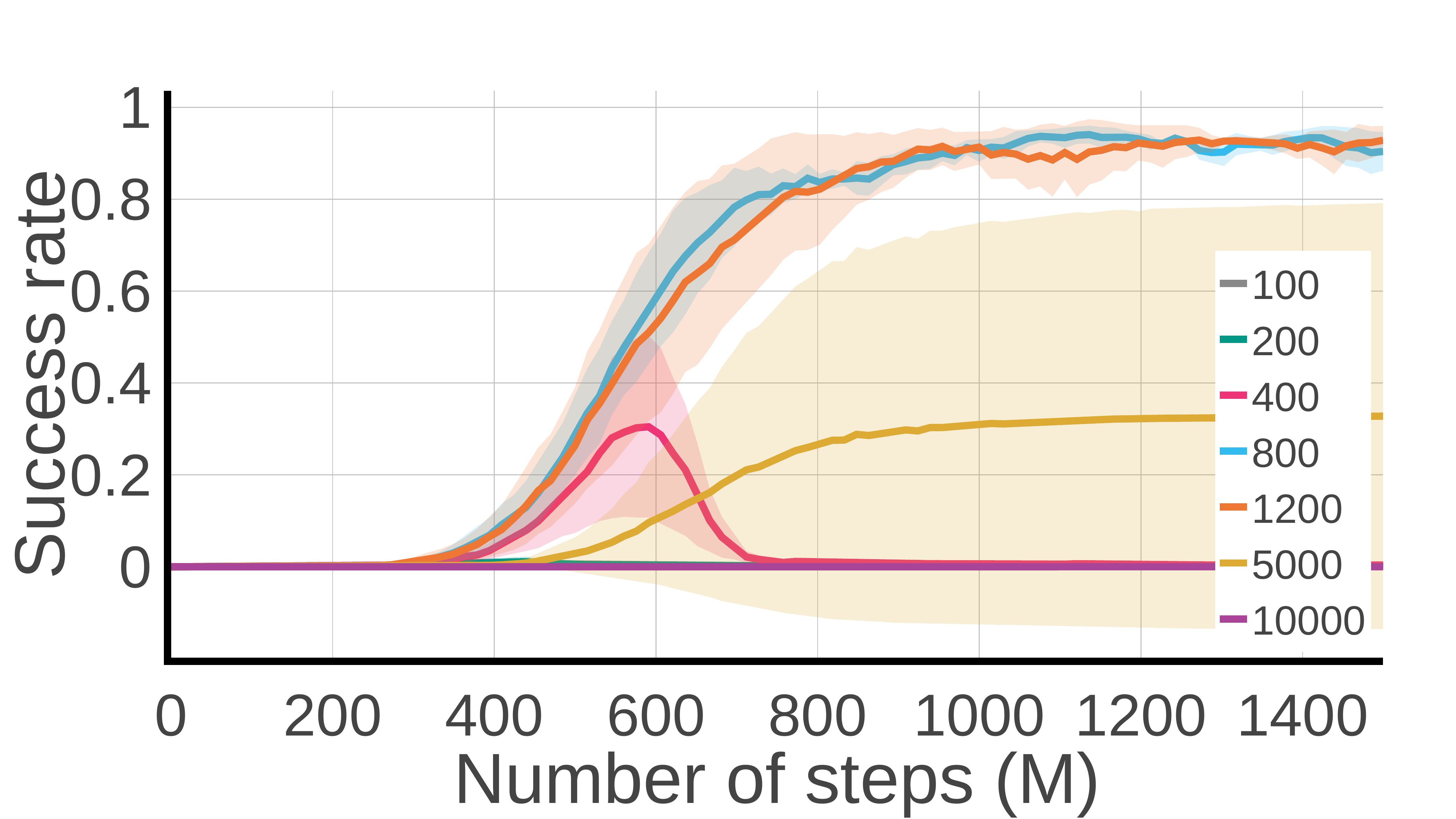}
        \caption{$c_1$}
        \label{fig:c1}
    \end{subfigure}%
        \hfill
    \begin{subfigure}[t]{0.48\linewidth}
        \centering
        \includegraphics[width=0.99\linewidth]{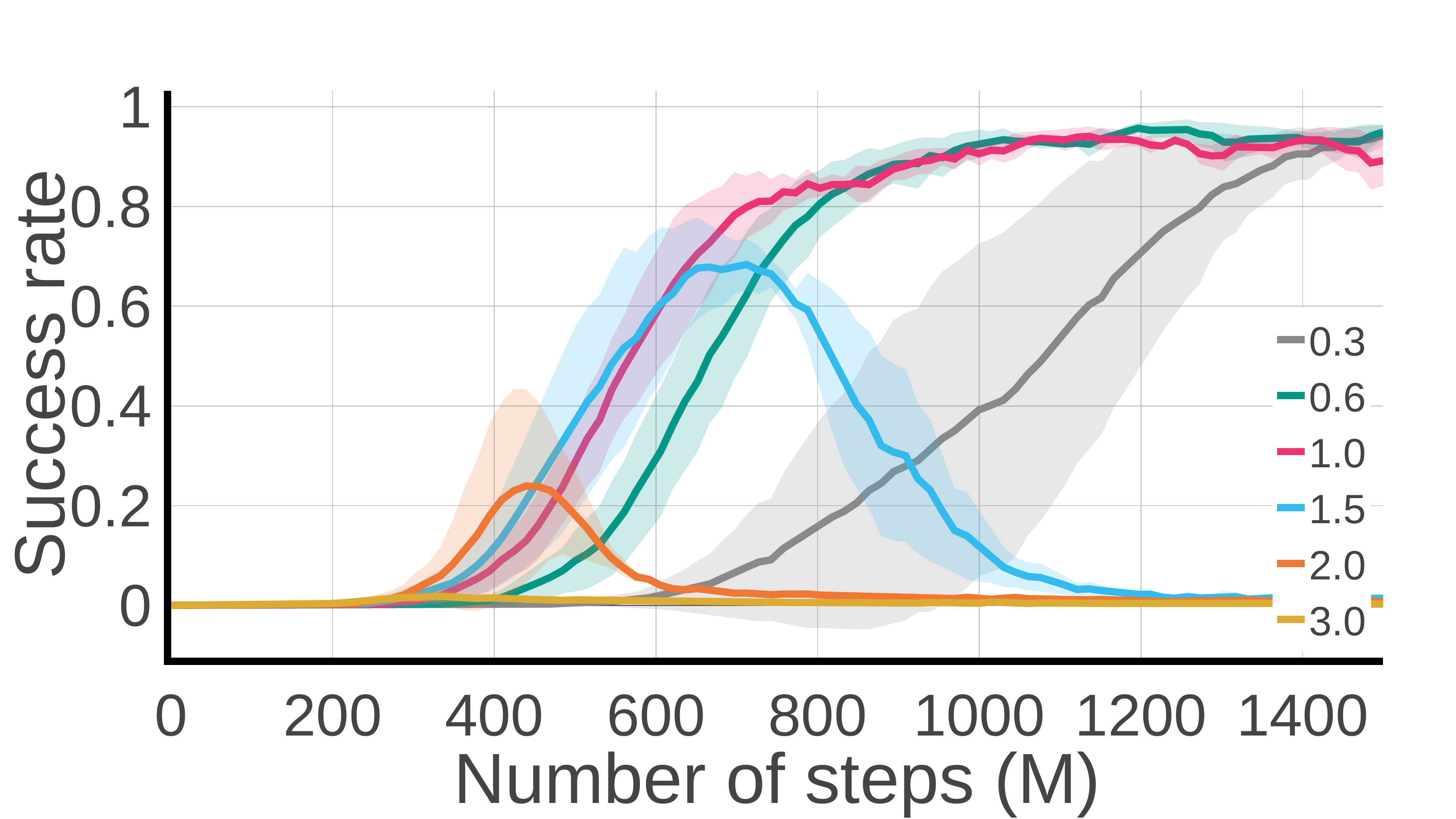}
        \caption{$c_2$}
        \label{fig:c2}
    \end{subfigure}\\
        \begin{subfigure}[t]{0.48\linewidth}
        \centering
        \includegraphics[width=0.99\linewidth]{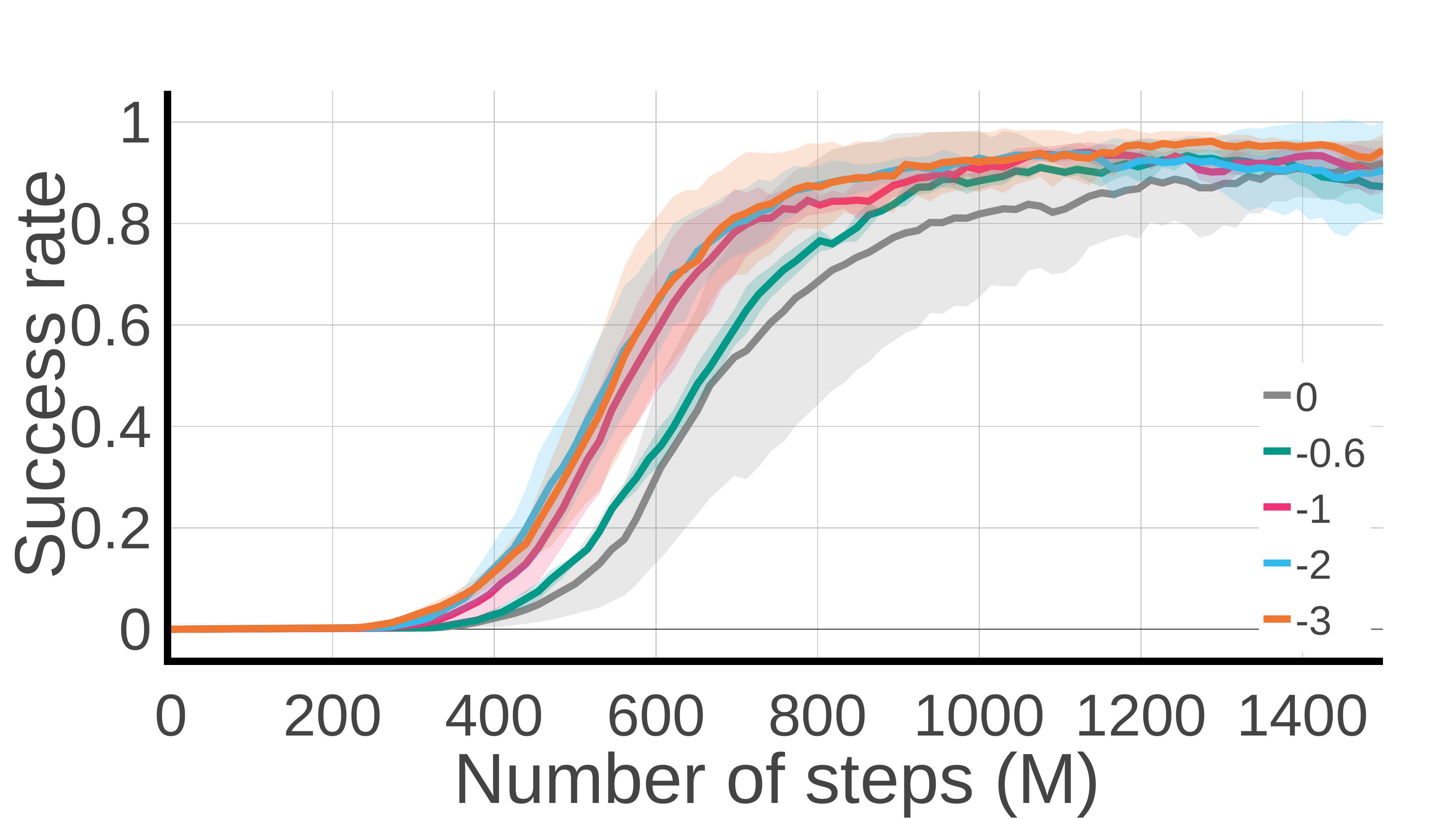}
        \caption{$c_3$}
        \label{fig:c3}
    \end{subfigure}%
        \hfill
    \begin{subfigure}[t]{0.48\linewidth}
        \centering
        \includegraphics[width=0.99\linewidth]{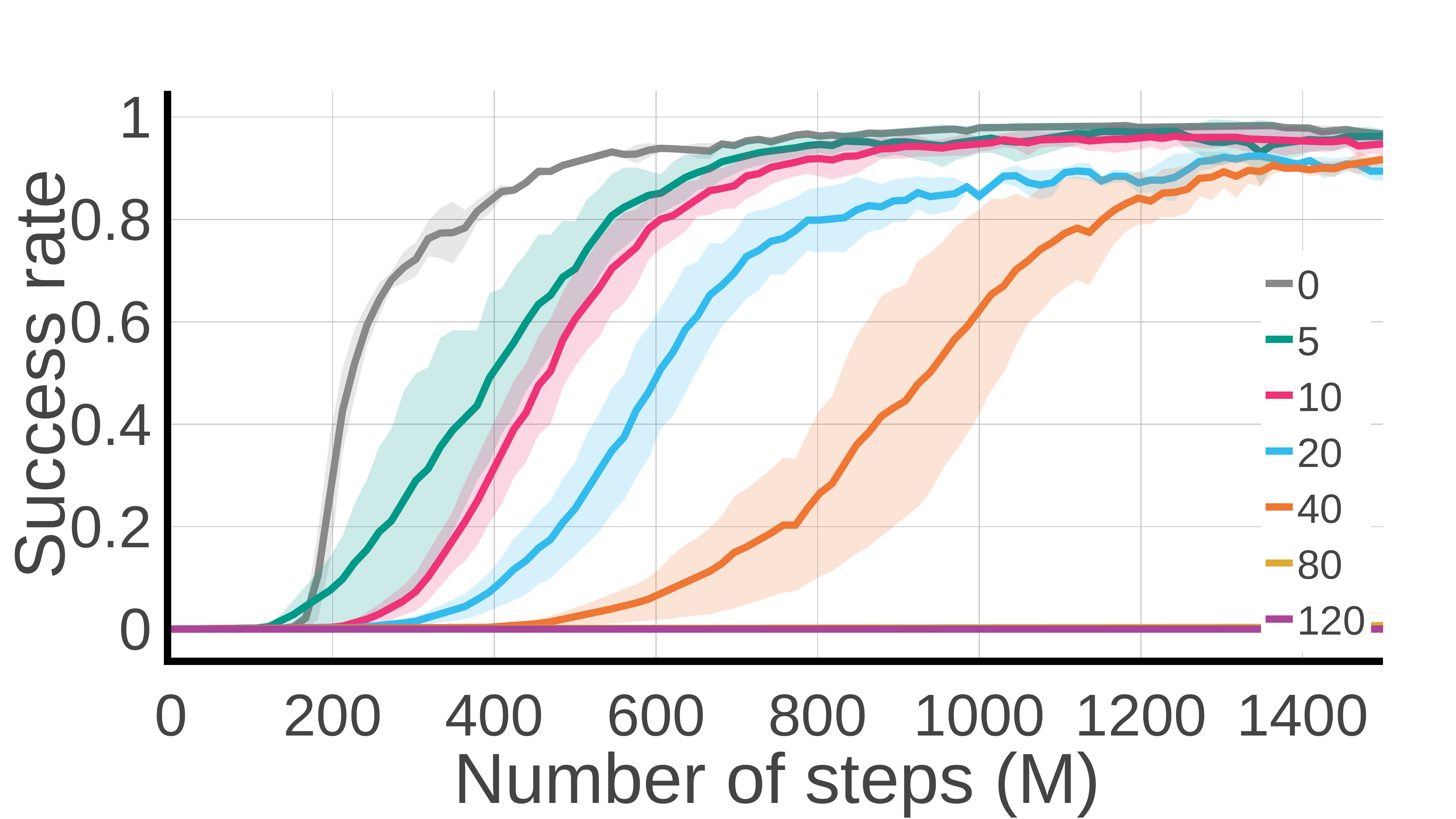}
        \caption{$c_4$}
        \label{fig:c4}
    \end{subfigure}\\
        \begin{subfigure}[t]{0.48\linewidth}
        \centering
        \includegraphics[width=0.99\linewidth]{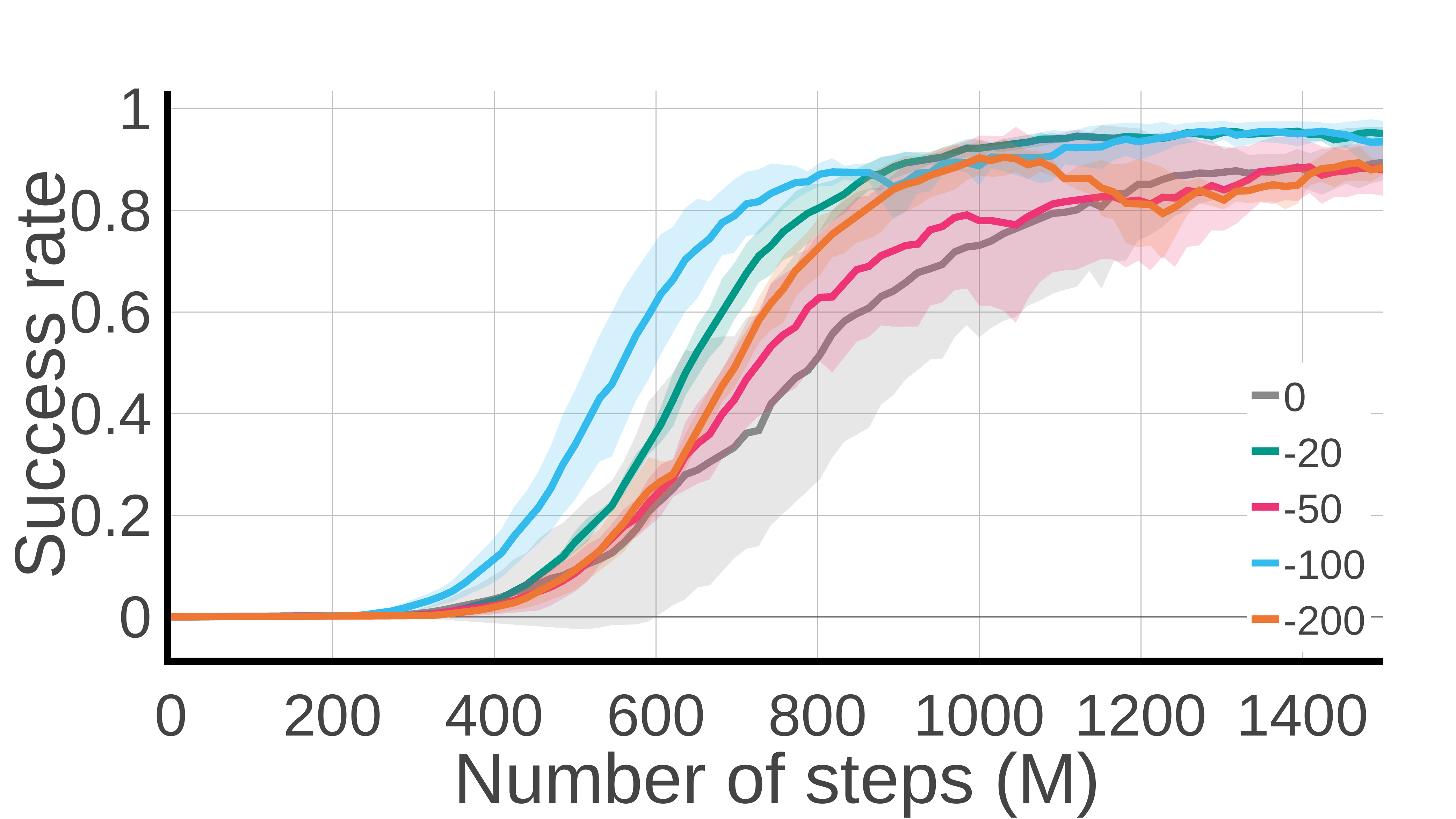}
        \caption{$c_5$}
        \label{fig:c5}
    \end{subfigure}%
    \caption{\textbf{Reward function ablation}. Different learning curves as we vary the values of $c_1$ \textbf{(a)}, $c_2$ \textbf{(b)}, $c_3$ \textbf{(c)}, $c_4$ \textbf{(d)}, $c_5$\textbf{(e)}.
 }
    \label{fig:reward_ab}
\end{figure}

As illustrated in \figref{fig:reward_ab}, increasing the value of $c_1$ led to an improvement in policy learning, but too large a value of $c_1$ resulted in a decline in performance. Similarly, increasing $c_2$ improved policy learning, but after $c_2=1.0$, the performance began to deteriorate. In contrast, policy learning was found to be less sensitive to $c_3$, which encourages fingers to remain near objects, and to $c_5$, which discourages fingers from pushing objects away. However, we did observe that having $c_3<0$ and $c_5<0$ was advantageous, as the learning curves exhibited substantially higher variance when $c_3=0$ and $c_5=0$. The fourth term ($c_4$) imposes an energy penalty to prevent excessive energy usage on the motors. As shown in \figref{fig:reward_ab}, a $c_4$ value of 0 allows the policy to learn the fastest, as there are no energy constraints. However, as $c_4$ increases, the learning speed decreases. When $c_4$ becomes too large, the learning process begins to fail.

A contact penalty term ($c_6$, Equation \ref{eqn:reward_table_penalty}) promotes in-air object reorientation by reducing table dependency. Policies trained with the penalty achieved $87\%$ in-air success, as those without achieved only $4.1\%$. This clearly shows that the contact penalty term benefits the in-air reorientation.

\subsection*{Using a different encoder for goal}
\label{appsubsec:diff_goal_encoder}

\begin{figure}[!htb]
    \centering
    \begin{subfigure}[t]{0.48\linewidth}
        \centering
        \includegraphics[width=\linewidth]{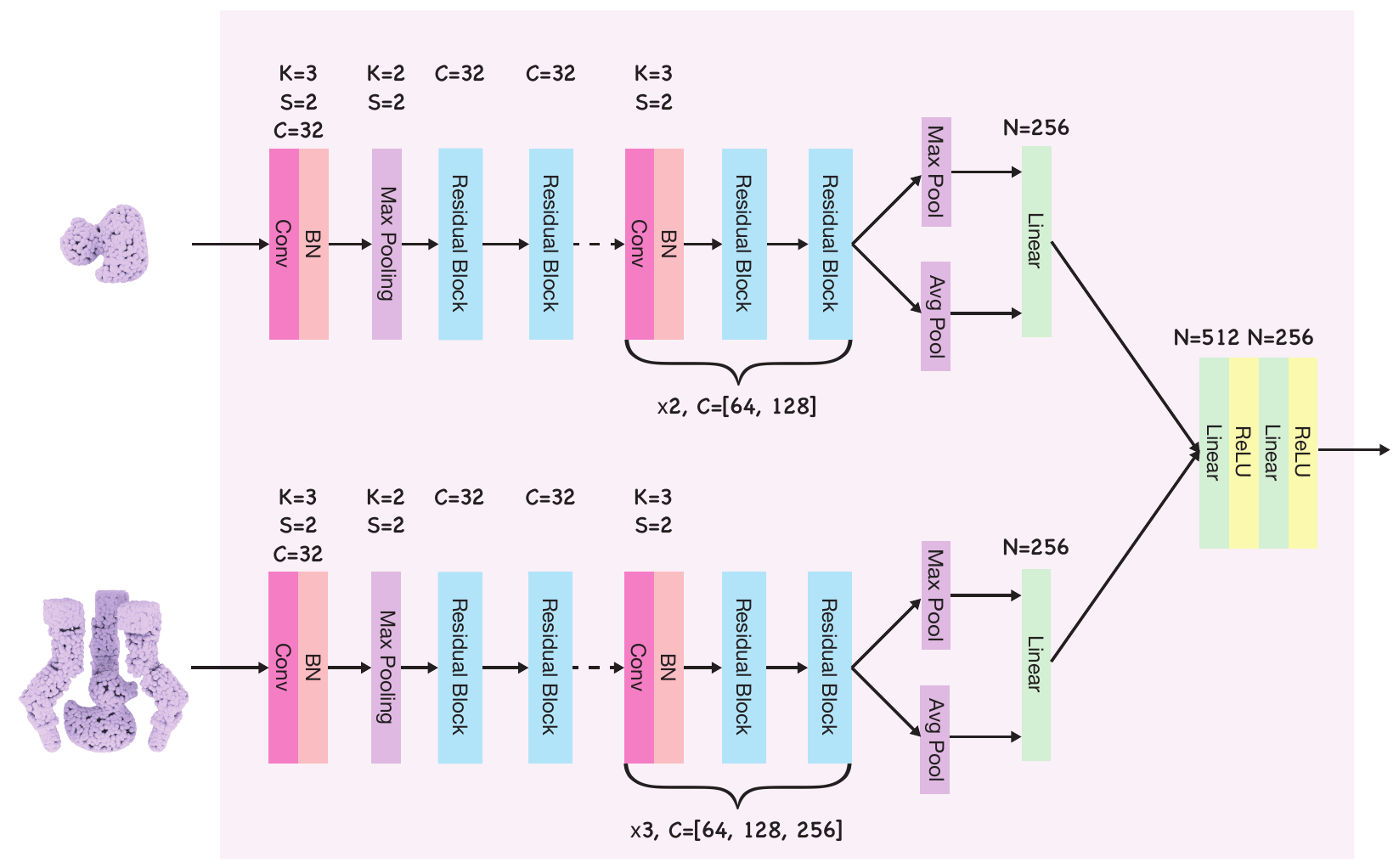}
        \caption{}
        \label{fig:separate_goal_cnn}
    \end{subfigure}%
    \hfill
    \begin{subfigure}[t]{0.48\linewidth}
        \centering
        \includegraphics[width=\linewidth]{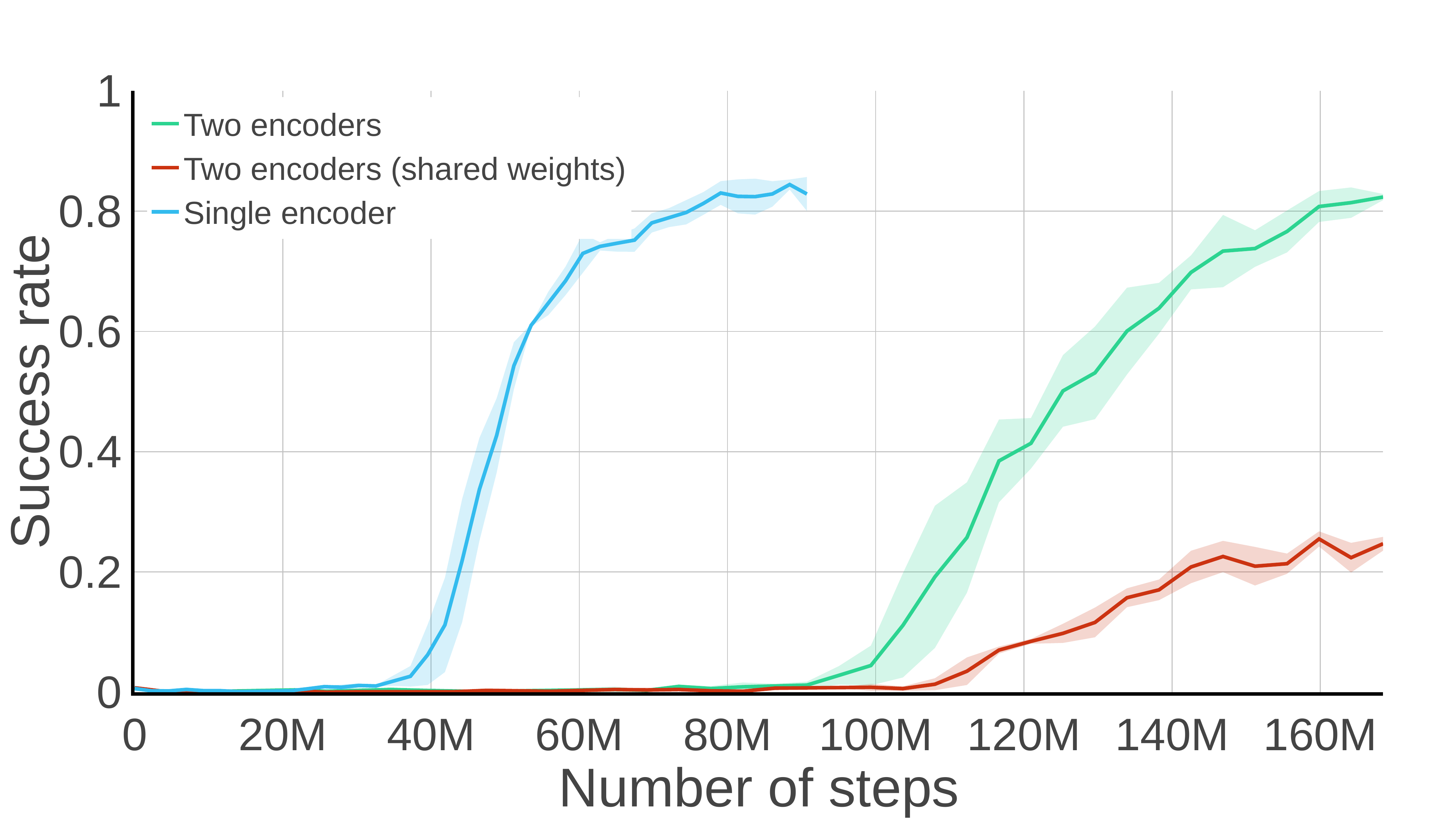}
        \caption{}
        \label{fig:separate_goal_curve}
    \end{subfigure}
    \caption{\textbf{Encoder architectures}. \textbf{(a)}: we tried using separate encoders for the goal point cloud and the scene point cloud. \textbf{(b)} shows that using separate encoders leads to considerably slower policy learning than using a single encoder on merged goal and scene point clouds.}
    \label{fig:separate_goal}
\end{figure}

As shown in \figref{fig:separate_goal_curve}, we also found that stacking the goal object point cloud onto the scene point cloud and feeding it as a whole into the 3D CNN encoder (Figure 7B) leads to faster policy learning than using two separate 3D CNN encoders (\figref{fig:separate_goal_cnn}) to process the scene point cloud and the goal object point cloud.

\subsection*{Stage 0: speeding up vision policy training with visual pre-training}
\label{appsubsec:stage_0}
\begin{figure}[!htb]
    \centering
    \includegraphics[width=\linewidth]{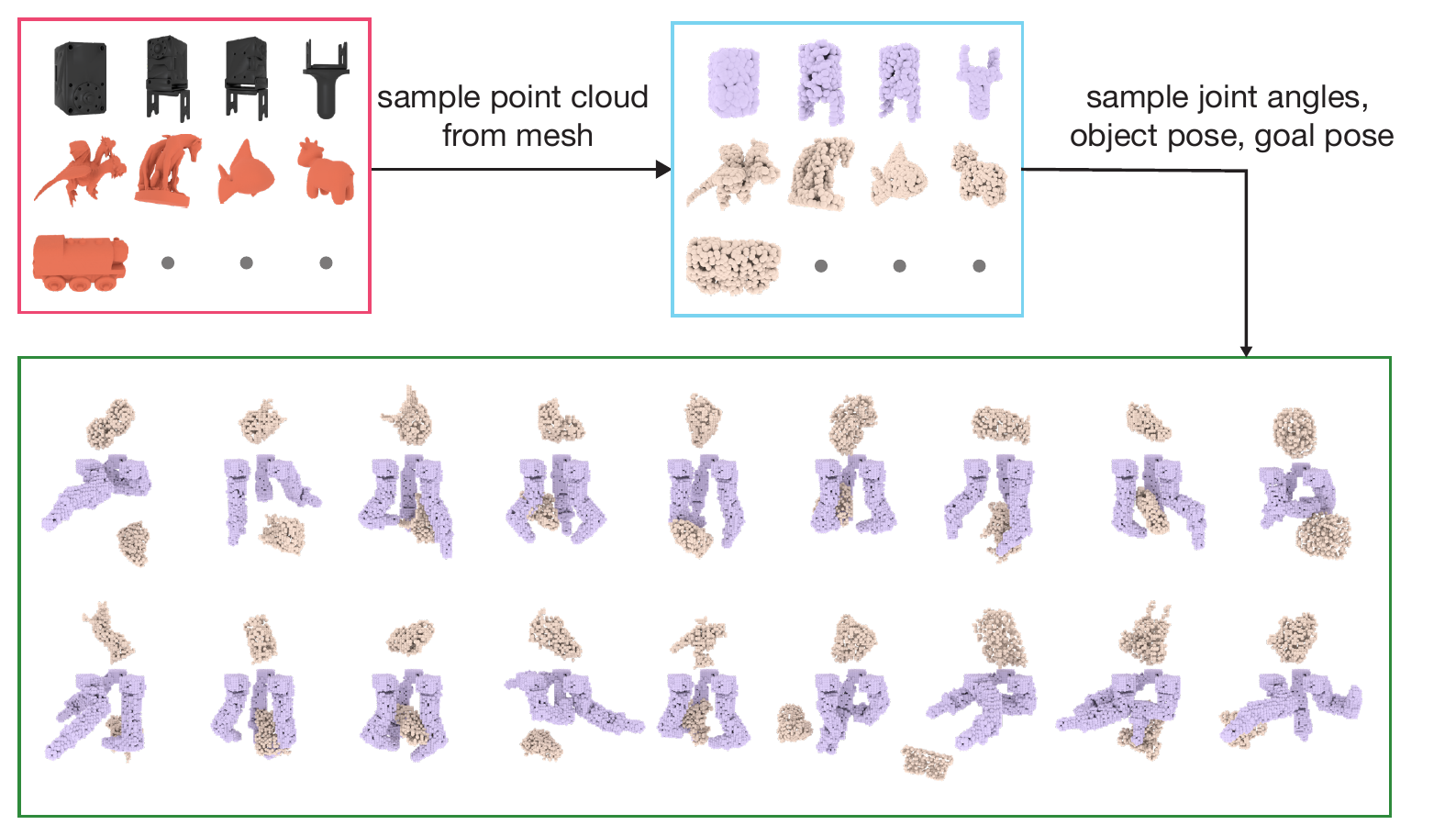}
    \caption{\textbf{Synthetic data}. The synthetic data generation in Stage 0.}
    \label{fig:synthetic_data}
\end{figure}

Can we further speed up the vision policy learning? As shown in Figure 7A, our policy network is a recurrent network. Training a sequence model can take longer training time. We investigated whether we can first pre-train the vision network component (sparse 3D CNN) in the policy network without involving the RNN component (Stage 0), and then fine-tune the policy with the pre-trained vision network (Stage 1). To pre-train the vision network, we explored various representation learning techniques, such as learning a forward/inverse dynamics model and reconstructing the input point cloud to pre-train the vision network. To our surprise, although most pre-training techniques lead to mild, if any, improvement in the policy learning speed, it's beneficial to first train the vision network to predict some low-dimensional state information of the system. More specifically, in the pre-training stage, the vision network is trained to predict the object category, the distance between the object's orientation and the goal orientation, and the joint positions of the robot hand ($\bm{q}_t$). Note that we do not predict object pose, which would require a definition of reference frame on the objects. To further speed up the training, we do not use simulation at all to generate the training data at this stage. Instead, we generate completely random synthetic data for training (\figref{fig:synthetic_data}). First, we convert meshes of the robot links and objects into their corresponding canonical point clouds. Next, we randomly sample joint angles and use forward kinematics to get the pose of each robot link and randomly sample object poses and goal poses. Finally, we transform the canonical point clouds of each part according to their poses and get the point cloud of the entire scene.

\begin{figure}[!htb]
    \centering
    \begin{subfigure}[t]{0.96\linewidth}
        \centering
        \includegraphics[width=\linewidth]{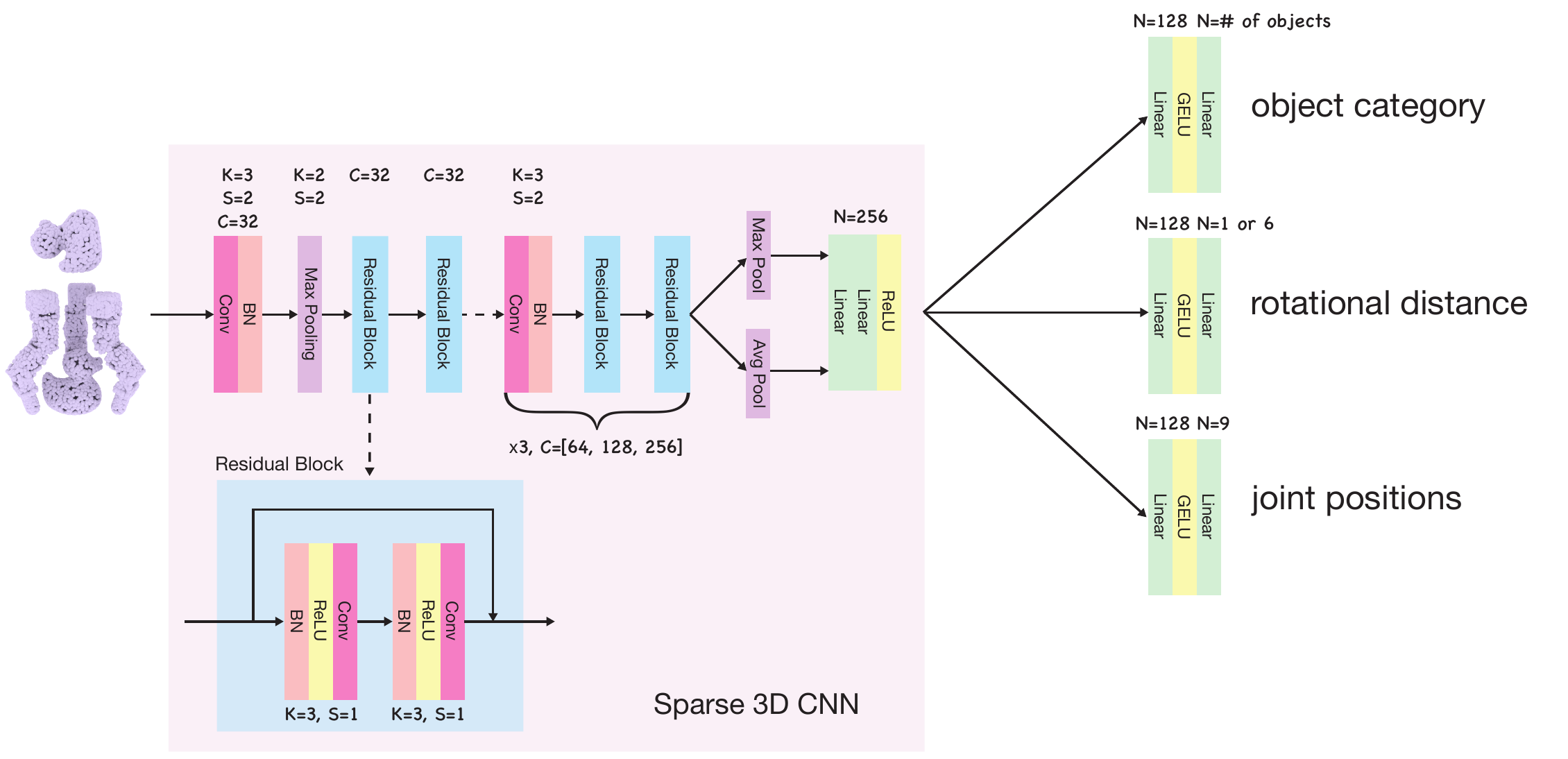}
        \caption{}
        \label{fig:pretrain_net_single_embedding}
    \end{subfigure}\\
    \begin{subfigure}[t]{0.96\linewidth}
        \centering
        \includegraphics[width=\linewidth]{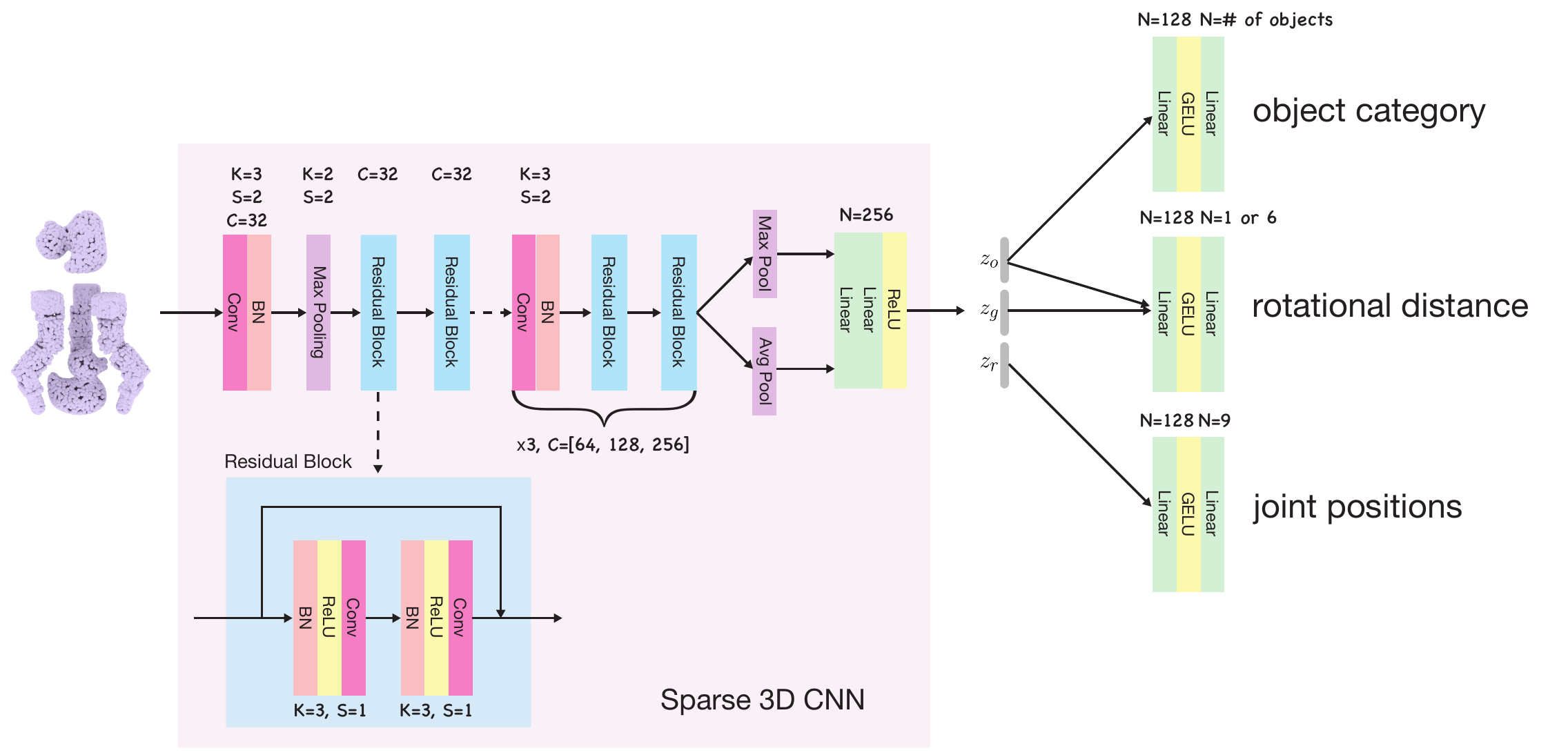}
        \caption{}
        \label{fig:pretrain_net_split_embedding}
    \end{subfigure}
    \caption{\textbf{Different architectures for prediction}. In \textbf{(a)} and \textbf{(b)}, we designed two architectures, with the difference being whether the output of the vision network is split into entity-specific embeddings or not.}
    \label{fig:pretrain_arch}
\end{figure}

\begin{figure}[!htb]
    \centering
    \begin{subfigure}[t]{0.48\linewidth}
        \centering
        \includegraphics[width=\linewidth]{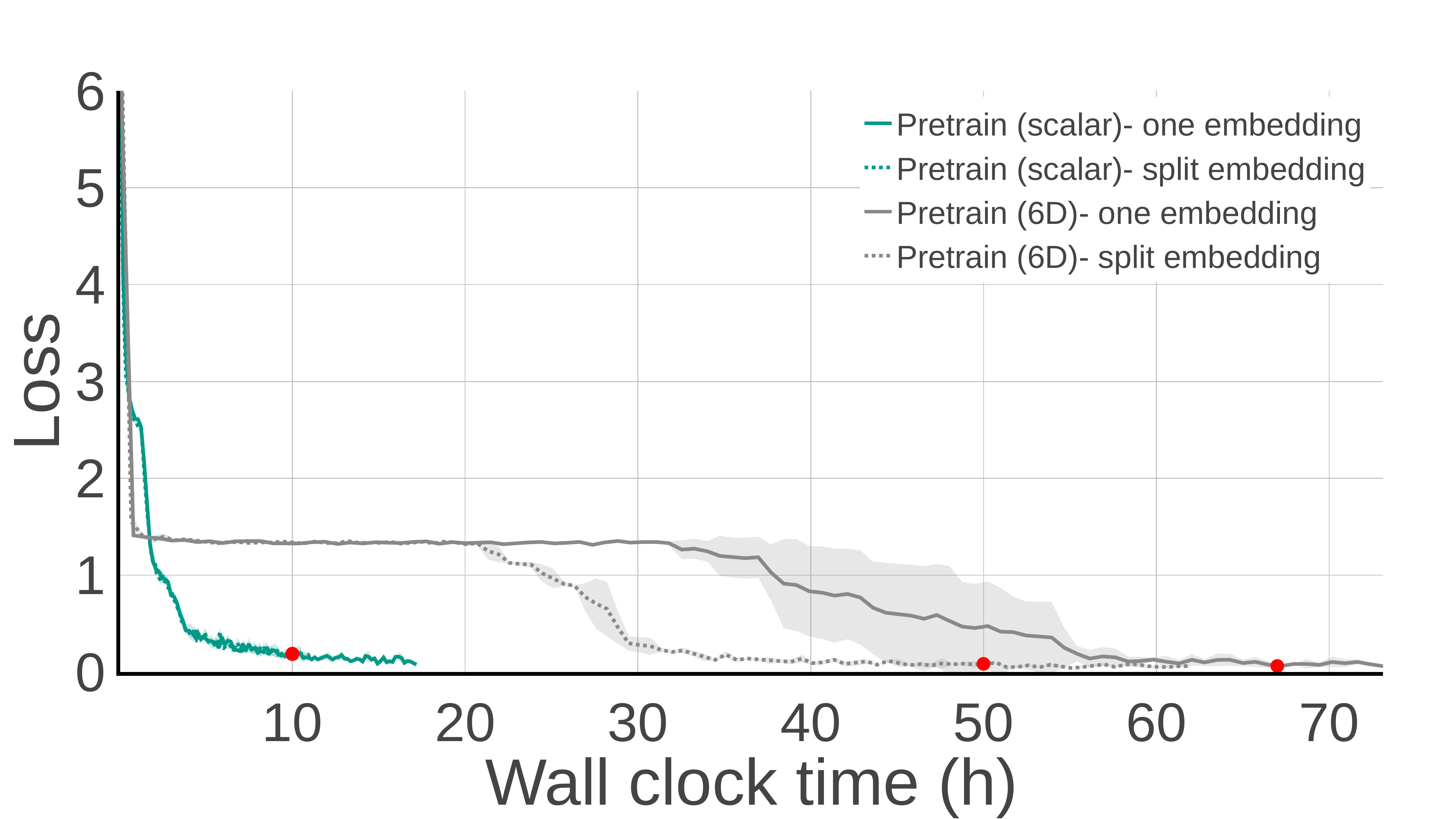}
        \caption{}
        \label{fig:pretrain_loss_time}
    \end{subfigure}
    \hfill
    \begin{subfigure}[t]{0.48\linewidth}
        \centering
        \includegraphics[width=\linewidth]{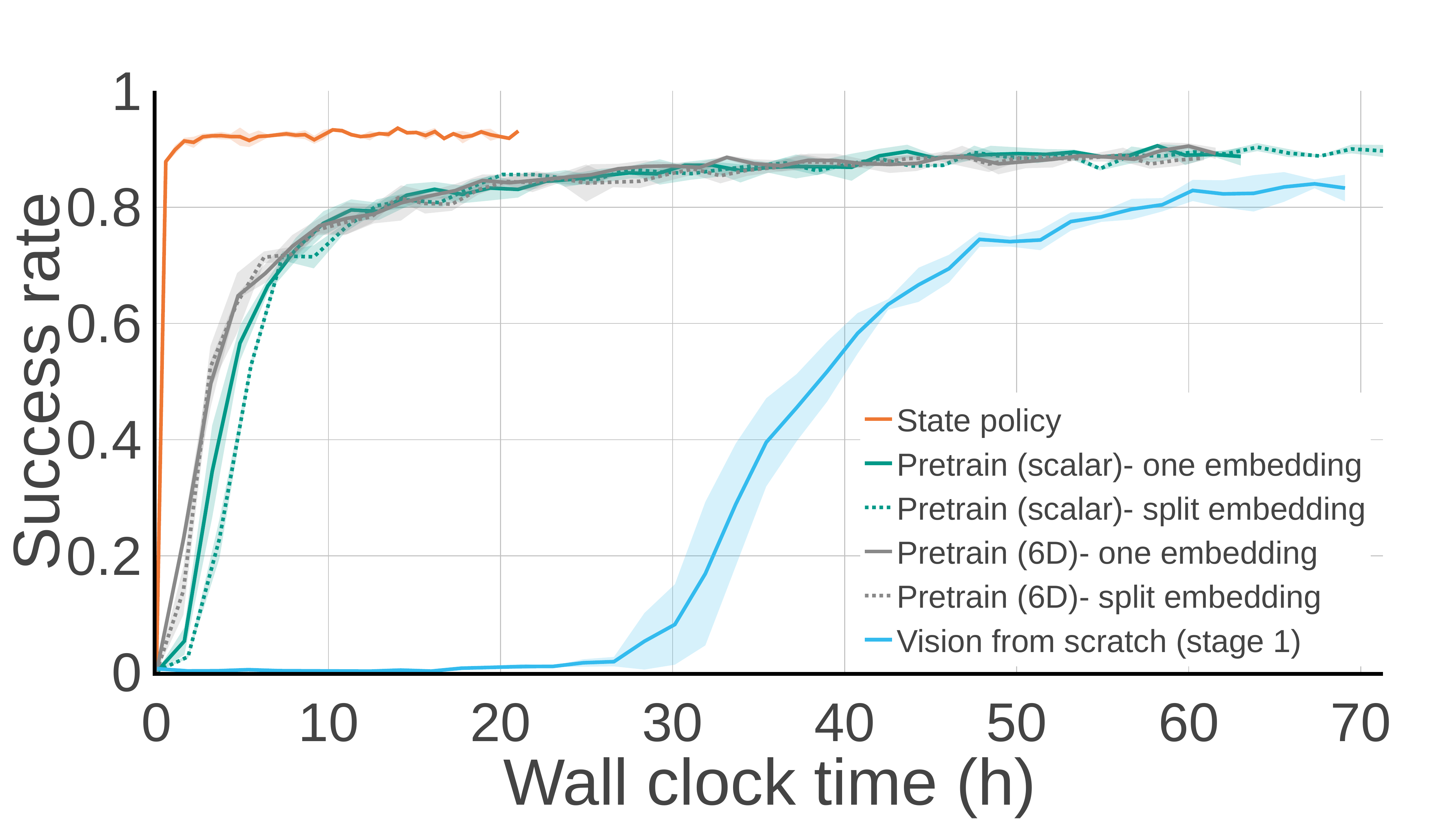}
        \caption{}
        \label{fig:pretrain_fptd_time}
    \end{subfigure}
    \caption{\textbf{Learning curves for pretraining}. \textbf{(a)} shows the learning curves under different training conditions. The red dots represent the checkpoints we took for policy learning. \textbf{(b)}: After the pre-training, we use the vision network as the policy backbone and train the policy with BC. The \textbf{State policy} is a student policy that takes as input the joint positions, rotation matrix of the relative orientation, and object position, which can be seen as an upper bound for the vision policy. \textbf{Vision from scratch (stage 1)} means the vision policy learning in stage 1 only without stage 0.}
    \label{fig:pretrain_loss}
\end{figure}

For rotational distance prediction, we experimented with two representations. The first one is the $6$D representation~\cite{zhou2019continuity} of the relative rotation matrix $\bm{R}_t^o(\bm{R}^g)^{-1}$ between the object's current orientation $\bm{R}_t^o$ and goal orientation $\bm{R}^g$. The second one is the scalar distance between the two rotation matrices ($\Delta\theta_t$). We found that pre-training the vision network to predict the scalar distance leads to faster convergence during pre-training than predicting the $6$D representation of the relative rotation matrix. In addition, we experimented with two ways of using the output of the vision network for the state prediction tasks: (1) three prediction tasks use the same embedding (\figref{fig:pretrain_net_single_embedding}), (2) we split the vision network output into three parts (object embedding $\bm{z}^o$, goal embedding $\bm{z}^g$, robot embedding $\bm{z}^r$), and each prediction task only uses the relevant embedding (\figref{fig:pretrain_net_split_embedding}). When training to predict the scalar rotational distance, we don't find these two ways of using the vision network output to make a difference. However, when training to predict the $6$D representation of the rotational distance, splitting the embedding leads to a notably faster pre-training (\figref{fig:pretrain_loss_time}). After the pre-training converges, we proceed to Stage $1$ using the pre-trained vision network backbones. We found that all four pre-training schemes lead to a substantial and similar speedup for policy learning (\figref{fig:pretrain_fptd_time}). We also did an ablation study on the importance of different pre-training tasks in \secref{appsec:pretrain_ablation}.

\subsection*{Ablation on the prediction tasks for vision network pre-training}
\label{appsec:pretrain_ablation}

\begin{figure}[!htb]
    \centering
    \begin{subfigure}[t]{0.49\linewidth}
        \centering
        \includegraphics[width=0.95\linewidth]{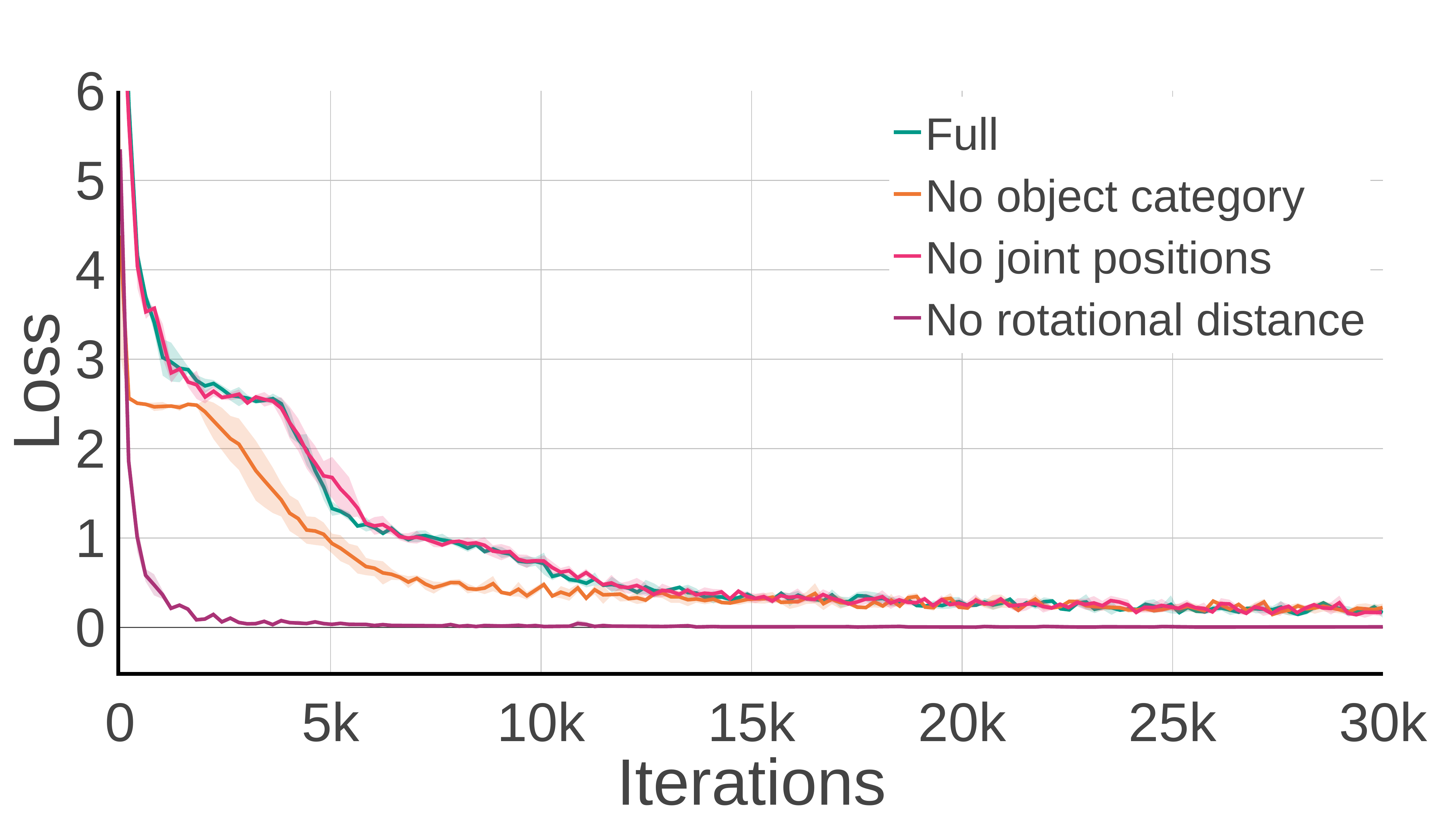}
        \caption{}
        \label{fig:ablation_pretrain_stage0}
    \end{subfigure}%
        \hfill
    \begin{subfigure}[t]{0.49\linewidth}
        \centering
        \includegraphics[width=0.95\linewidth]{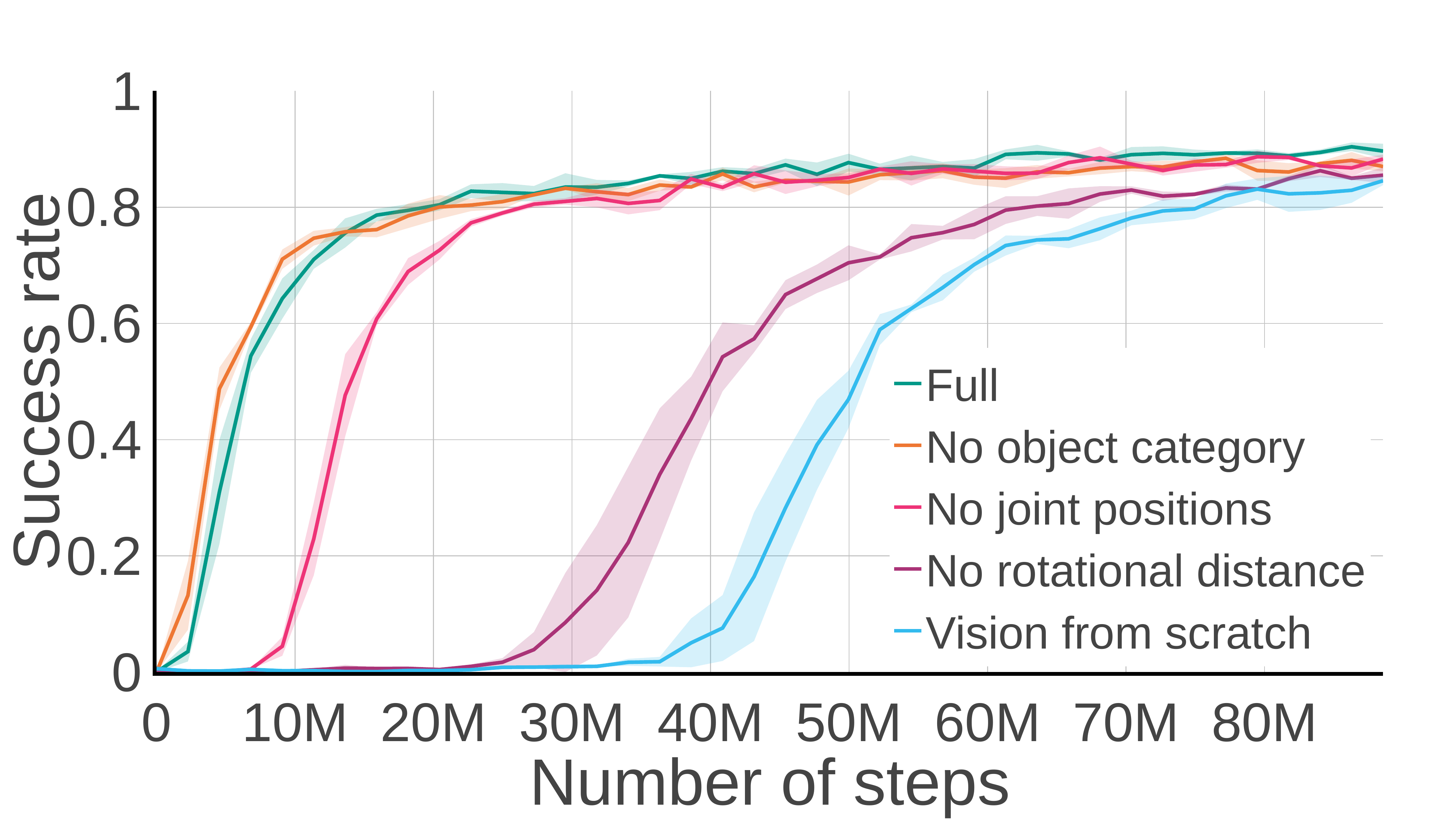}
        \caption{}
        \label{fig:ablation_pretrain_stage1}
    \end{subfigure}%
    \caption{\textbf{Effects of different prediction tasks}. \textbf{(a)}: loss curves for the pre-training in Stage 0. \textbf{(b)}: learning curves for training the vision policies with the pre-trained vision networks in Stage 1.}
    \label{fig:ablation_pretrain}
\end{figure}

We perform ablation study on different prediction tasks for pre-training the vision networks in Stage 0. In \figref{fig:ablation_pretrain}, Full represents the case of training to predict all three tasks (the scalar rotational distance, the joint positions, and the object category) with the single embedding architecture. And our ablation studies remove each prediction task individually. As shown in \figref{fig:ablation_pretrain}, removing the rotational distance prediction task makes the pretraining much easier (loss goes down much faster), but it also negatively affects the benefit of pre-training the most. It implies that the task of predicting the rotational distance is most useful for pre-training the vision network. Removing the task of predicting the joint positions slightly reduces the benefit of pre-training. Removing the task of predicting the object category has almost no effect on the policy learning.

\subsection*{Analysis for object reorientation in the air in simulation}
\label{app_subsec:failure_analysis}

We can get three outcomes when using the four-fingered hand to reorient objects in the air: the episode succeeds (Success), the controller stops the object with an orientation error bigger than $0.4$ radians (Orientation error), the object falls (Object falls), or the controller runs out of time and fails to reach the goal orientation (Time out). To see the ratio of each case, we tested the vision controller, which takes as input the realistically rendered point cloud, on the twelve objects in \figref{fig:objects_id_app} in simulation. We set the testing episode length to 180 time steps, which is equivalent to 15 seconds. For the first 24 time steps (2 seconds), we keep the table below the hand so that the hand can first grasp the objects. After the 24th step, we remove the table and check if the controller can reorient the objects in the air. Note that this is a rough approximation of the real-world testing scenario where we hand over the object to the robot hand and release the object after the hand grasps the object, because when we remove the table, there is no guarantee that the hand happens to grasp the object stably at the 24th time step. Nonetheless, empirically, we found only $10.3\%$ of the $1200$ testing episodes have the issue of object falling immediately (we check if the object falls from 24 steps to 30 steps) after the table is removed, suggesting that this is still a reasonable approximation of the real-world testing scenarios. To emulate how we stop the policy in the real world, we also stop the policy in the simulation if $|\Delta\theta_t^{pred}|<\bar{\theta}$ (we check the predicted orientation distance from the policy network) and $||\bm{a}_t||<\bar{a}$.

Each object is tested 100 times with a random initial pose and goal orientation. We plot the percentage of each case (Success, Orientation error, Time out, Object falls) in \figref{fig:failure_ratio_air}. The figure shows that the majority of failures occur because objects fall out of the hand. For object \#12, the percentage of failures due to large orientation errors is particularly high because this object is nearly symmetric in the point cloud representation. \figref{fig:air_violin_sim_real} and \figref{fig:air_violin_sim_real_time} show the comparison of two objects between the orientation error and episode time in simulation and in the real world, respectively. The results suggest that there are still gaps in the policy performance between simulation and the real world. Nonetheless, even in the real world, the median time for successful reorientation in the full SO(3) space is less than 7 seconds, demonstrating the fast and dynamic manipulation capability of the system. \figref{fig:sim_air_violin} and \figref{fig:sim_air_violin_time} show the distribution of the orientation error and episode time of the non-dropping episodes for all twelve objects in the simulation.

\subsection*{Discussion on precise manipulation}

\begin{figure}[!htb]
    \centering
    \begin{subfigure}[b]{\linewidth}
         \centering
         \includegraphics[width=\linewidth]{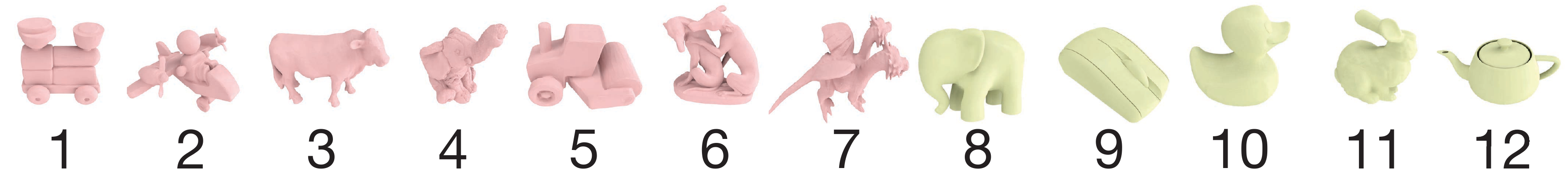}
         \caption{}
         \label{fig:objects_id_app}
     \end{subfigure}\\
         \begin{subfigure}[b]{0.43\linewidth}
         \centering
         \includegraphics[width=0.9\linewidth]{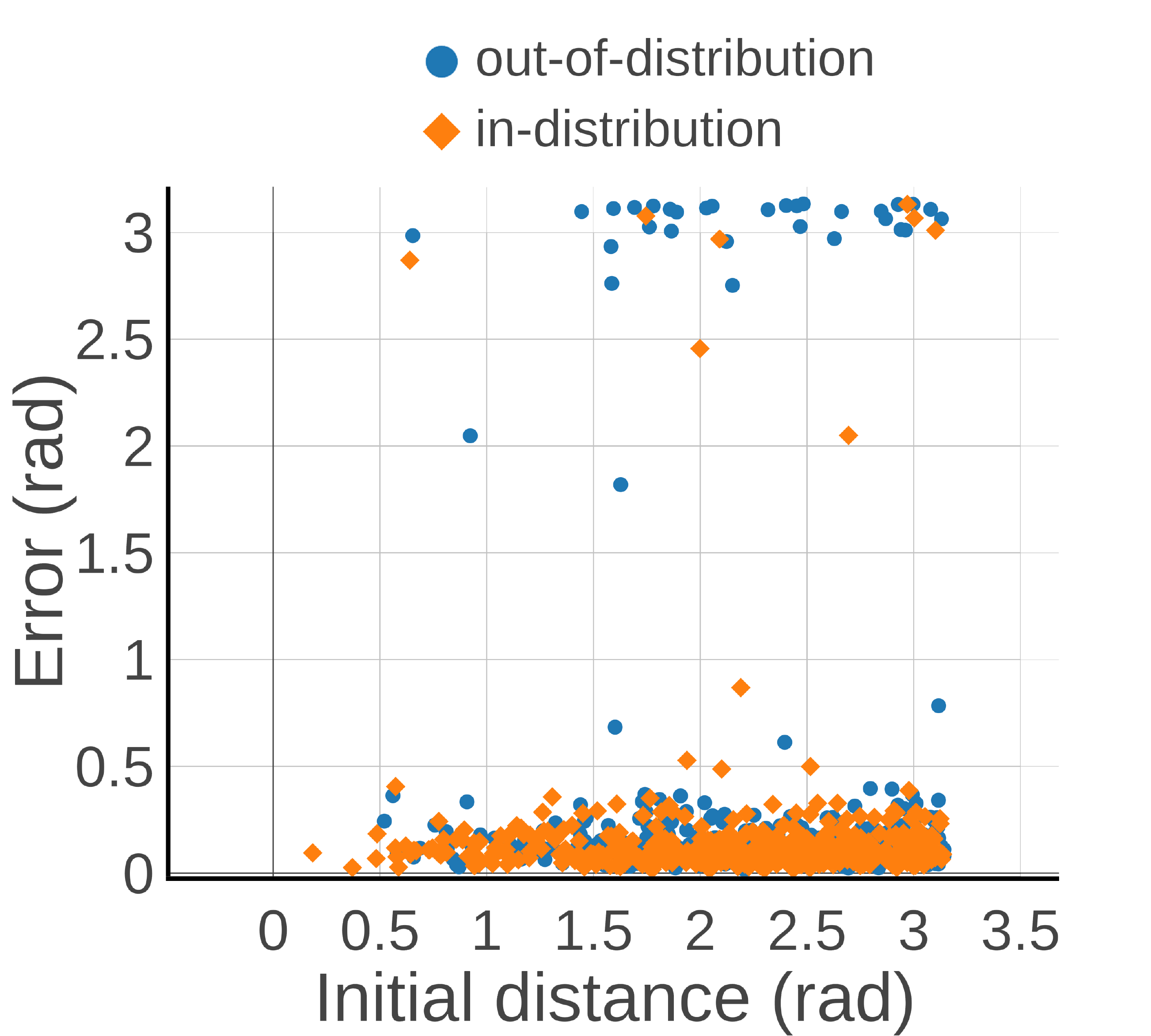}
         \caption{}
         \label{fig:sim_distance_dist}
     \end{subfigure}\hfill
         \begin{subfigure}[b]{0.55\linewidth}
         \centering
         \includegraphics[width=\linewidth]{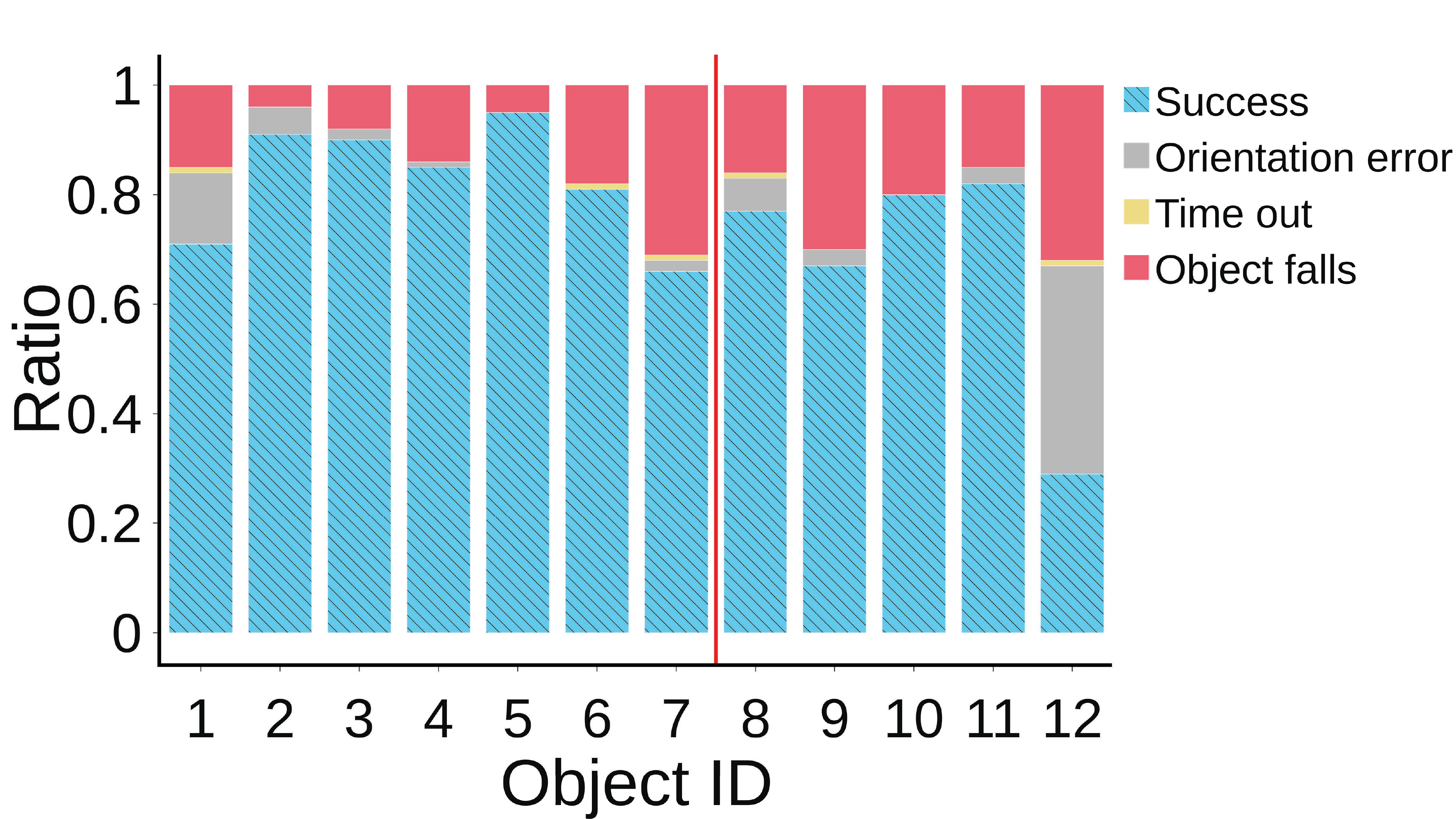}
         \caption{}
         \label{fig:failure_ratio_air}
     \end{subfigure}
    \caption{
    \textbf{Precise manipulation analysis}. Tests for object reorientation in the air in simulation. We tested each object in \figref{fig:objects_id_app} 100 times with random initial pose and goal orientation. \textbf{(a)}: the objects we used for testing (same as Figure 3A). \textbf{(b)}: We show the relationship between the reorientation error and the distance ($\Delta\theta_0$) between the object's initial and target orientation on non-dropping tests (around $90\%$). We randomly sub-sample the tests on in-distribution objects to make sure the total numbers of points are the same for in-distribution objects and out-of-distribution objects in this plot. \textbf{(c)}: We categorize the testing results for each object into four cases: Success, Orientation error (where the controller stops the object with an orientation error greater than $0.4$ radians), Time out, Object falls.
    }
    \label{fig:failure_analysis_air}
\end{figure}

\begin{figure}[!htb]
    \centering
         \begin{subfigure}[b]{0.48\linewidth}
         \centering
         \includegraphics[width=\linewidth]{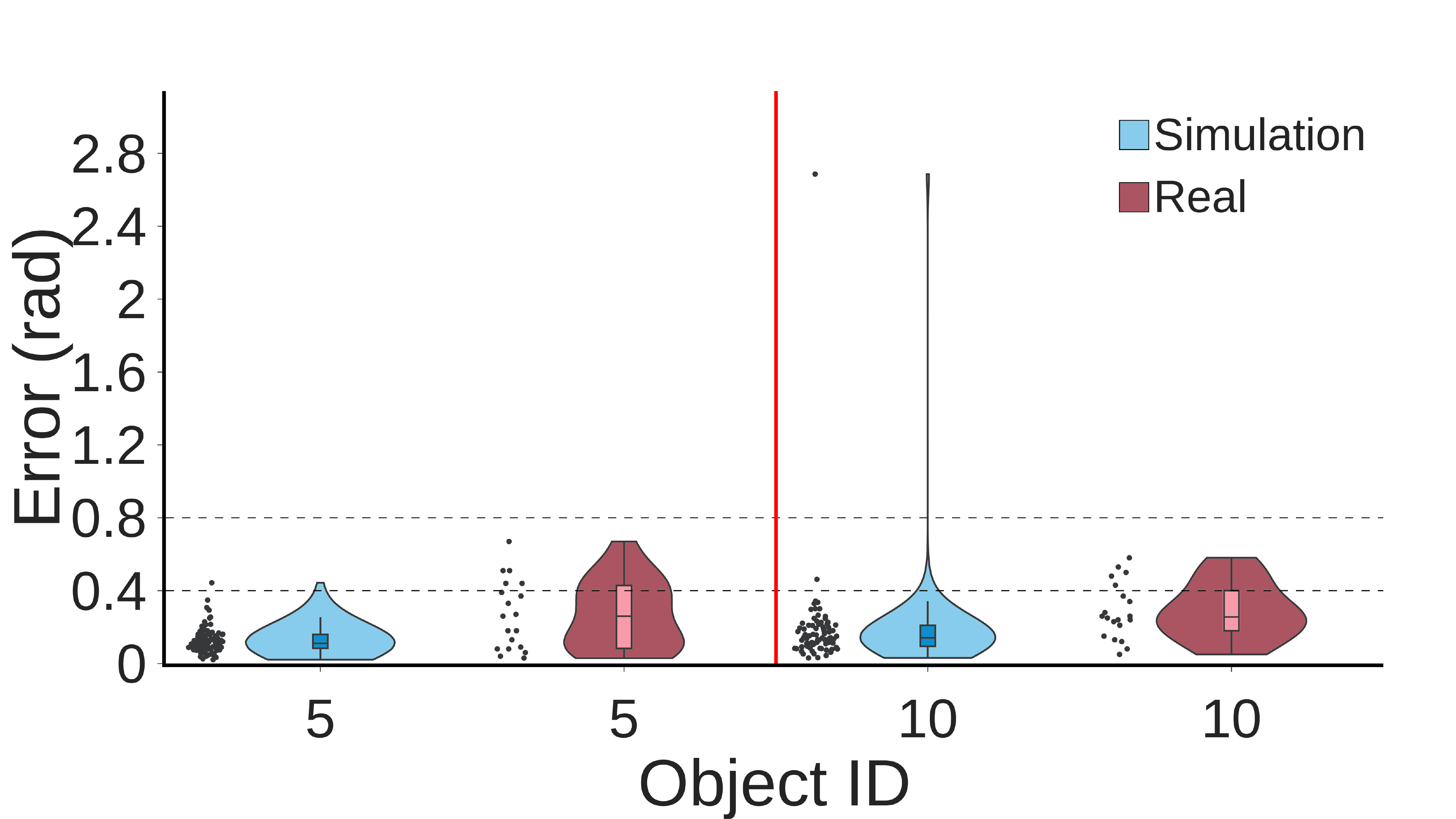}
         \caption{}
         \label{fig:air_violin_sim_real}
     \end{subfigure}\hfill
              \begin{subfigure}[b]{0.48\linewidth}
         \centering
         \includegraphics[width=\linewidth]{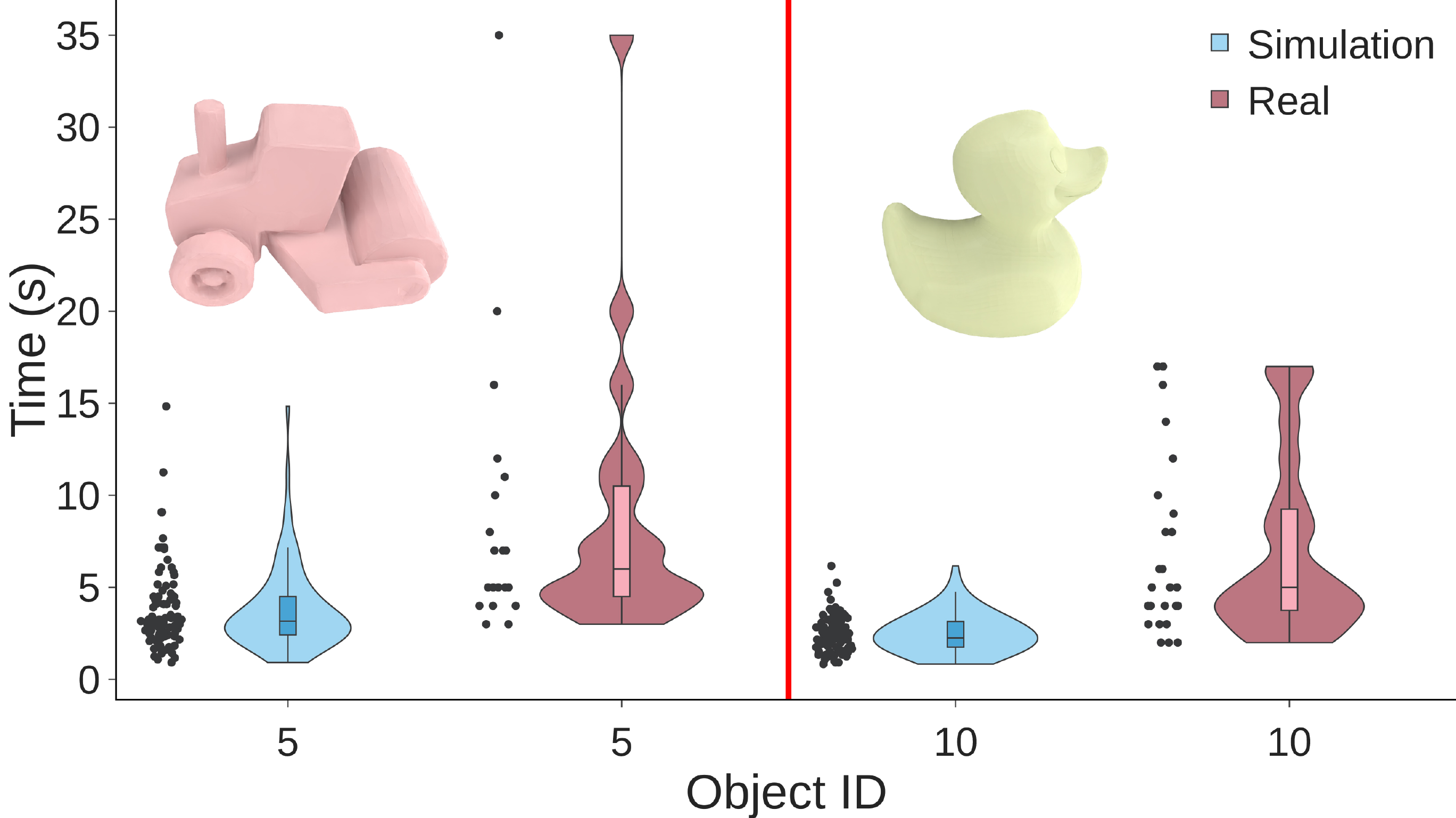}
         \caption{}
         \label{fig:air_violin_sim_real_time}
     \end{subfigure}
     \\
         \begin{subfigure}[b]{0.48\linewidth}
         \centering
         \includegraphics[width=\linewidth]{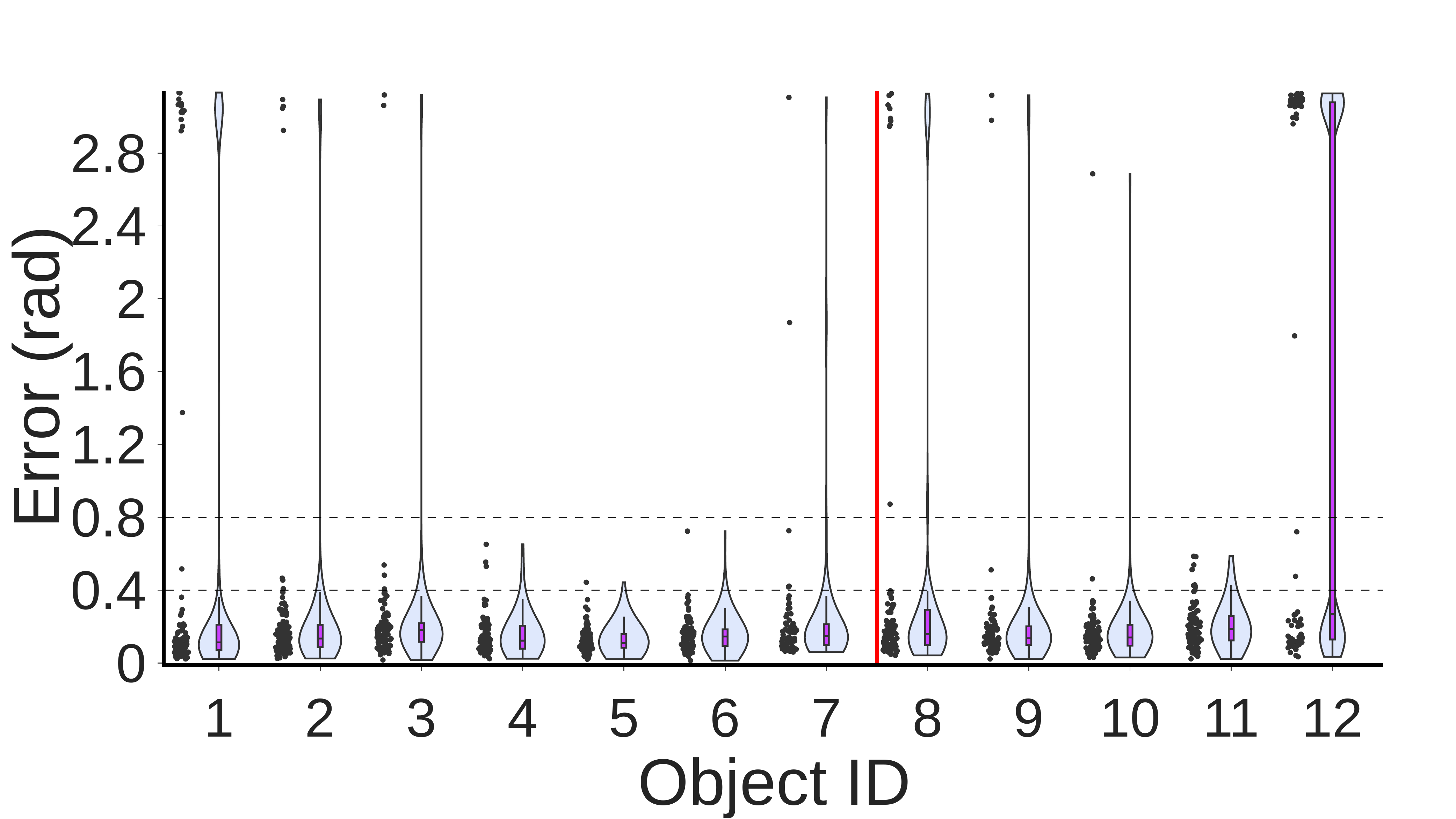}
         \caption{}
         \label{fig:sim_air_violin}
     \end{subfigure}
     \hfill
    \begin{subfigure}[b]{0.48\linewidth}
         \centering
         \includegraphics[width=\linewidth]{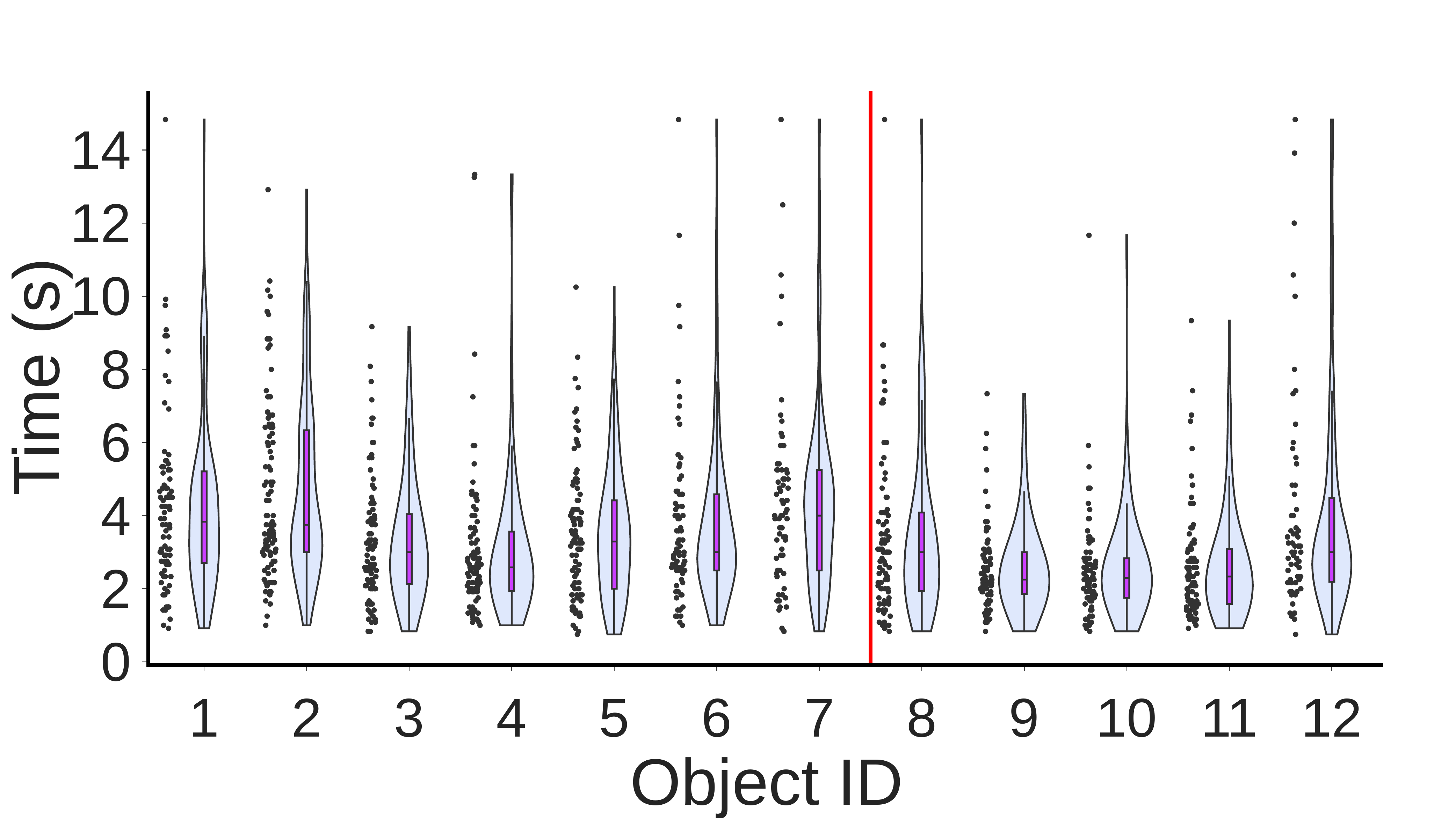}
         \caption{}
         \label{fig:sim_air_violin_time}
     \end{subfigure}
    \caption{\textbf{Precise manipulation analysis}. Following \figref{fig:failure_analysis_air}, \textbf{(a)} and \textbf{(b)}: With object $\#5$ and $\#10$, we compare the distribution of the orientation error and the elapsed time of the episodes in simulation and in the real world. The controller achieves lower error and uses shorter time in simulation. We can see there is still a gap between the simulation and real-world performance. \textbf{(c)} and \textbf{(d)}: we show the distribution of the reorientation error and episode time of the non-dropping testing episodes on all twelve objects in simulation.
    }
    \label{fig:air_error_time}
\end{figure}

In this study, we adopted a success threshold of $0.4$ radians, consistent with the definition used in a previous study~\cite{andrychowicz2020learning}. It is natural to wonder if our controller can accurately reorient objects with a smaller reorientation error. To provide more insight into the system performance at a stricter success criterion of $0.1$ radians, we did more analysis in simulation. We find that at $0.4$ radians, the success rate is $72.3\%$ (from $1200$ tests on the $12$ objects shown in \figref{fig:objects_id_app}), but drops to $25.9\%$ at $0.1$ radians. However, this drop is not due to the inability of our controller to perform precise manipulation. It is, in fact, largely attributed to the failure of the module that predicts the rotational distance between the object’s current and target orientation, which in turn is used to stop the hand. For instance, if we use the ground-truth distance to stop the controller, success rates of $67.4\%$ and $80.9\%$ are achieved at $0.1$ radians and $0.4$ radians thresholds, respectively. 

\begin{figure}[!htb]
    \centering
    \begin{subfigure}[t]{0.48\linewidth}
        \centering
        \includegraphics[width=\linewidth]{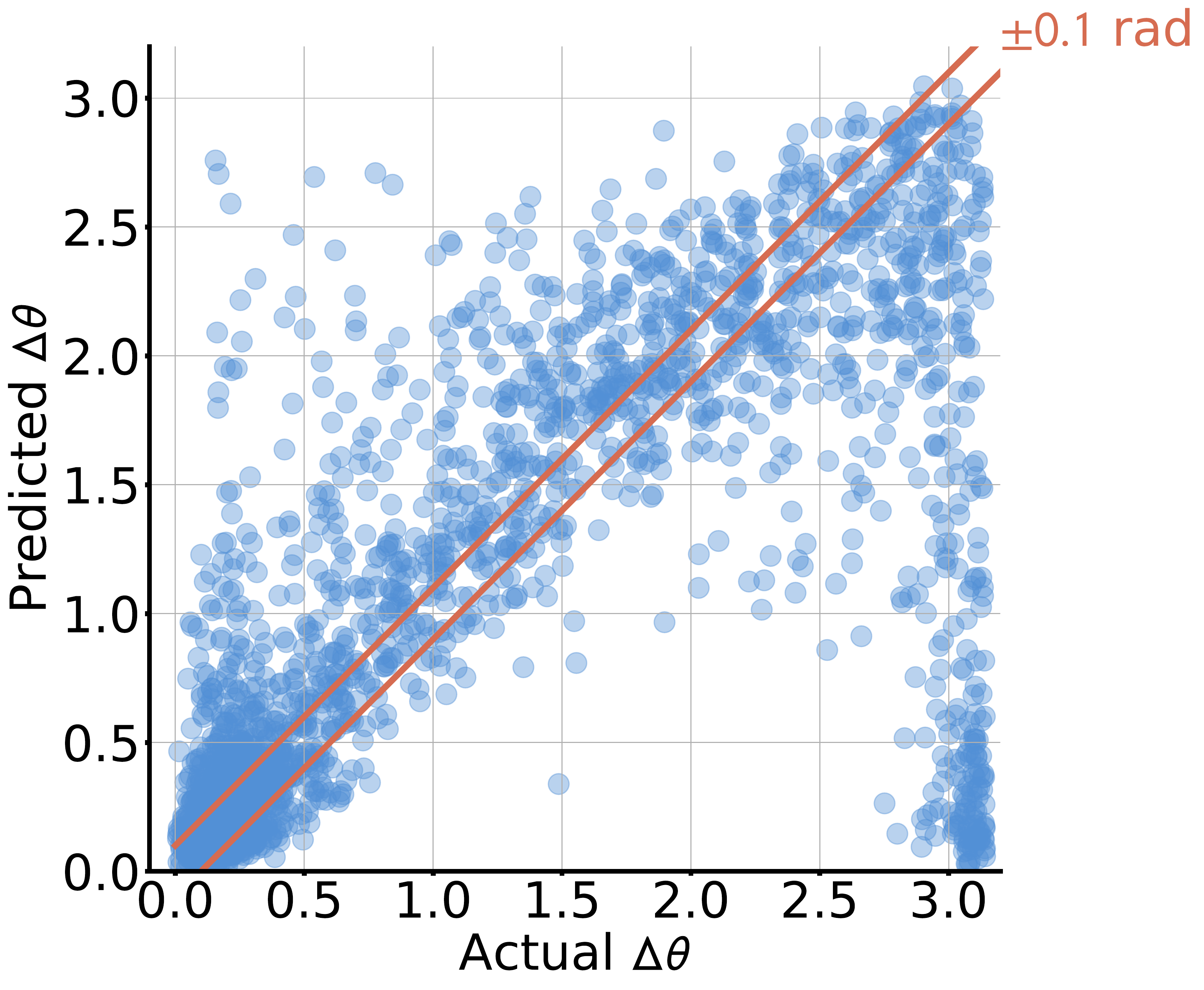}
        \caption{}
        \label{fig:pred_dist_error_whole}
    \end{subfigure}
    \hfill
    \begin{subfigure}[t]{0.48\linewidth}
        \centering
        \includegraphics[width=\linewidth]{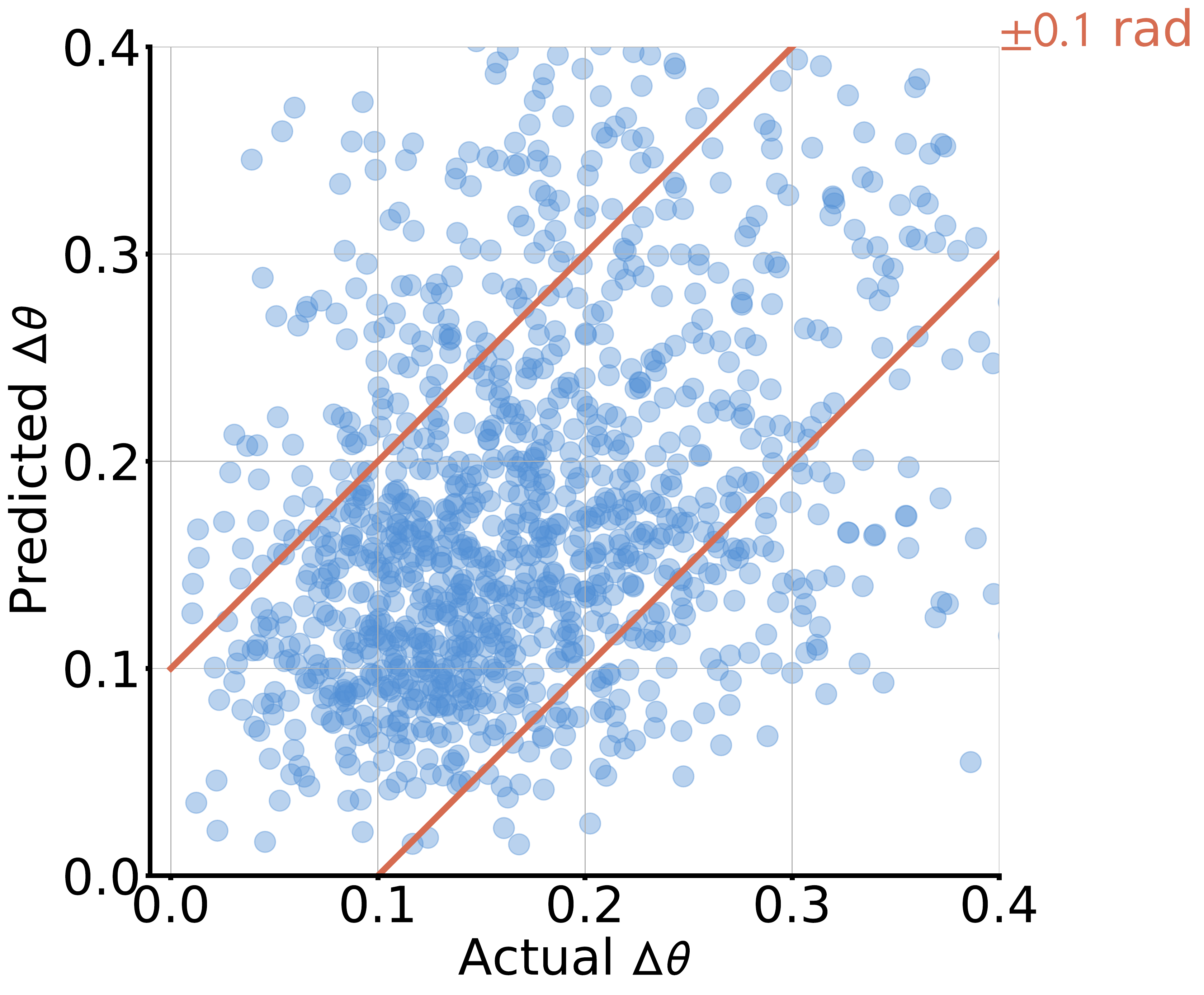}
        \caption{}
        \label{fig:pred_dist_error_detail}
    \end{subfigure}
    \caption{\textbf{Reorientation error analysis}. We plotted the actual and predicted rotational distance between the object and goal orientation from 1200 testing episodes on twelve objects in \figref{fig:objects_id_app}. \textbf{(b)} is a zoomed-in version of \textbf{(a)} for $x\in[0, 0.4]$ radians and $y\in[0,0.4]$ radians. The region between the two red lines indicates an error of less than $0.1$ radians. Overall, our rotational distance predictor performs reasonably well, but it has limited accuracy when the actual distance is $\Delta\theta\leq 0.4$ radians, indicating the difficulty of precisely predicting the rotational distance.}
    \label{fig:pred_dist_error}
\end{figure}

To provide readers with a better understanding of the accuracy of the rotation distance predictor, we have also included scatter plots comparing ground-truth and predicted distance in \figref{fig:pred_dist_error}. More specifically, we conducted simulation tests on each of the twelve objects $100$ times, with a success threshold $\bar{\theta}$ set to $0.1$ radians. For each trial, we recorded the trajectory and predicted rotational distance at each time step. We then plotted the actual and predicted rotational distances between the object and the goal orientation in \figref{fig:pred_dist_error}. The figure demonstrates that although the prediction model performs reasonably well overall, it suffers from providing sufficient accuracy in the region where $\Delta\theta\leq0.4$ radians, which is of particular interest for precise manipulation. For example, among the data points for which the actual $\Delta\theta\leq0.1$ radians, only $29.4\%$ of the predictions correctly estimated distances within $0.1$ radians. When the actual $\Delta\theta\leq0.4$ radians, $85.9\%$ of the predictions estimated the distance to be less than $0.4$ radians. Inaccurate predictions of rotational distance, coupled with observation and command delays in the real system, make precise manipulation with $\Delta\theta\leq 0.1$ radians challenging. This indicates one research direction for improving orientation accuracy is to train a better predictor for orientation distance or a better classifier for identifying whether the goal orientation has been reached.

\clearpage
\renewcommand{\arraystretch}{1.2}
\begin{table}[!htb]
\centering
\caption{Hyper-parameter Setup}
\label{tbl:hyper_params}
\resizebox{\columnwidth}{!}{
\begin{tabular}{cccccc} 
\hline
Hyperparameter                                                                                      & Value   & Hyperparameter               & Value                                     & Hyperparameter                                                                          & Value  \\ 
\hline
\multicolumn{6}{c}{\textbf{Teacher policy}}                                                                                                                                                                                                                                                 \\ 
\hline
\# of envs                                                                                          & 32000   & batch size                   & \textcolor[rgb]{0.125,0.129,0.141}{64000} & \begin{tabular}[c]{@{}c@{}}\# of rollout steps~\\per policy update\end{tabular}         & 8      \\
GAE lambda                                                                                          & 0.95    & Reward discount              & 0.99                                      & \begin{tabular}[c]{@{}c@{}}\# of policy update epochs \\after each rollout\end{tabular} & 12     \\
Actor learning rate                                                                                 & 0.0003  & Critic learning rate         & 0.001                                     & PPO clip range                                                                          & 0.1    \\
$\epsilon_\theta$                                                                                   & 0.4 radians & $c_1$                        & 800                                         & $c_2$                                                                                   & 1    \\
$c_3$                                                                                               & -1      & $c_4$                        & -20                                       & $c_5$                                                                                   & -100   \\
$c_6$                                                                                               & -1      & $c_7$                        & -2                                        & $\bar{p}$                                                                               & 0.15   \\
$\bar{p}_z$                                                                                         & 0.16    & $\bar{\dot{q}}$              & 0.25                                      & $\bar{v}$                                                                               & 0.04   \\
$\bar{\omega}$                                                                                           & 0.5     &  $c_d$                            &                        15                   &                                                                                         &        \\ 
\hline
\multicolumn{6}{c}{\textbf{Student policy}}                                                                                                                                                                                                                                                 \\ 
\hline
\# of envs (Stage 1/Stage 2)                                                                        & 400/260 & batch size (Stage 1/Stage 2) & 40/20                                     & \begin{tabular}[c]{@{}c@{}}\# of pts sampled from \\each CAD model\end{tabular}         & 500    \\
\begin{tabular}[c]{@{}c@{}}\# of pts sampled from realistically \\rendered point cloud\end{tabular} & 6000    & learning rate                & 0.0003                                    & \begin{tabular}[c]{@{}c@{}}\# of rollout steps~\\per policy update\end{tabular}         & 80     \\
\hline
\end{tabular}
}
\end{table}
\renewcommand{\arraystretch}{1.}

\renewcommand{\arraystretch}{1.4}
\begin{table}[!htb]
\centering
\caption{Object mass.}
\label{tbl:object_mass}
\begin{tabular}{cccccccc} 
\hline
Object & Mass (g) & Object & Mass (g) & Object & Mass (g) & Object & Mass (g)  \\ 
\hline
 \begin{minipage}{.06\textwidth}
\includegraphics[width=\linewidth]{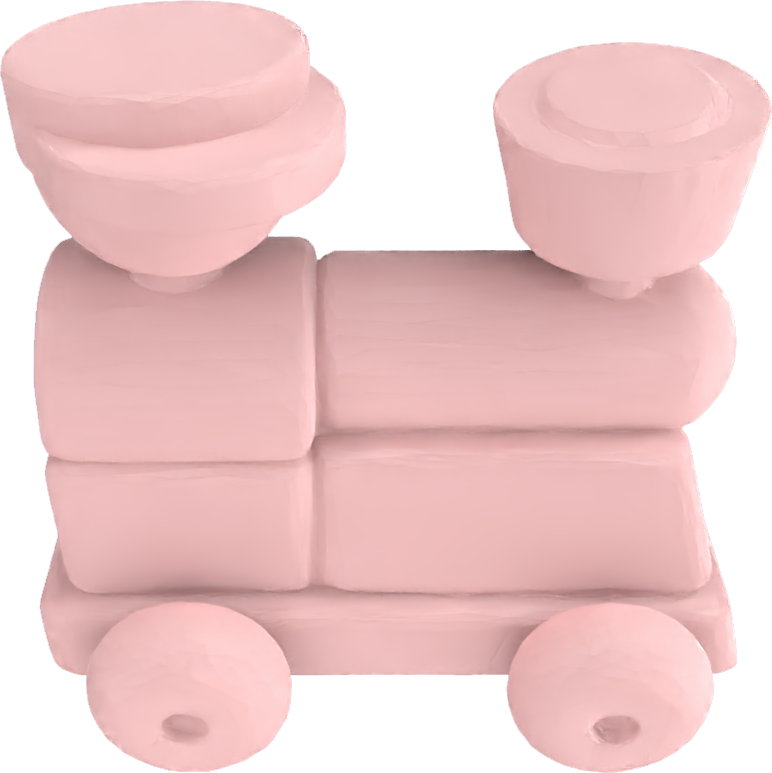}
\end{minipage}      &      158.0    &     \begin{minipage}{.06\textwidth}
\includegraphics[width=\linewidth]{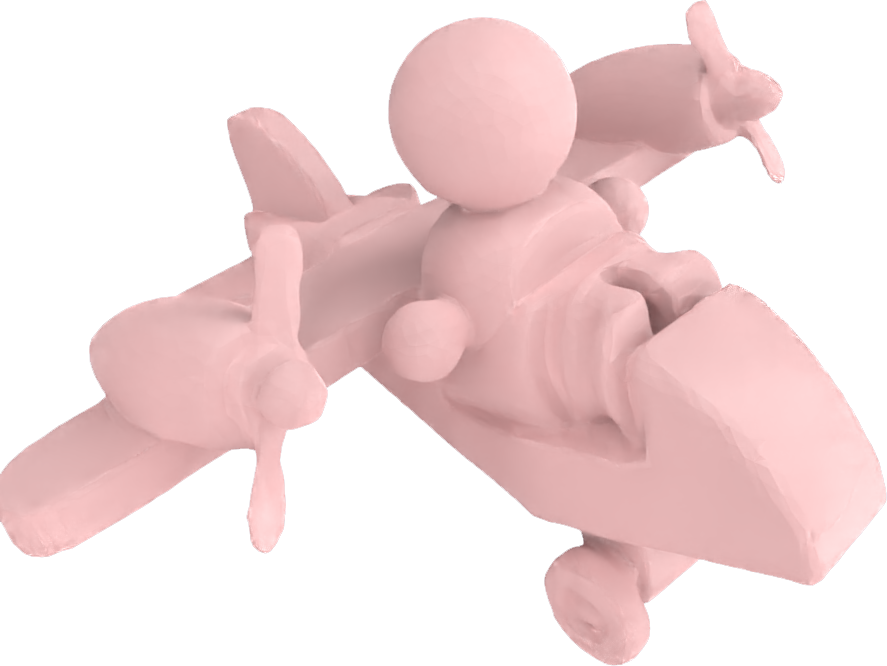}
\end{minipage}    &     95.0      &
 \begin{minipage}{.06\textwidth}
\includegraphics[width=\linewidth]{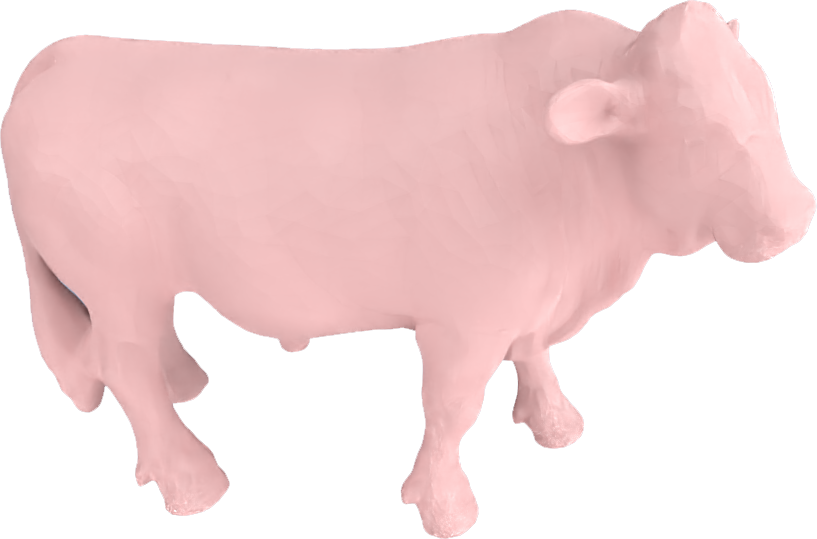}
\end{minipage}      &     111.1    &     \begin{minipage}{.06\textwidth}
\includegraphics[width=\linewidth]{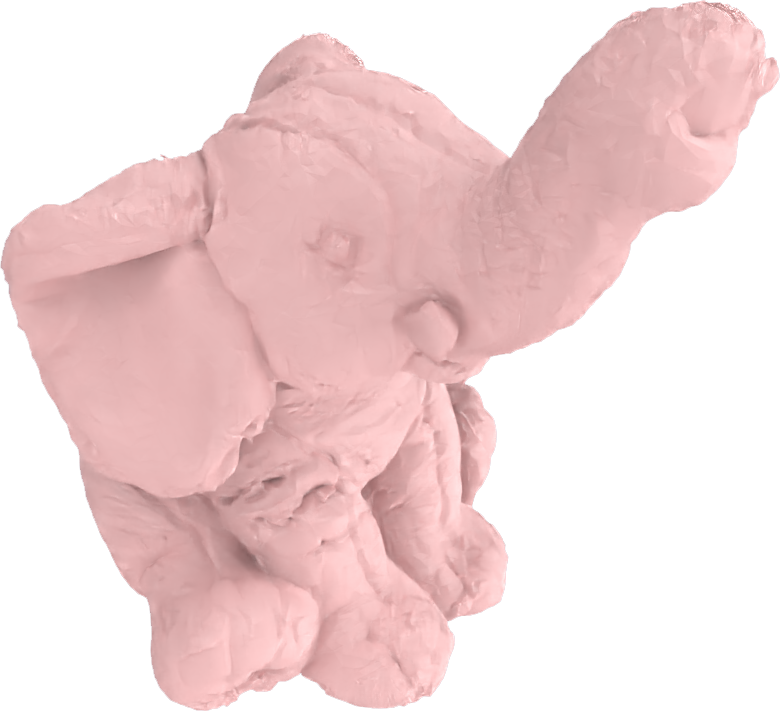}
\end{minipage}    &     116.9      \\
 \begin{minipage}{.06\textwidth}
\includegraphics[width=\linewidth]{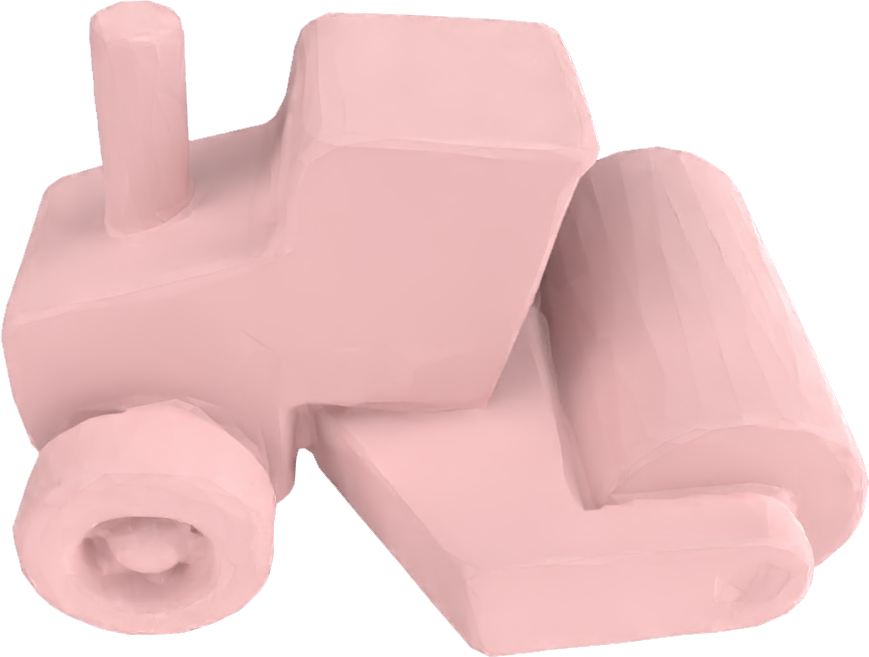}
\end{minipage}      &      151.8    &     \begin{minipage}{.06\textwidth}
\includegraphics[width=1.3\linewidth]{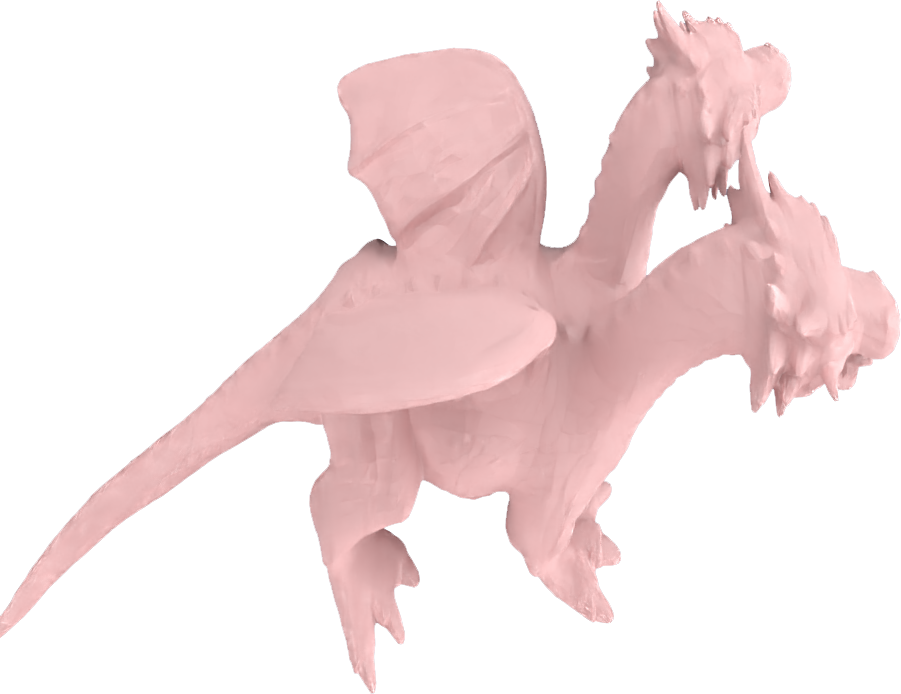}
\end{minipage}    &     70.3      &
 \begin{minipage}{.06\textwidth}
\includegraphics[width=\linewidth]{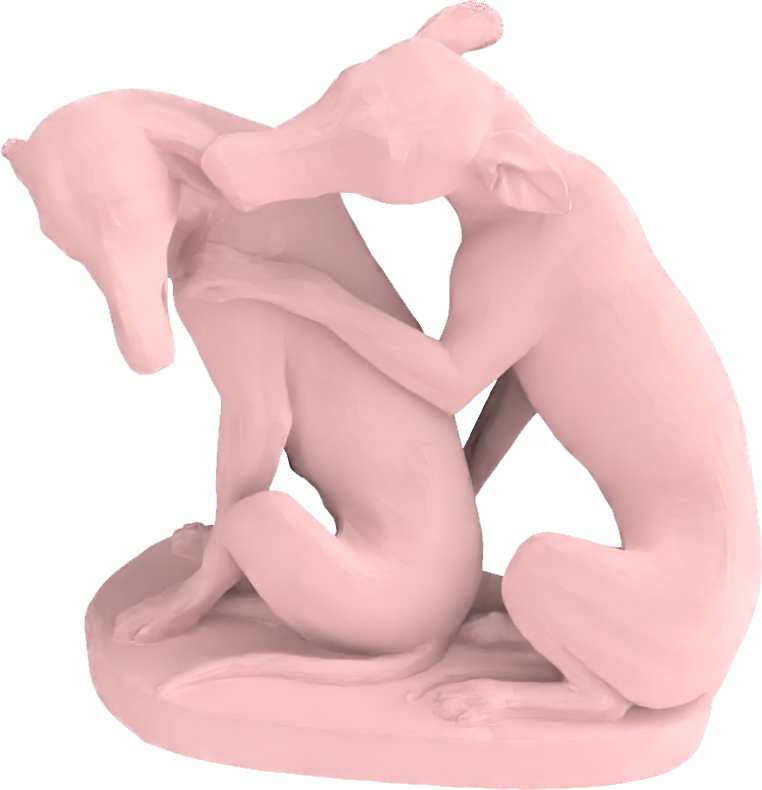}
\end{minipage}      &      104.3    &     \begin{minipage}{.06\textwidth}
\includegraphics[width=\linewidth]{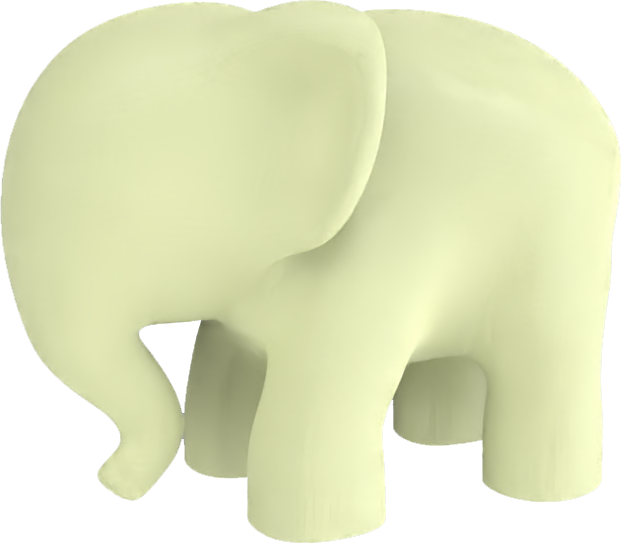}
\end{minipage}    &     92.1      \\
 \begin{minipage}{.06\textwidth}
\includegraphics[width=\linewidth]{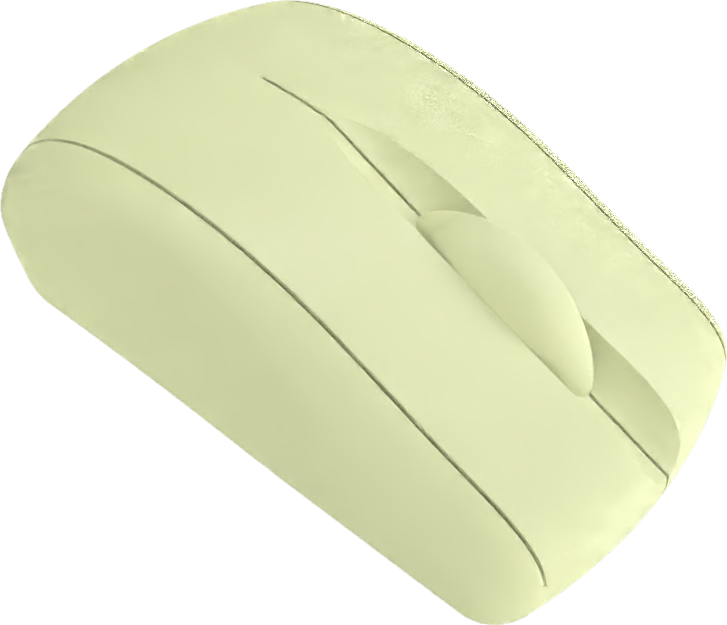}
\end{minipage}      &      106.3    &     \begin{minipage}{.06\textwidth}
\includegraphics[width=\linewidth]{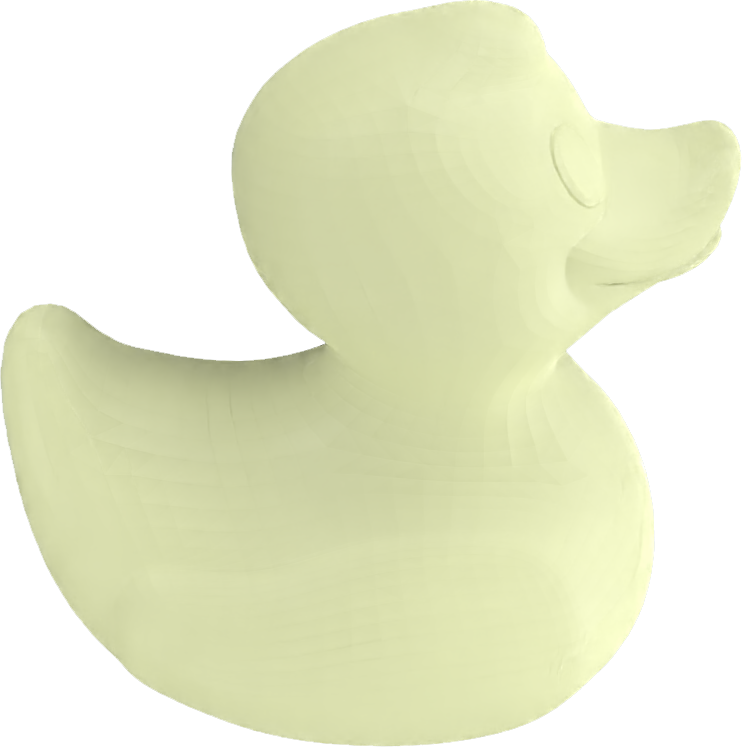}
\end{minipage}    &     140.9      &
 \begin{minipage}{.06\textwidth}
\includegraphics[width=\linewidth]{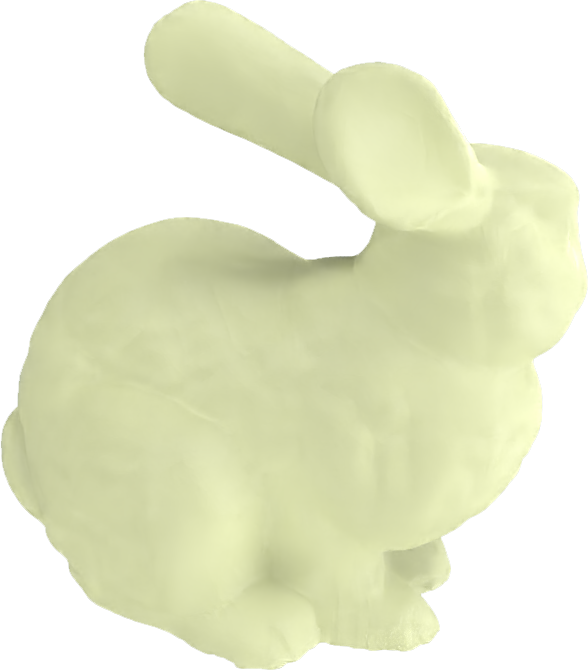}
\end{minipage}      &      86.8    &     \begin{minipage}{.06\textwidth}
\includegraphics[width=1.4\linewidth]{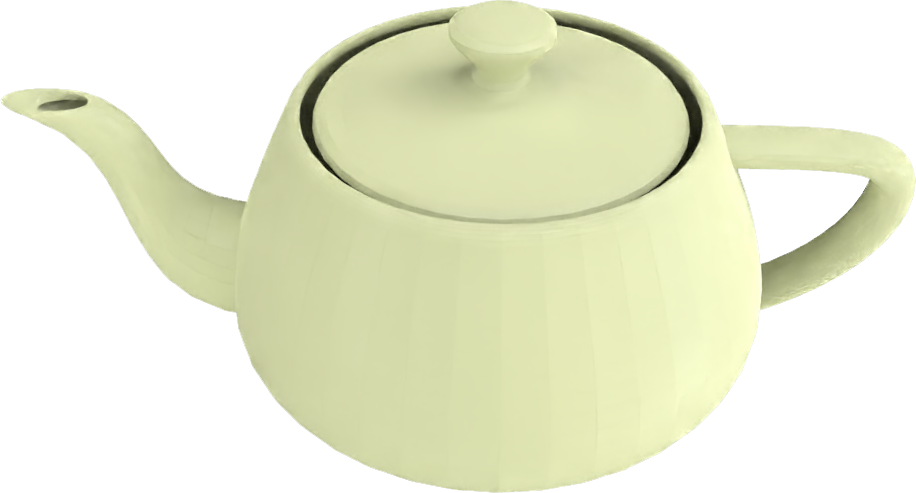}
\end{minipage}    &     117.6      \\
 \begin{minipage}{.06\textwidth}
\includegraphics[width=\linewidth]{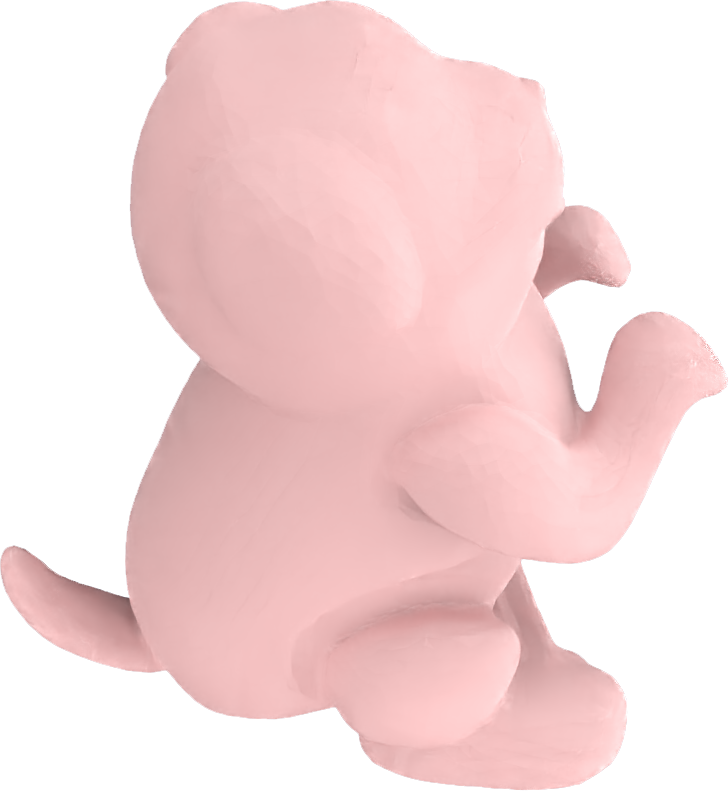}
\end{minipage}      &      148.1    &     \begin{minipage}{.06\textwidth}
\includegraphics[width=\linewidth]{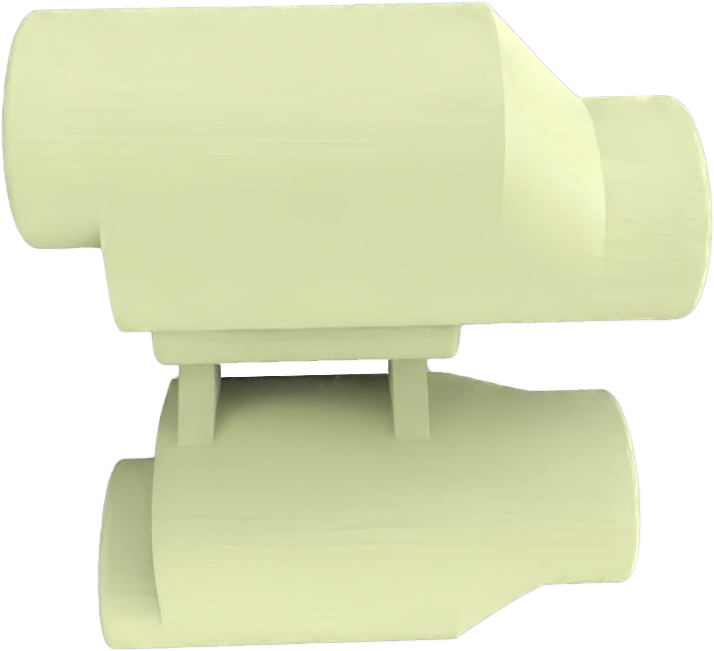}
\end{minipage}    &     162.1     &
 \begin{minipage}{.06\textwidth}
\includegraphics[width=\linewidth]{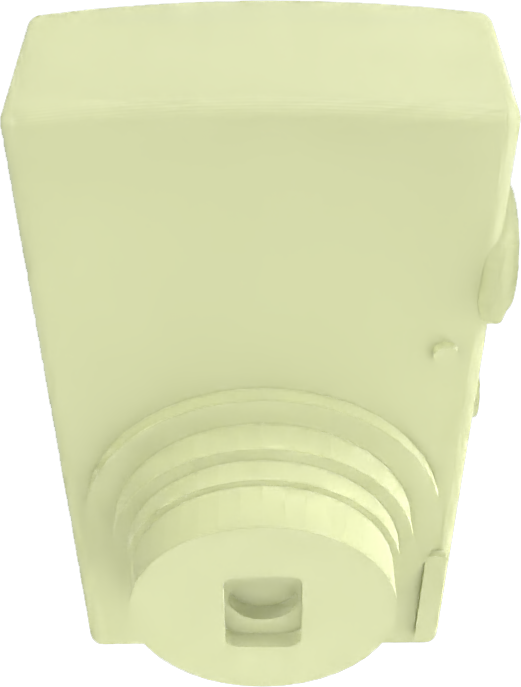}
\end{minipage}      &      106.1    &     \begin{minipage}{.06\textwidth}
\includegraphics[width=1.4\linewidth]{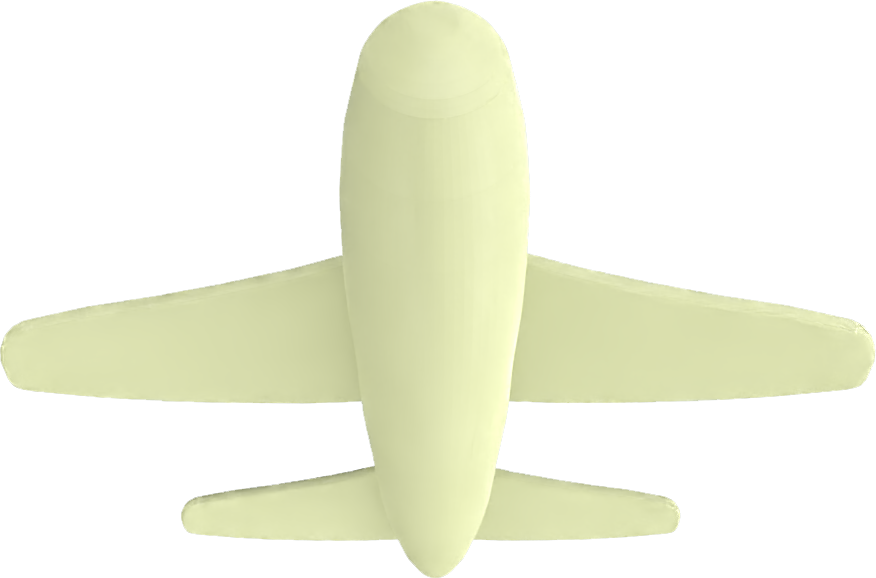}
\end{minipage}    &     50.1     \\
 \begin{minipage}{.06\textwidth}
\includegraphics[width=\linewidth]{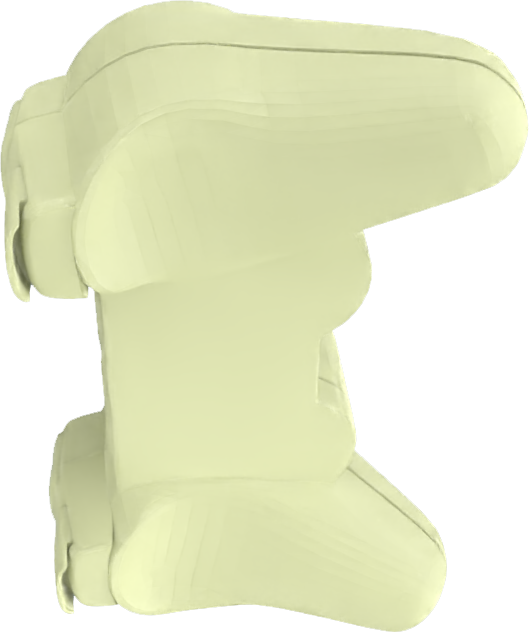}
\end{minipage}      &      60.5 &
 \begin{minipage}{.06\textwidth}
\includegraphics[width=\linewidth]{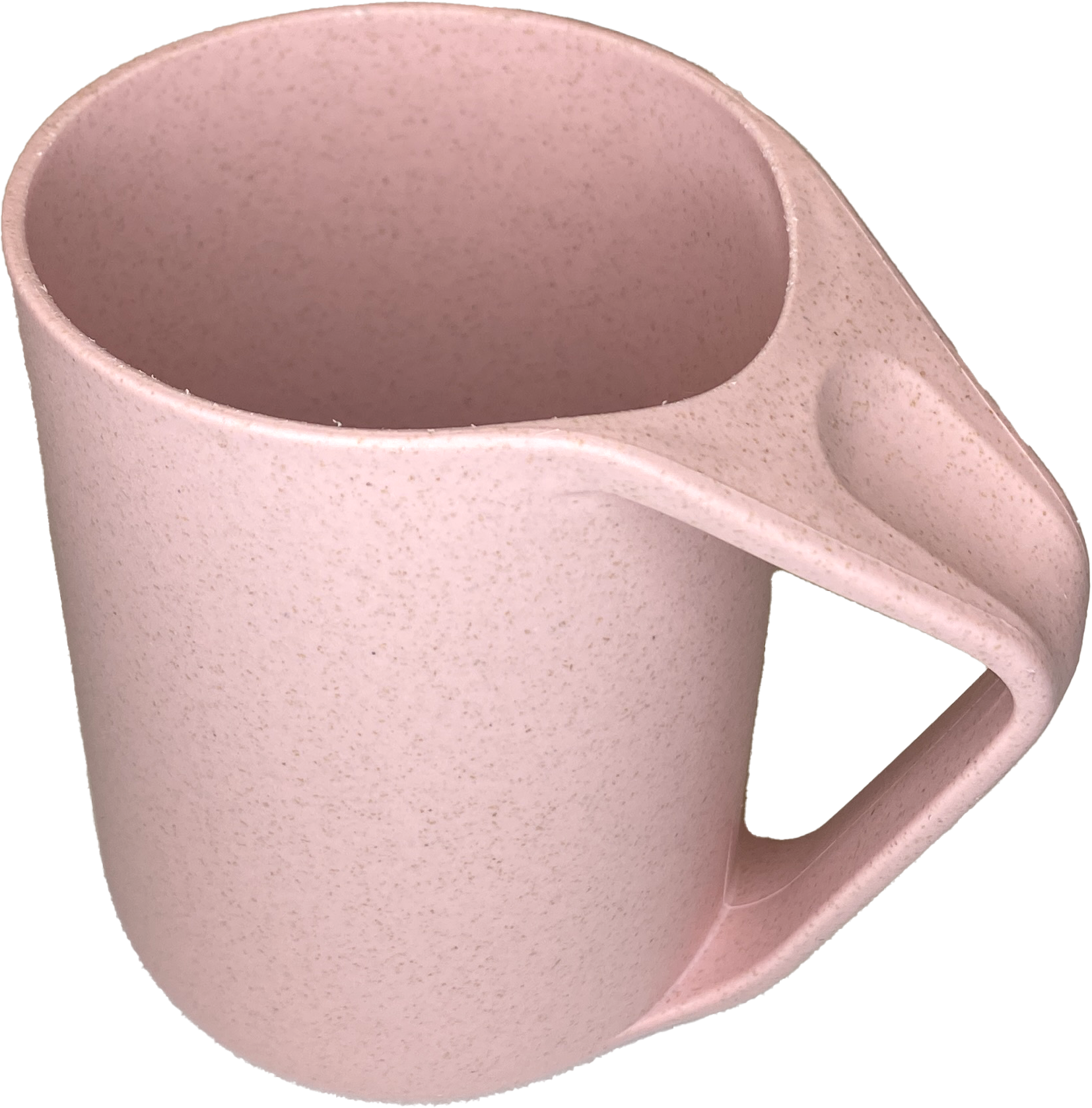}
\end{minipage}      &      104.1    &     \begin{minipage}{.06\textwidth}
\includegraphics[width=1.4\linewidth]{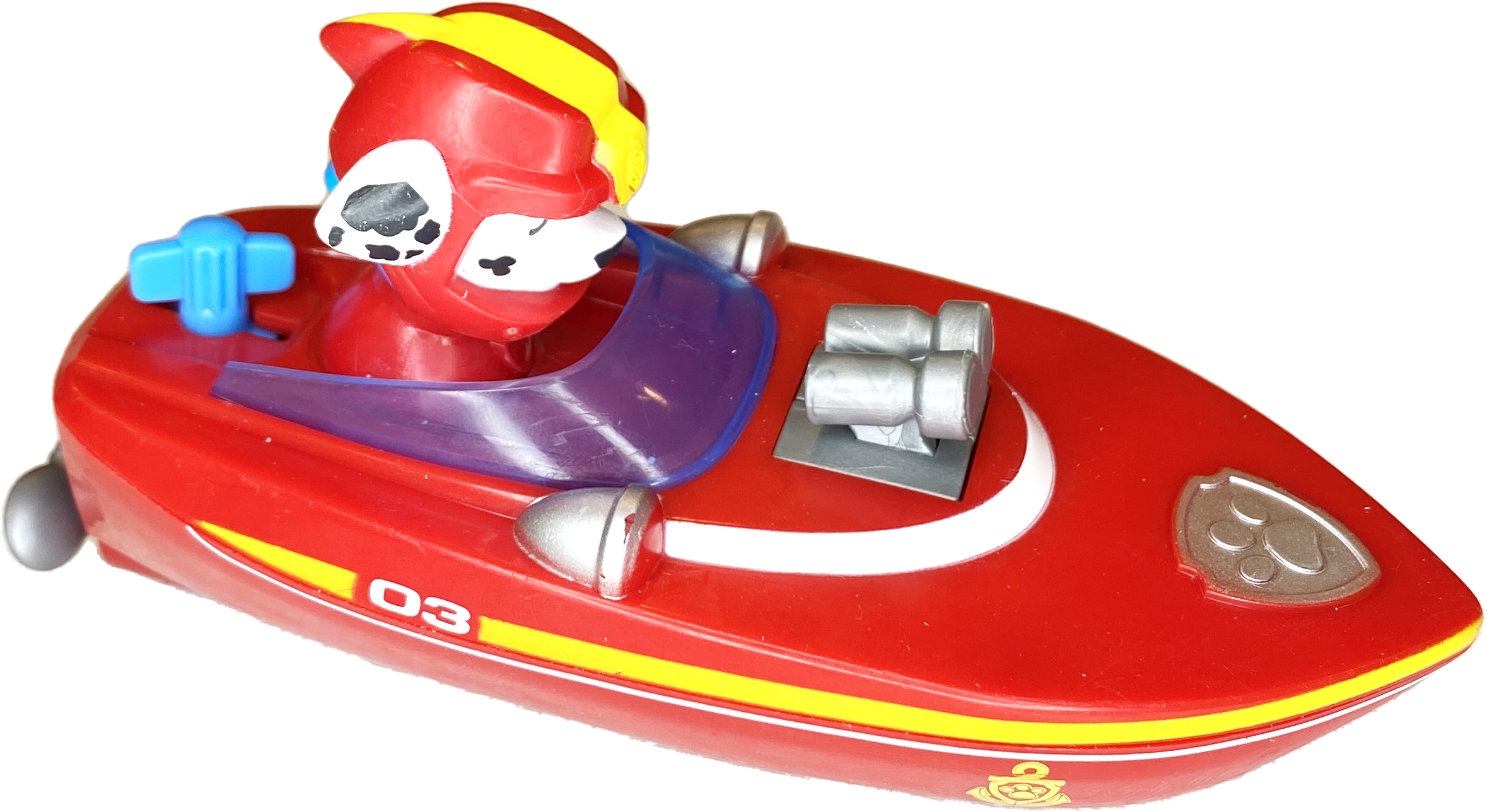}
\end{minipage}    &     85.7      &
 \begin{minipage}{.06\textwidth}
\includegraphics[width=1.5\linewidth]{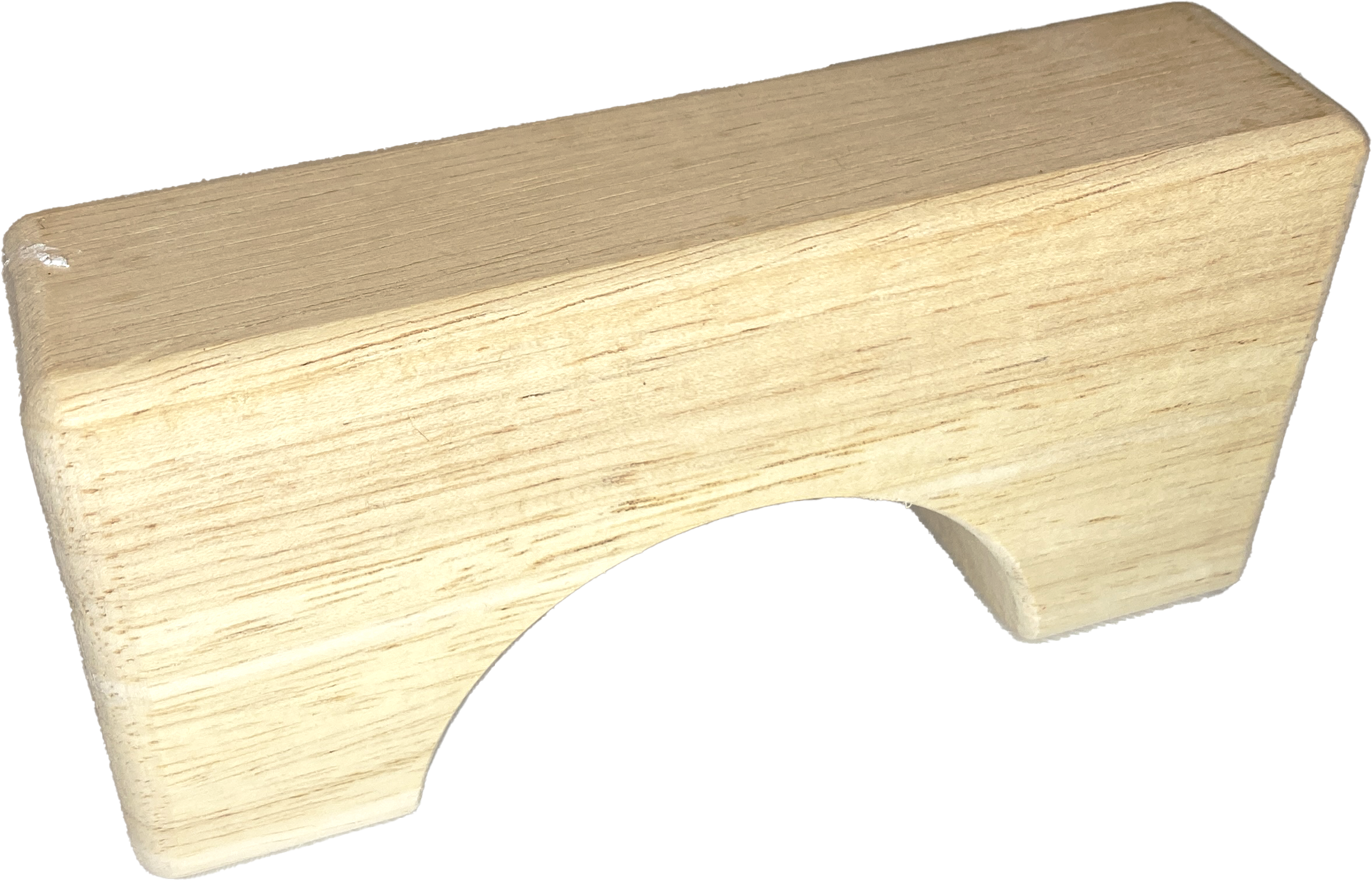}
\end{minipage}      &      180.9   \\   \begin{minipage}{.06\textwidth}
\includegraphics[width=0.6\linewidth]{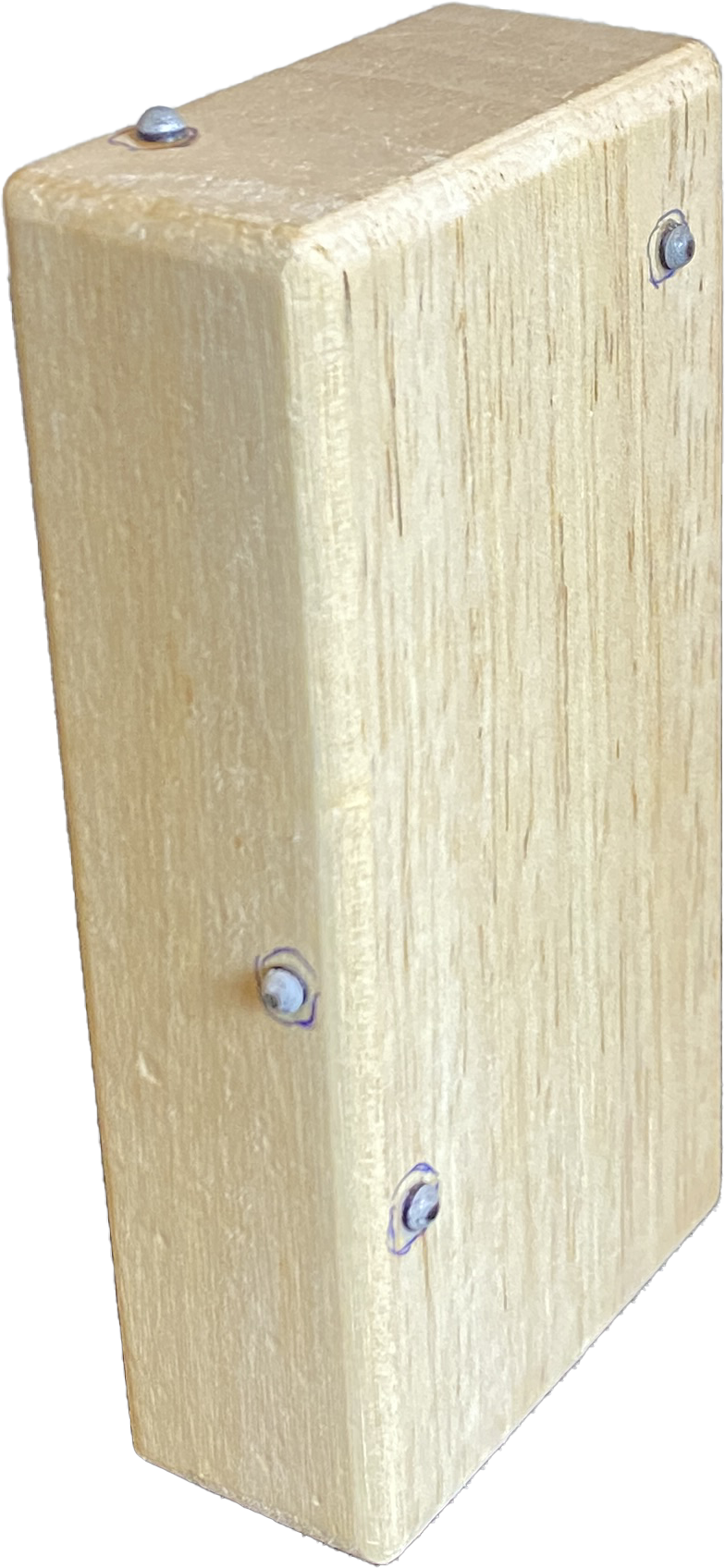}
\end{minipage}    &     244.0     &
 \begin{minipage}{.06\textwidth}
\includegraphics[width=\linewidth]{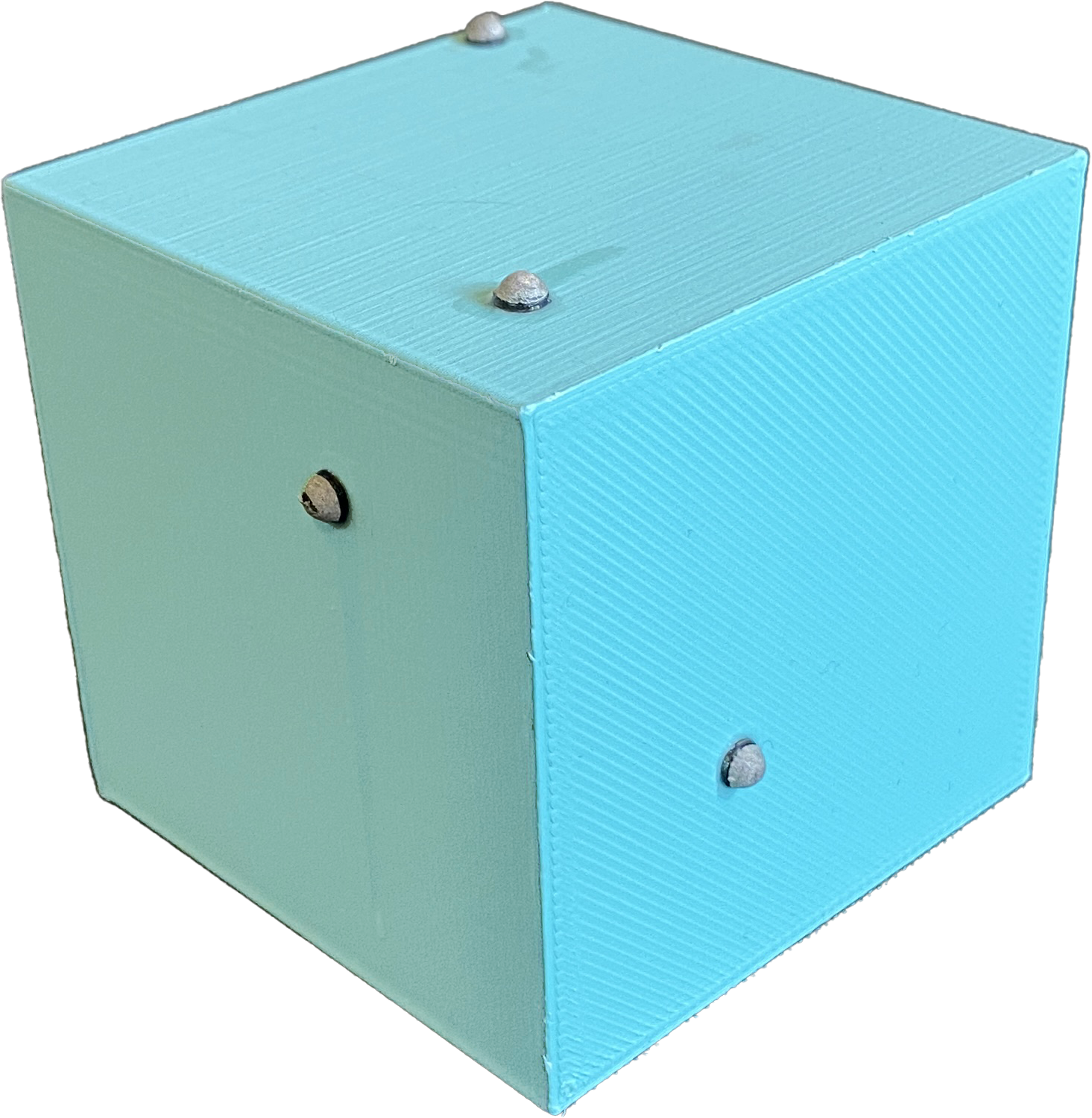}
\end{minipage}      &      127.9    &     \begin{minipage}{.06\textwidth}
\includegraphics[width=1.5\linewidth]{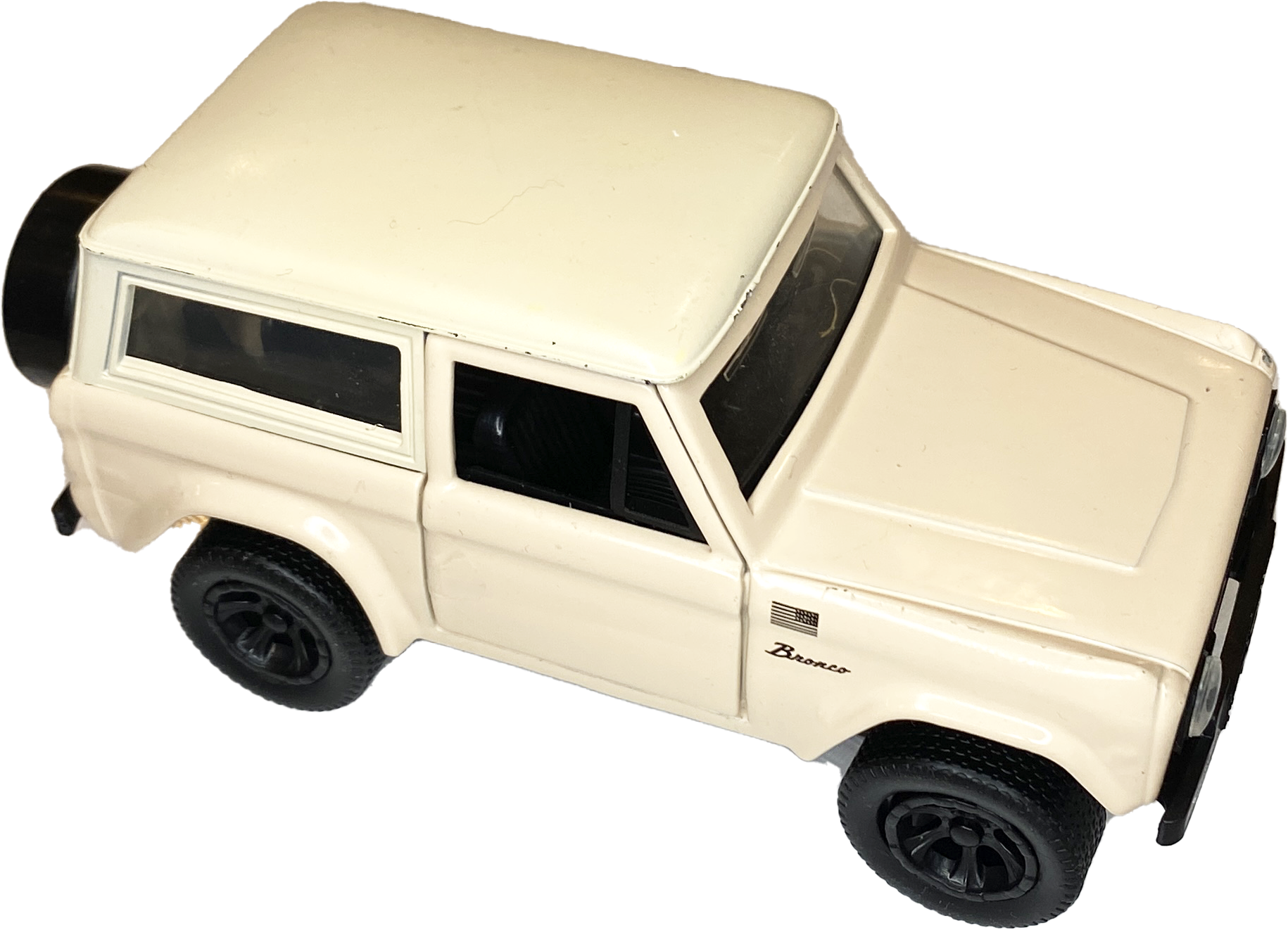}
\end{minipage}    &     137.1      &
 \begin{minipage}{.06\textwidth}
\includegraphics[width=\linewidth]{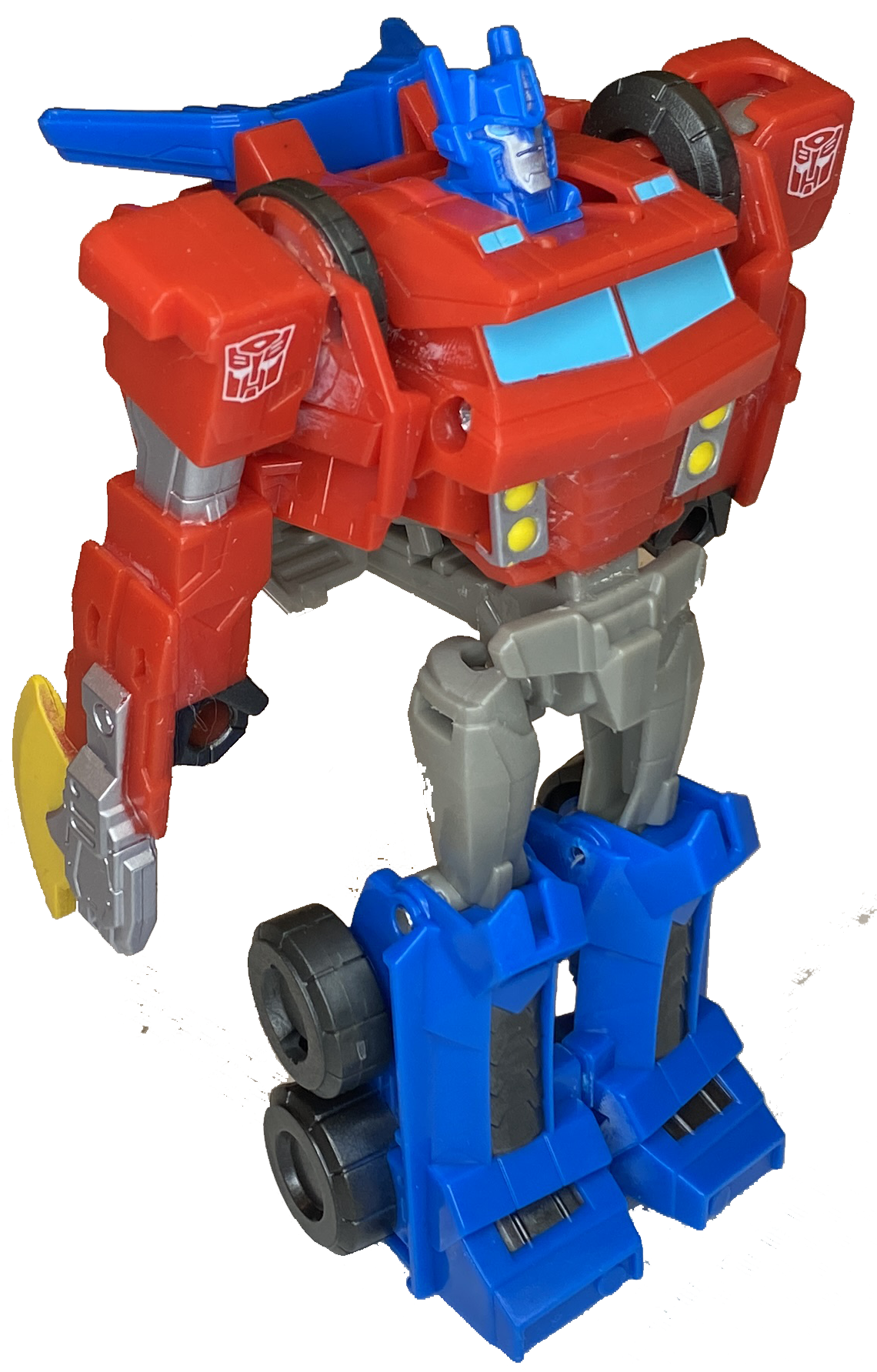}
\end{minipage}      &      67.1         \\
\hline
\end{tabular}
\end{table}
\renewcommand{\arraystretch}{1.}

\renewcommand{\arraystretch}{1.2}
\begin{table}[!tb]
\centering
\caption{Dynamics Randomization and Noise}
\label{tbl:dyn_ran_params}
\resizebox{\columnwidth}{!}{
\begin{tabular}{cccccc} 
\hline
Parameter          & Range                       & Parameter                             & Range                      & Parameter                      & Range                                                                                                                                                                                                       \\ 
\hline
state observation  & $+\gaussian(-0.002, 0.002)$ & action                                & $+\gaussian(0, 0.05)$      & joint stiffness                & $\times\uniform(0.8, 1.2)$                                                                                                                                                                                  \\
joint damping      & $\times\uniform(0.8, 1.2)$  & link mass                             & $\times\uniform(0.8, 1.2)$ & friction for robot, objects & $\uniform(0.24, 1.6)$                                                                                                                                                                                       \\
friction for table & $\uniform(0.05, 1.0)$       & restitution & $\uniform(0.0, 1.0)$       & object size scale              & $\uniform(0.95, 1.05)$                                                                                                                                                                                             \\
object mass        & $\uniform(0.009,0.324)$ kg    &                                       &                            &                                &                                                                                                                                                                                                             \\
\hline
\multicolumn{6}{l}{\begin{tabular}[c]{@{}l@{}}$\gaussian(\mu, \sigma)$: Gaussian distribution with mean $\mu$ and standard deviation $\sigma$.\\ $\uniform(a, b)$: uniform distribution between $a$ and $b$.~\\ $+$: the sampled value is added to the original value of the variable. $\times$: the original value is scaled by the sampled value.\end{tabular}}    \\
\hline
\end{tabular}}
\end{table}
\renewcommand{\arraystretch}{1.}

\newpage

\end{appendices}

\end{document}